\documentclass[10pt]{article}
\usepackage{amsmath}
\usepackage{graphicx}
\usepackage{enumerate}
\usepackage{natbib}
\usepackage{url} 
\usepackage{xr-hyper}

\usepackage{algorithm}
\usepackage{algpseudocode}
\usepackage{booktabs} 
\usepackage{customerized_notations}
\usepackage{amsmath}
\usepackage{amssymb}
\usepackage{mathtools}
\usepackage{amsthm}
\usepackage{multirow}
\newtheorem{theorem}{Theorem}[section]
\newtheorem{proposition}[theorem]{Proposition}
\newtheorem{lemma}[theorem]{Lemma}
\newtheorem{corollary}[theorem]{Corollary}
\theoremstyle{definition}

\newtheorem{assumption}[theorem]{Assumption}
\theoremstyle{remark}

\usepackage{marginnote}

\def\spacingset#1{\renewcommand{\baselinestretch}%
{#1}\small\normalsize} \spacingset{1}

\title{\bf Feature-wise and Sample-wise Adaptive Transfer Learning for High-dimensional Linear Regression}

\author{
    Zelin He\textsuperscript{a}, Ying Sun\textsuperscript{b}, Jingyuan Liu\textsuperscript{c,*}, Runze Li\textsuperscript{a}
}

\date{} 

\begin{document}

\maketitle

\renewcommand{\thefootnote}{\alph{footnote}} 
\footnotetext[1]{Department of Statistics, Pennsylvania State University.}
\footnotetext[2]{School of Electrical Engineering and Computer Science, Pennsylvania State University.}
\footnotetext[3]{MOE Key Laboratory of Econometrics, Department of Statistics and Data Science in School of Economics, Wang Yanan Institute for Studies in Economics, and Fujian Key Lab of Statistics, Xiamen University.}

\setcounter{footnote}{0} 
\renewcommand{\thefootnote}{\fnsymbol{footnote}} 
\footnotetext[1]{Corresponding author. Email: jingyuan@xmu.edu.cn.}

\bigskip
\begin{abstract}
We consider the transfer learning problem in the high dimensional {linear regression} setting, where the feature dimension is larger than the sample size. To learn transferable information, which may vary across features or the source samples, we propose an adaptive transfer learning method that can detect and aggregate the feature-wise (\textit{F-AdaTrans}) or sample-wise (\textit{S-AdaTrans}) transferable structures. We achieve this by employing a fused-penalty, coupled with weights that can adapt according to the transferable structure. 
To choose the weight, we propose a theoretically informed, data-driven procedure, enabling \textit{F-AdaTrans} to selectively fuse the transferable signals with the target while filtering out non-transferable signals, and \textit{S-AdaTrans} to obtain the optimal combination of information transferred from each source sample. We show that, with appropriately chosen weights, \textit{F-AdaTrans} achieves a convergence rate close to that of an oracle estimator with known transferable structure,
 and \textit{S-AdaTrans}  recovers existing near-minimax optimal rates as a special case. The effectiveness of the proposed method is validated using both synthetic and real data, demonstrating favorable performance compared to the existing methods. 
\end{abstract}

\noindent%

\spacingset{1.9}

\section{Introduction}
Transfer learning is a technique that incorporates knowledge from various source tasks into a target task 
 \citep{torrey2010transfer}. This approach is especially useful in high-dimensional data analysis, such as medical imaging and genetic data analysis, where the number of features is significantly larger than the target sample size. 
A fundamental challenge in transfer learning, especially under the high-dimensional setting, is addressing the negative transfer issue: if different source samples are equally aggregated into the training procedure, it could lead to a significant deterioration of the estimation accuracy \citep{li2022transfer, li2023transfer, li2023estimation}. Although several  studies have explored the identification of transferable samples, they often pre-select these samples prior to training via some ad-hoc fitting methods \citep{tian2022transfer, liu2023unified}. These methods typically result in a binary ``all-or-none" decision regarding the transfer of information, which may not be optimal in practice.

\begin{figure}[ht]
    \centering    \includegraphics[width=\linewidth]{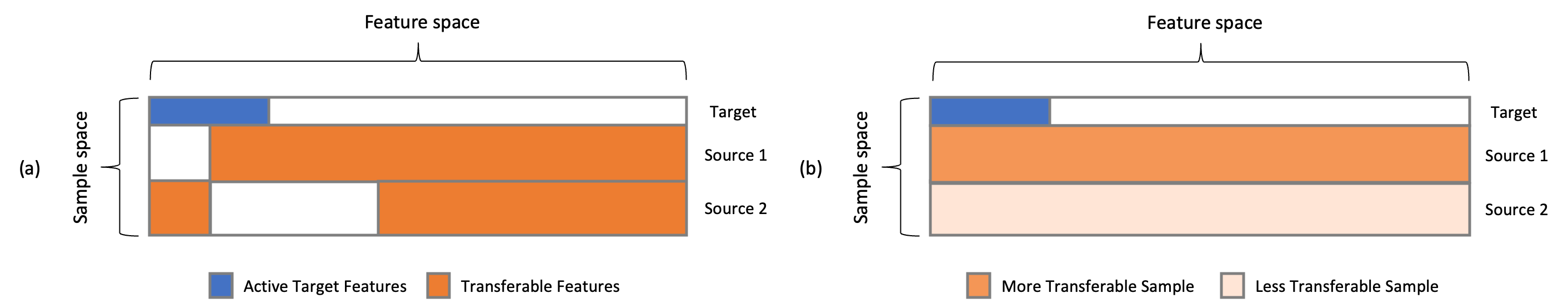}
    \caption{
    (a) A special case of feature-specific transferable structure, where two sources possess non-overlapping non-transferable features. For each source, transferable features are shaded in orange, while non-transferable ones are left blank. 
    (b) Sample-specific transferable structure. The darkness of orange depicts the transferability of each source. In both (a) and (b), truly sparse active target signals are highlighted in blue.}
    \label{fig:parameter-settings}
\end{figure}

In fact, we notice that in high-dimensional setting, the transferable structure often \textit{varies across features} within the same source sample. One such example is the analysis of high-dimensional brain functional connectivity patterns, where patterns learned from healthy samples and various disease samples can be transferred to a sample with a target disorder \citep{li2018novel}. While the feature spaces of brain connectivity are accordant across samples, each source sample is observed to possess a distinct set of features that are non-transferable due to variations in brain conditions, as depicted in Figure \ref{fig:parameter-settings}(a).
In other applications, the overall source informative level \textit{varies across source samples}, affected by factors such as data quality, reliability, and relevance to the target task. For instance, 
in genetic-disease association studies, due to patient demographics and measurement methodologies, source-task similarity with the target often varies in a whole-sample manner \citep{tang2016fused, wahlsten2003different}. Figure \ref{fig:parameter-settings}(b) illustrates such a scenario, where the darker orange color in the first source indicates that this source is more relevant to the target task 
while the second source is less transferable.

The above feature-specific and sample-specific transferable structures inspire us to propose an adaptive transfer learning framework under the high-dimensional setting, to adaptively adjust the information transferred from the sources to the target while estimating parameters.  Specifically, we present a solution for achieving \textit{feature-} and \textit{sample-wise} adaptive transfer learning, called {\it AdaTrans}, in high-dimensional linear regression involving $K$ source tasks. The target model is $p$-dimensional with sparsity level $s$.
The method is based on a fused-penalty 
that extracts the non-transferable signals from the sources and merges the transferable part with the target. We then tailor the method to address the unique challenges of feature-wise and sample-wise adaptation respectively. Our contributions are as follows:

For scenarios where the source samples contain different feature-wise transferable structures, we propose {\it F-AdaTrans}, which can detect non-transferable features for each source. We achieve this by assigning feature-wise weights to the fused-penalty in the objective function. When the transferable structure is known, \textit{F-AdaTrans} yields an ``oracle'' estimator, 
the convergence rate of which is shown to be of the order $\kappa_{F} \sqrt{s \log s /N}$, where $N=n_{T} + Kn_{S}$ is the full sample size, $n_T$ and $n_S$ are respectively the target and source sample size. The factor $\kappa_{F}$, ranging from $O(1)$ to $O(\sqrt{N/n_T})$, determines the enhanced estimation efficiency compared with single-task benchmarks. When the transferable structure is unknown, we introduce a weighting procedure based on folded concave penalization \citep{fan2001variable}. We prove that the proposed method is able to detect the transferable structure with high probability, yielding a “sub-oracle” estimator. As the detection power of the target sample increases, this estimator progressively approaches the performance of the oracle estimator.



For scenarios where the non-transferable signals are dense,  we propose \textit{S-AdaTrans} where {sample-wise} weight $\boldsymbol{w} \in \mathbb{R}^{(K+1)}$ are imposed to the penalty term. We prove that  the estimator attains a non-asymptotic rate $O\big(\kappa_{S}(\boldsymbol{w}) \cdot s \log p/N + \bar{h}(\boldsymbol{w})\sqrt{\log p/n_{T}} \big)$. The factors $\kappa_{S}(\boldsymbol{w})$ and $\bar{h}(\boldsymbol{w})$, dependent on $\boldsymbol{w}$, reveal a tradeoff between the loss of effective sample size and the suppression of negative transfer.
The bound recovers near minimax optimal rates in special cases, showing the tightness of our analysis. {We then provide closed-form solutions of the optimal weights in special cases}. The result provides a natural guideline for determining the weights: we show with the optimal weights, \textit{S-AdaTrans} provably improves over the single-task benchmark,  achieving robustness against negative transfer. To determine the practically unknown weights, we implement a cross-validation-based weight selection procedure, with its effectiveness demonstrated across diverse simulation settings. In addition, \textit{AdaTrans} is designed to jointly learn the target and all source parameters, rather than pooling the samples all together with a unified parameter vector. Thus, it can automatically account for the covariate-shift issue, which is another typical challenge for high-dimensional transfer learning.

\subsection{Related Works}
There are several transfer learning methods for high-dimensional regression problems  [e.g. \cite{li2022transfer, tian2022transfer, li2023estimation, liu2023unified, he2024transfusion}]. Despite the growing interest, none of existing works have considered \textit{learning features adaptively} from the source. 
As for the \textit{sample adaptive learning}, most of these works adopt a discontinuous decision strategy, where a source sample is either included or completely discarded. This fundamentally differs from our sample-adaptive method, where the weights can take continuous values. 
Consequently, our method includes and continuously interpolates the binary decision options, achieving a better performance by exploring a larger search space.

In the low-dimensional setting, 
there are some existing works focusing on adaptively learning from heterogeneous sources. For instance, in domain adaptation, sample reweighing technique has been studied to align the source and target distributions \citep{shimodaira2000improving,cortes2010learning,fang2020rethinking}. When considering multiple sources,  
the generalization error of the model parameter obtained by minimizing empirical risk weighted across the tasks has been derived \citep{ben2010theory,zhang2012generalization,konstantinov2019robust}. Then weight can be chosen by minimizing the bound. 
{The problem has also been studied in the multi-task learning setting \citep{li2019spatial, zhang2022learning, tang2016fused, duan2023adaptive}.} While non-exhaustively listed, methods that fall into the above categories usually require estimating the source and target data probability distribution or the sample size is larger than the feature dimension. Consequently, they are not applicable in the high dimensional setting where the feature dimension far exceeds the sample size.

Compared to sample-adaptive methods, the works related to feature-wise adaptive learning are scarce. One line is vertical federated learning~\citep{yang2019federated}, where the data is partitioned by features. The study therein focuses on computation methods rather than statistical performance. In this context, the fused penalty, as employed in our paper, provides a natural framework for incorporating adaptive transfer across features. Previous works \citep{liu2023unified, he2024transfusion} have explored similar fused penalties, but they do not incorporate an adaptive weighting strategy. Moreover, these methods rely on restrictive assumptions about the source-target relationship to identify the target parameter: \citet{liu2023unified} assumes a target sample size comparable to that of the source samples to achieve optimal performance, while \citet{he2024transfusion} requires sufficient diversity among source parameters (cf. Theorem 1) to avoid fundamental bias. These assumptions can be difficult to satisfy in heterogeneous transfer learning settings.

\subsection{Notation and Organizaiton} 
Throughout the paper, we define bold upper- and lower-case letters as matrices and vectors, respectively.
For a matrix $\mathbf{A} \in \mathbb{R}^{m \times n}$, we denote its $(i, j)$-th element by $\mathbf{A}_{ij}$, the $i$-th row by $\mathbf{A}_{i\cdot}$, the column-submatrix indexed by a set $S$ as $\mathbf{A}_{S}$, $\|\mathbf{A}\|_{q}$ as the operator norm induced by the $q$-norm and $ \Lambda_{\max}(\mathbf{A}) $ and $ \Lambda_{\min}(\mathbf{A}) $ as its maximum and minimum eigenvalues, respectively. For two matrices $\boldsymbol{A} \in \mathbb{R}^{m \times n}$ and $\boldsymbol{B} \in \mathbb{R}^{m \times p}$, we say $\boldsymbol{A} \perp \boldsymbol{B}$ if $\boldsymbol{A}^\top \boldsymbol{B} = \boldsymbol{0}$. We set $[K] := \{1, \ldots, K\}$. We let $a \vee b:=\max \{a, b\}$ and $a \wedge b:=\min \{a, b\}$. We use $c, c_0, c_1, \ldots$ to denote generic constants independent of $n$, $p$ and $K$. 
Let $a_n=O\left(b_n\right)$ and $a_n \lesssim b_n$ denote $\left|a_n / b_n\right| \leq c$ for some constant $c$ when $n$ is large enough; $a_n = o\left(b_n\right)$ or $b_n \gg a_n$ if $a_{n} = O(c_{n}b_{n})$ for some $c_{n} \to 0$; $a \asymp b$ if $a = O(b)$ and $b = O(a)$. 

{The structure of this paper is organized as follows: Section 2 introduces the transfer learning settings within a high-dimensional linear regression framework and outlines basic assumptions. Sections 3 and 4 delve into the \textit{F-AdaTrans} and \textit{S-AdaTrans} algorithms, respectively. 
Section 5 presents extensive simulations that demonstrate the favorable performance of \textit{AdaTrans} over other methods 
. 
In Section 6, we apply the \textit{AdaTrans} methods to predict stock recovery during the COVID-19 pandemic. Section 7 concludes the paper. Proofs, implementation details, and codes for reproducing the simulation and real data results are provided in the Supplementary materials.}

\section{Preliminaries}
We consider a high-dimensional transfer learning problem involving one target task and $K$ source tasks. For the target task, we observe a sample $(\bXz, \byz)$ generated from the linear  model
\begin{align}\label{m:target_model}
   \by^{(0)}=\bXz \bz +\bez, 
\end{align}
    where $\byz \in \mathbb{R}^{n_T}$ is the response vector, $\bXz \in \mathbb{R}^{n_T \times p}$ is the feature matrix, $\bz \in \mathbb{R}^{p}$ is the target parameter of primary interest and $\bez\in \mathbb{R}^{n_T}$ is the observation noise. The target parameter is sparse with $s:=\|\bz\|_{0}$ nonzero elements, and the feature dimension $p$ is allowed to be larger than $n_{T}$.
    

In addition to the primary target sample, suppose we have access to $K$ sets of auxiliary source samples $\{(\bXk, \byk)\}_{k=1}^K$, each generated from 
\begin{align}\label{m:source_model}
 \by^{(k)}
 =\bXk (\bz + \bdk) +\bek.   
\end{align}
The source parameter differs from that of the  target by 
 $\bdk \in \mathbb{R}^{p}$, which we termed as ``target-source contrast" or ``non-transferable signal". For simplicity, we assume the source samples share a common size, i.e.,  $n_k = n_S,\   \forall k \in [K]$, which is larger than  $n_T$.

Our goal is to estimate the target parameter $\bz$ using both target and source samples  where only (a) part of source features in each source sample are transferable or (b) part of source samples are transferable, which will be discussed in Section \ref{feature_wise} and Section \ref{sample_wise}, respectively.

{This section ends with two  standard assumptions
for high-dimensional regression analysis on feature matrix $\bXk$ and observation error $\bek$. Notice that we only impose a mild tail condition on the distribution of $\bXk$
but allow other distribution characteristics, such as the
correlation structures, to vary across the tasks.}
\begin{assumption}
\label{A1}
For any $0 \leq k \leq K$, $0 \le i \le n_k$, $\bXk_{i \cdot}$'s are independent sub-Gaussian random vectors with mean zero and covariance $\Sigk$. Furthermore, 
    there exists some constant $c$ such that $1/c \leq \min_{0 \leq k \leq K}\Lambda_{\min}(\Sigk) \leq \max_{0 \leq k \leq K}\Lambda_{\max}(\Sigk) \leq c$. 
    
\end{assumption}
\begin{assumption}
\label{A2}
    For any $0 \leq k \leq K$, $0 \le i \le n_k$, the $\bek_i$'s are independent Gaussian random variables with zero mean and variance $\sigma_k$ such that $\max_{0 \le k \le K} \sigma_k \le c$ for some constant $c$ and $\bek$ is independent of $\bXk$.
\end{assumption}

\section{Feature-wise Adaptive Transfer Learning}
\label{feature_wise}

In transfer learning, each source task may have its unique transferable structure: some of its features are transferable, whereas others are not. The transferability of the $j$th feature in the $k$-th source task can be assessed by the magnitude of target-source contrast $\bdk_{j}$ in (\ref{m:source_model}): If $\bdk_{j}$ is negligible, indicating the $j$th feature has a similar effect between the $k$-th source and the target, it is transferable. Conversely, if $\bdk_{j}$ is non-negligible, the corresponding feature should be treated as non-transferable. Specifically, we consider the following parameter space:
\begin{align}
\label{para_space_1}
\Theta_{1}:=\left\{\bz, \{\bdk\}_{k=1}^K:  \operatorname{supp}(\bz) = S_0, \operatorname{supp}(\bdk) = S_k, k \in [K]\right\}.
\end{align}
where $\operatorname{supp}(\boldsymbol{x}):=\{j \in [p]: \boldsymbol{x}_{j} \neq 0\}$.
Here $S_0$ depicts the sparsity structure of the primary parameter $\bz$. For $k \in [K]$, the features in the $k$-th source can be divided  
into two sets - the transferable set $S^c_{k}=\{j \in [p]: \bdk_j=0\}$, with zero target-source contrasts, and the non-transferable set $S_k$ with non-negligible contrasts. Note that we assume zero contrasts for the transferable sets for technical simplicity, and they can be relaxed to sufficiently small contrasts in practice. 

Next we introduce a feature-wise adaptive transfer learning  framework (\textit{F-AdaTrans}) that detects transferable features in each source, by assigning different penalty strengths 
based on the target-source contrast $\bdk_j$'s. Setting $\bdz = \boldsymbol{0}$, we consider estimating $\bz$ by solving the problem
\begin{align}\label{obj2}
& \min_{\bz, \bd^{(1)}, \dots, \bd^{(K)}}   \Big\{ \frac{1}{N} \sum^{K}_{k=0} \|\byk-\bXk (\bz + \bdk)\|_{2}^{2} +  \lz \sum^{p}_{j=1}  w^{(0)}_{j} |\bz_{j}| +  \lambda_{1} \sum^{K}_{k=1} \sum^{p}_{j=1} w^{(k)}_{j} |\bdk_{j}|  \Big\},
\end{align}
where $N = K n_{S} + n_{T}$ is the total sample size, $\lambda_{0}$ and $\lambda_{1}$ are tuning parameters and $w^{(k)}_{j}$s are the weights assigned for different target-source contrasts. The first term of \eqref{obj2} measures the overall fit of the models with parameters $\{\bz + \bdk\}_{k = 0}^K$, and the second term imposes a penalty to achieve sparsity of $\bz$. The third term of \eqref{obj2} is a fused penalty with weights that adaptively shrinks $\bdk_j$'s: ideally, it applies stronger penalties to transferable features with negligible $\bdk_{j}$, shrinks them to zero, so that the information from the $j$th feature of the $k$-th source is successfully transferred to help estimate $\bz$; meanwhile, it does not impose excessive penalties to non-transferable features with relatively large $\bdk_{j}$'s to prevent introducing bias. 

We remark that the definition of transferable/informative features inverts the usual logic of penalized regression. Recall that in the penalized regression, non-informative features are penalized to zero while informative features are preserved in the model. However, for the penalties imposed on target-source contrasts $\bdk_j$'s, ``shrinking to zero" indeed represents a transferable feature that is ``informative" to the model, while $\bdk_j$'s that are not penalized to zero correspond to ``non-informative" and non-transferable features.

We also note that the proposed weighting strategy not only enables feature-wise adaptive learning but, more importantly, enhances the identifiability of the target parameter \(\bz\). Without the weights $w_{j}^{(k)}$'s, it is challenging to distinguish between \(\bz\) and \(\bdk\) in the $k$th source data. Other methods, such as \cite{li2023estimation}, and the S-AdaTrans algorithm introduced in the next section address this issue by incorporating additional nonsmooth constraints. However, these constraints complicate the optimization process, whereas the F-AdaTrans formulation in (\ref{obj2}) can be reformulated as a standard Lasso regression problem, allowing it to be efficiently solved via algorithms like iterative soft thresholding.
\subsection{Weight Choice with Known Transferable Structure}
\label{feature_known}
A natural question in feature-wise adaptive transfer learning is how to determine the weights $w^{(k)}_j$'s. To answer this, we first consider an ideal scenario where the sparsity and transferable structure are known. 
In this setting, 
we can define ``oracle estimator" $\hbz_{\operatorname{ora}}, \hat{\bd}^{(1)}_{\operatorname{ora}}, \dots, \hat{\bd}^{(K)}_{\operatorname{ora}} $ via
\begin{equation}\label{oracle_def}
    \begin{aligned}
        & \min_{\bz, \{\bd^{(k)}\}_{k = 1}^K}&& \hspace{-0.2cm} \frac{1}{N} \sum^{K}_{k=0} \|\byk-\bXk (\bz + \bdk)\|_{2}^{2} \\
        & \qquad \text{s.t.} && \bb^{(0)}_{S^c_0} = 0,\  \bd^{(k)}_{S^c_k} = 0, \forall k \in [K].
    \end{aligned}
\end{equation}
Based on the above definition, the optimal weights in this oracle scenario should be chosen such that solving problem (\ref{obj2}) would yield the oracle estimator in (\ref{oracle_def}). The choice of such oracle weight is provided in the following theorem. 
\begin{theorem}
\label{oracle_with_known_structure}
Given Assumptions \ref{A1} and \ref{A2}, and provided that $n_{S} \gtrsim \log p$, if we choose $w^{(k)}_{j} = \boldsymbol{1}_{\{j \in S^c_k\}}$ for $k = 0, \dots, K$, $\lambda_{0} \gtrsim \sqrt{\frac{\log p}{N}}$ and $\lambda_{1} \gtrsim \frac{n_S}{N} \sqrt{\frac{\log p}{n_S}}$, then with probability larger than $1-c_{1} \exp \left(-c_{2} \log p \right)$, the solution of problem (\ref{obj2}) satisfies (\ref{oracle_def}).
\end{theorem}
Theorem \ref{oracle_with_known_structure} states that if we choose the weight such that only the transferable set of features are penalized, then with a sufficiently strong penalty, solving the problem (\ref{obj2}) could yield an oracle estimator that fully utilizes the transferable structure. Next, we provide the explicit form of the oracle estimator of $\bz$.
\begin{proposition}
\label{OLS_trans_form}
If $|S_0| < n_T$ and $\max_{1 \le k \le K} |S_k| < n_S$, the solution to problem (\ref{oracle_def}) satisfies $\hbz_{\operatorname{ora}, S_0^c} = \boldsymbol{0}$ and
{\small
\begin{align}
\label{oracle_form}
\begin{split}
        \hbz_{\operatorname{ora}, S_0} = \left[ (\bXz_{S_0})^\top \bXz_{S_0} + \sum^{K}_{k=1}(\bX^{(k)}_{S_0})^\top (\boldsymbol{I} - \boldsymbol{H}_{S_k}^{(k)})\bX^{(k)}_{S_0} \right]^{-1} \left[ (\bXz_{S_0})^\top \byz + \sum^{K}_{k=1}(\bX_{S_0}^{(k)})^\top (\boldsymbol{I} - \boldsymbol{H}_{S_k}^{(k)})\byk \right], 
\end{split}
\end{align}
}
where $\Hk:=\bX_{S_k}^{(k)}[(\bXk_{S_k})^\top \bXk_{S_k}]^{-1}(\bXk_{S_k})^\top$ is the projection matrix onto the column space of $\bXk_{S_k}$.  
\end{proposition}
To better understand $\hbz_{\operatorname{ora}, S_0}$, we rewrite 
$$
\hbz_{\operatorname{ora}, S_0} = [\tbX_{S_0}^\top\tbX_{S_0}]^{-1}\tbX_{S_0}^\top\by \quad \text{and} \quad \hbz_{\operatorname{ora}, S_0^c} = \boldsymbol{0},
$$
where $\tbX_{S_0} =((\bXz_{S_0})^\top, (\tilde{\bX}_{S_0}^{(1)})^\top, \dots, (\tilde{\bX}_{S_0}^{(K)})^\top)^\top$, 
with  $\tbXk_{S_0}=(\boldsymbol{I} - \Hk)\bXk_{S_0}$.
Note that $\tbXk_{S_0}$ is the projection of $\bXk_{S_0}$, our active feature matrix, onto the \textit{null space} of $\bXk_{S_k}$, the non-transferable feature matrix. The projection essentially filters out the non-transferable components, and the ``overlap'' between $\bXk_{S_0}$ and $\bXk_{S_k}$ represents the non-transferable proportion of the $k$-th source:
if the non-transferable feature matrix $\bXk_{S_k}$ is orthogonal to the target feature matrix $\bXk_{S_0}$ (i.e., $\bXk_{S_k} \perp \bXk_{S_0}$), then $\tbXk_{S_0} = \bXk_{S_0}$; in this case, the $k$-th source sample is fully transferable. On the other hand, if $\bXk_{S_k}$ is linearly related to $\bXk_{S_0}$,  as would be the case when $S_0 \subset S_k$, then $\tbXk_{S_0} = 0$, indicating that the entire $k$-th source is non-transferable. Thus, $\tbX_{S_0}$ is indeed the collection of the ``transferable part'' of each source sample, along with the full target sample.


In the following proposition, we study the estimation error of the oracle estimator.
\begin{proposition}\label{oracle_rate}
Under Assumptions \ref{A1} and \ref{A2}, if $s = |S_0| < n_T$, $\max_{1 \le k \le K} |S_k| < n_S$ and $N \ge \log p$, then we have with probability larger than$1-c_{1} \exp \left(-c_{2} \log p \right)$,
\begin{align}
\label{ora-rate}
\|\hbz_{\operatorname{ora}} - \bz\|_{2} \lesssim \kappa_{F}  \left\| \hat{\boldsymbol{\Omega}}_{S_0, N}\right\|_{\infty} \sqrt{\frac{s \log s}{N}},
\end{align}
 where $\bX_{S_0}$ is column-submatrix indexed by $S_0$ of the full-sample design matrix $\bX$, and
\begin{align}
\label{kappa}
\hat{\boldsymbol{\Omega}}_{S_0, N} := \left(\frac{\bX_{S_0}^\top\bX_{S_0}}{N}\right)^{-1}, \quad
  \kappa_{F} := \frac{\left\| [\tbX_{S_0}^\top\tbX_{S_0}]^{-1}\tbX_{S_0}^\top \boldsymbol{\epsilon} \right\|_{\infty}}{\left\|  [\bX_{S_0}^\top\bX_{S_0}]^{-1}\bX_{S_0}^\top \boldsymbol{\epsilon} \right\|_{\infty}}.  
\end{align}
\begin{itemize}
    \item[(1)] If for $k \in [K]$, $\bXk_{S_0} \perp \bXk_{S_k}$, we then have $\|\hbz_{\operatorname{ora}} - \bz\|_{2} \lesssim \left\| \hat{\boldsymbol{\Omega}}_{S_0, N}\right\|_{\infty} \sqrt{s\log s / N}$.
    \item[(2)] If for $k \in [K]$, $S_0 \subset S_k$, we then have $\|\hbz_{\operatorname{ora}} - \bz\|_{2} \lesssim \left\| \hat{\boldsymbol{\Omega}}_{S_0, N}\right\|_{\infty} \sqrt{s\log s / n_{T}}$.
\end{itemize}
\end{proposition}
In error bound (\ref{ora-rate}), the term $\sqrt{s \log s / N}$ is the nearly \textit{oracle rate} for estimating $\bz$ with \textit{known support} using the full samples of size $N$ - it reveals the benefit of transferring the source information when estimating the $\bz$.
$\left\| \hat{\boldsymbol{\Omega}}_{S_0, N}\right\|_{\infty}$ is a full-sample estimate of the precision matrix and thus has a bounded $\ell_{\infty}$ norm with high probability. 
The term $\kappa_{F}$ in \eqref{kappa}  measures the transferability of source datasets. When all source datasets are transferable in the sense that $\bXk_{S_k} \perp \bXk_{S_0}$ for all $k \in [K]$, then $\kappa_{F} = 1$. In this ideal case, we can utilize all information from size-$N$ full sample when estimating $\bz$. Otherwise, if all source tasks are completely non-transferable in the sense that $S_0 \subset S_k$ for all $k\in [K]$, we have $\kappa_{F} \asymp \sqrt{N/n_{T}}$. In this case, the estimation rate becomes $\sqrt{s \log s/{n_{T}}}$, which is actually the single-task minimax optimal rate for estimating $\bz$ with known support up to a logarithm order. Between these two extremes, each source contains both transferable and non-transferable features, hence $\kappa_{F}$ reflects the proportion of transferable information among the full sample.

\subsection{Weight Choice with Unknown Transferable Structure}
\label{sec:feature_unknown_weights}

We now introduce a data-driven approach to determine the weights in \eqref{obj2} when the non-transferable signals $\bdk_j$s are unknown,  which is often the case in practice.
Recall that to achieve feature-wise adaptive transfer learning, one should impose a relatively larger penalty on negligible $\bdk_{j}$ and a relatively smaller penalty on non-negligible $\bdk_{j}$. A suitable candidate for this purpose is the family of folded-concave penalties, denoted by $\mR_{\lambda}(t)$, including SCAD \citep{fan2001variable}, MCP  \citep{zhang2010nearly}, etc. See a formal definition of a folded-concave penalty in Appendix C.1. 
\begin{figure}
    \centering
\includegraphics[width = 0.8\linewidth]{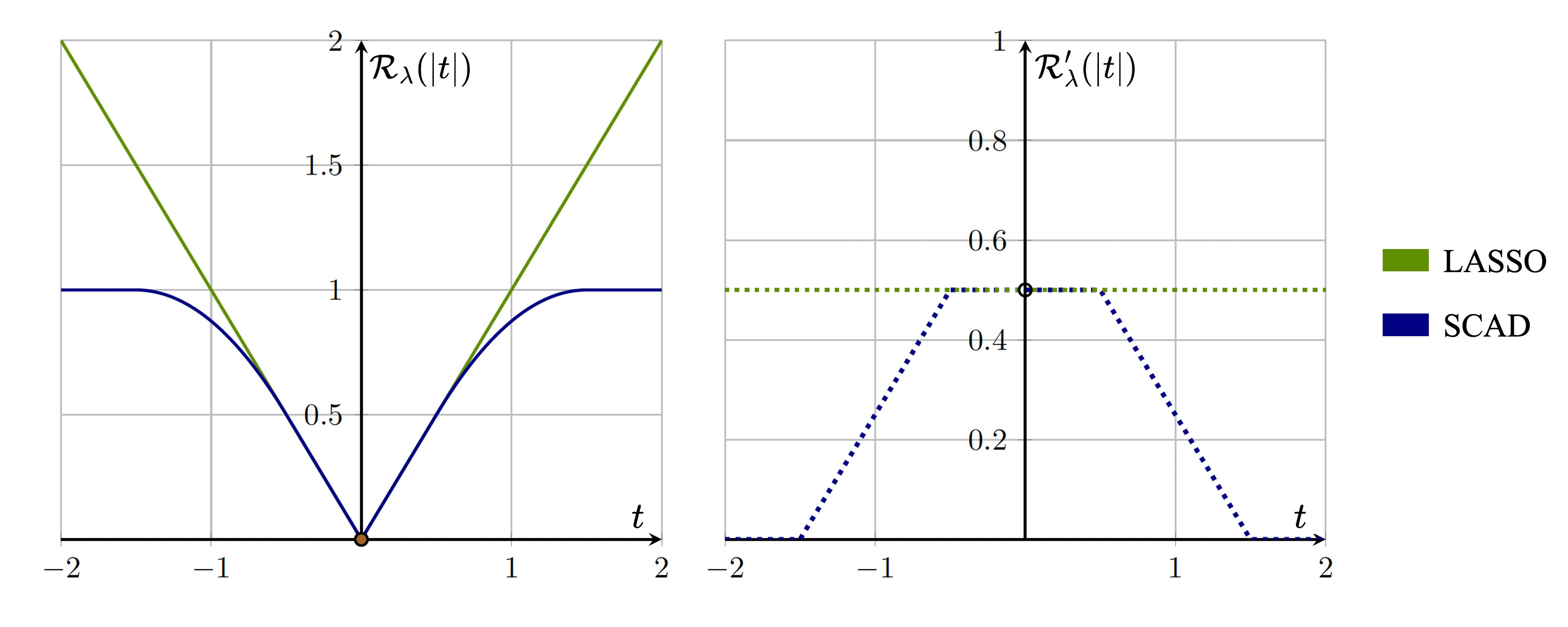}
 \caption{Penalty Functions. SCAD (blue) and Lasso (green)  shown on the left; their first-order derivatives are on the right.}
 \label{fig:SCAD}
\end{figure}
Figure \ref{fig:SCAD} depicts the SCAD penalty function along with its derivative, which demonstrates a linearly increasing penalty to small entries to promote sparse solution and information transfer, and a constant penalty to large entries to reduce bias and negative transfer. Inspired by the local linear approximation (LLA) algorithm \citep{zou2008one, fan2014strong}, we consider the first-order Taylor approximation of the folded-concave penalties for weight construction:
\begin{align*}
 \mR_\lambda(|\bz_{j}|) &\approx \mR_\lambda(|\hbz_{\text{init}, j}|)+\mR_\lambda^{\prime}(|\hbz_{\text{init}, j}|) (|\bz| - |\hbz_{\text{init}, j}|), \\
 \mR_\lambda(|\bdk_{j}|) &\approx \mR_\lambda(|\hbdk_{\text{init}, j}|) 
 + \mR_\lambda^{\prime}(|\hbdk_{\text{init}, j}|) (|\bdk| - |\hbdk_{\text{init}, j}|), \ k \in [K],
\end{align*}
where $\mR_{\lambda}^{\prime}(t)$ is the first-order derivative of $\mR_{\lambda}(t)$, and $\hbdk_{\text{init}, j}$ and $\hbz_{\text{init}, j}$ are some initial estimators. Based on this, we set the weights in \eqref{obj2} as $\lambda_{0}w^{(0)}_{j} =\mR_{\lz}^{\prime}(|\bz_{\text{init}, j}|)$ and $\lambda_{1} w^{(k)}_{j} = \mR_{\lambda_{1}}^{\prime}(|\bdk_{\text{init}, j}|)$. With an appropriate initial estimator, this approach should mirror the effects of the non-convex folded concave penalty while being much more computationally efficient. The full procedure of \textit{F-AdaTrans} is presented in Algorithm \ref{alg:fadatrans}.

\begin{algorithm}
\caption{\textit{F-AdaTrans}}
\label{alg:fadatrans}
\begin{algorithmic}[1]
\State \textbf{Input:} target data $\left(\bXz, \byz\right)$, source data $\left\{\left(\bXk, \byk\right)\right\}_{k=1}^K$, tuning parameters $\lambda_0$ and $\lambda_1$, folded concave penalty function $\mR_{\lambda}^{\prime}(t)$.
\State \textbf{Output:} the estimated target coefficient $\hat{\boldsymbol{\beta}}^{(0)}$.
\State \textit{Weight Construction:} Run an initialization algorithm (e.g. separate Lasso regression) to obtain the initial estimator $\hbz_{\text{init}}$ and $\hbdk_{\text{init}}$, then construct the feature-wise adaptive weight $\hat{w}_{j}^{(0)} = \lz^{-1}\mR_{\lz}^{\prime}(|\hbz_{\text{init}, j}|)$ and $\hat{w}_{j}^{(k)} = \lk^{-1}\mR_{\lk}^{\prime}(|\hbdk_{\text{init}, j}|)$.

\State \textit{Feature-wise Adaptive Transfer:} Compute $\hbz$ by solving
\Statex $ \arg \min_{\bz, \{\bd^{(k)}\}_{k=1}^K}   \Big\{ \frac{1}{N} \sum^{K}_{k=0} \|\byk-\bXk (\bz + \bdk)\|_{2}^{2} +  \lz \sum^{p}_{j=1}  \hat{w}^{(0)}_{j} |\bz_{j}| +  \lambda_{1} \sum^{K}_{k=1} \sum^{p}_{j=1} \hat{w}^{(k)}_{j} |\bdk_{j}|  \Big\}.$
\end{algorithmic}
\end{algorithm}

In most applications, recovering the entire transferable structure with a limited target sample is challenging. Therefore, we first introduce the idea of detectable and non-detectable transferable structures.
We partition each non-transferable set $S_k, \ k=[K]$, into two parts, the non-detectable non-transferable set $A_k \subset S_k$ and the detectable non-transferable set $S_k \setminus A_k$.  Similarly, the truly active set $S_0$ can be partitioned into $A_0$ and $S_0\setminus A_0$. Specifically, we have the following definition:
{\small
\begin{align}
\label{eq:partial_setup}
\begin{array}{ll}
\left\{
\begin{array}{ll}
|\bdk_j| \geq h_k^{\wedge}, & \text{if } j \in S_k \setminus A_k , \\
h_k^{\wedge} > |\bdk_{j}| > 0, & \text{if } j \in A_k, \\
\bdk_{j} = 0, & \text{if } j \in S_k^c , 
\end{array} \quad \text{for}\ k \in [K],
\right.
\end{array}
\begin{array}{ll}
\left\{
\begin{array}{ll}
|\bz_j| \geq h_0^{\wedge}, & \text{if } j \in S_0 \setminus A_0 , \\
h_0^{\wedge} > |\bz_{j}| > 0, & \text{if } j \in A_0, \\
\bz_{j} = 0, & \text{if } j \in S_0^c.
\end{array}
\right.
\end{array}
\end{align}
}
{Here, $h_k^{\wedge}$'s can be understood as
the minimum signals for a transferable structure to be detectable. The choice of $h_k^{\wedge}$'s depends on the detection capacity of the initial estimator and will be specified later.} When $A_k = \emptyset$ for all $k \in [K]$ and $A_0 = \emptyset$, all structures are detectable. Conversely, if $A_k = S_k$ and $A_0 = S_0$, the structures are entirely non-detectable. For simplicity, we treat the non-detectable set as a whole and define $A = \cup_{k=0}^K A_k$ as the full non-detectable set. Under such a setup, we consider a "sub-oracle estimator" as
{\small
\begin{equation}\label{sub_oracle_def}
    \begin{aligned}
        \hbz_{\operatorname{sub}}, \hat{\bd}^{(1)}_{\operatorname{sub}}, \dots, \hat{\bd}^{(K)}_{\operatorname{sub}} \in & \min_{\bz, \{\bd^{(k)}\}_{k = 1}^K}&& \hspace{-0.2cm} \frac{1}{N} \sum^{K}_{k=0} \|\byk-\bXk (\bz + \bdk)\|_{2}^{2}+\lambda_0\left\|\bz_{A}\right\|_1+\lambda_1 \sum_{k=1}^K\left\|\bd_{A}^{(k)}\right\|_1 \\
        & \qquad \text{s.t.} && \bb^{(0)}_{S^c_0 \backslash A} = 0,\  \bd^{(k)}_{S^c_k \backslash A} = 0, \forall k \in [K].
    \end{aligned}
\end{equation}
}

The sub-oracle estimator accounts for the existence of non-detectable components $A$ in the transferable structure due to the limited target sample size. When $A = \emptyset$, it degenerates to the oracle estimator defined in (\ref{oracle_def}). In the next theorem, we analyze the close form of the sub-oracle estimator and show that for any given initial estimator, with a proper choice of tuning parameter $\lz$ and $\lk$, the F-AdaTrans algorithm yields the sub-oracle estimator.


\begin{theorem}
\label{oracle}
Under Assumption \ref{A1} and \ref{A2}, suppose $n_{S} \gtrsim \log p$. For any given initial estimators $\hbz_{\text{init}}$ and $\hat{\bd}_{\text{init}}^{(k)}$, if we choose $\lz$ and $\lk$ such that
\begin{align}
\frac{a_2}{2}  \lambda_0 \geq \left\|\hbz_{\text{init}}-\bz\right\|_{\infty} \vee \left(c_{0} \sqrt{\frac{\log p}{N}} \right),  \  \frac{a_2}{2} \lambda_1 \geq \max_{k\in [K]}\left\|\hat{\bd}_{\text{init}}^{(k)}-\bd^{(k)}\right\|_{\infty}  \vee \left(c_{0} \frac{n_S}{N} \sqrt{\frac{\log p}{n_S}} \right),
\end{align}
where $a > a_{2} \ge 0$ and $a_{1} \ge 0$ are constants specified in Appendix \ref{SCAD} and $c_0$ is some universal constant, then setting $h_k^\wedge=\frac{3a}{2}\lk$ for $k \in [K]$ and $h_0^\wedge=\frac{3a}{2}\lz$ in (\ref{eq:partial_setup}), we have
\begin{itemize}
\item[(1)] Partially detectable case: if $|S_0| + |A| < n_T$, $|S_k| + |A| < n_S$ and under the incoherence assumption on $\{A_k\}_{k=0}^K$ outlined in Appendix \ref{mutual_cond:sec}, then with probability larger than $1-c_{1} \exp \left(-c_{2} \log p \right)$, the F-AdaTrans method (Algorithm \ref{alg:fadatrans}) yields a unique solution such that $\hbz_{\tSzc} = \hbz_{\text{sub}, \tSzc} = \boldsymbol{0}$, $\hbdk_{\tSkc} = \hbdk_{\text{sub}, \tSkc} = \boldsymbol{0}$, and
{\small
\begin{align}
\label{form:sub_oracle_est}
\hbz_{\tSz} = \hbz_{\text{sub}, \tSz}= & {\bigg[\left(\bXztSz\right)^{\top} \bXztSz+\sum_{k=1}^{K}\left(\bXktSz\right)^{\top}\left(\bI - \bH^{(k)}_{\tSk}\right)\bXk_{\tSz}\bigg]^{-1} }  \bigg[\left(\bXztSz\right)^{\top} \by^{(0)} \nonumber \\
&+\sum_{k=1}^{K}\left(\bXktSz\right)^{\top}\left(\bI - \bH^{(k)}_{\tSk}\right)\by^{(k)}-\frac{N}{2} \lambda_{0}  \hbzz_{A}+\frac{N}{2} \lambda_{1}\sum_{k=1}^{K}\hat{\boldsymbol{B}}_{\tSz, \tSk}^{\top}  \hbzk_{A}\bigg],
\end{align}
}
where $\tSz = S_0 \cup A$; for $k \in [K]$, $\tSk = S_k \cup A$, $\bH^{(k)}_{\tSk}:=\bX_{\tSk}^{(k)}[(\bXk_{\tSk})^\top \bXk_{\tSk}]^{-1}(\bXk_{\tSk})^\top$, 
\begin{align*}
\begin{array}{ll}
(\hbzk_{A})_{j}=\left\{
\begin{array}{ll}
\operatorname{sign}(\hbdk_{j}), & \text{if } j \in A, \\
0 & \text{if } j \in \tSk \backslash A, 
\end{array},
\right.
\end{array}
\begin{array}{ll}
(\hbzz_{A})_{j}=\left\{
\begin{array}{ll}
\operatorname{sign}(\hbz_{j}), & \text{if } j \in A, \\
0 & \text{if } j \in \tSz \backslash A, 
\end{array}
\right.
\end{array}
\end{align*}
and $\hat{\boldsymbol{B}}_{\tSz, \tSk} := \left[\left(\bX_{\tSk}^{(k)}\right)^{\top}\left(\bX_{\tSk}^{(k)}\right)\right]^{-1}\left(\bX_{\tSk}^{(k)}\right)^{\top} \bX_{\tSz}^{(k)}$ is the regression coefficient of $\bX_{\tSz}^{(k)}$ on $\bX_{\tSk}^{(k)}$.
\end{itemize}

\item[(2)] Fully detectable case: if $A_0 = \emptyset$ and $A_k = \emptyset$, $k \in [K]$, then with probability larger than $1-c_{1} \exp \left(-c_{2} \log p \right)$, Algorithm \ref{alg:fadatrans} yields a unique solution such that $\hbz = \hbzora$.
\end{theorem}
Theorem \ref{oracle} suggests that under certain incoherence conditions, the F-AdaTrans method (Algorithm \ref{alg:fadatrans}) yields the \textit{sub-oracle} estimator as the unique solution. The incoherence assumption, as outlined in Appendix \ref{mutual_cond:sec}, essentially requires that the non-detectable set $A$ does not exhibit a strong linear correlation with the detectable set. Compared with the oracle estimator (\ref{oracle_def}) which focuses on the true active set $S_0$, the sub-oracle estimator (\ref{sub_oracle_def}) takes the non-detectable set $A$ into estimation, leading to a higher variance and an additional bias term. A more precise initial estimator or a more identifiable transferable structure will lead to a smaller non-detectable set, making the F-AdaTrans estimator more closely approach the oracle estimator.
In the following corollary, we provide a choice of the initial estimator and establish the corresponding statistical rate of the F-AdaTrans estimator. To set up the statement of the corollary, we first introduce the following notation:
$$
\hat{\boldsymbol{\Omega}}_{\tSz, N} = \left[\left(\bX_{\tSz}\right)^{\top} \bX_{\tSz}/N\right]^{-1},\tilde{\boldsymbol{\Omega}}_{\tSz, N} = \left[\left(\tbX_{\tSz}\right)^{\top} \tbX_{\tSz}/N\right]^{-1}, \hat{\boldsymbol{B}} = \sum_{k=1}^{K}\hat{\boldsymbol{B}}_{\tSz, \tSk} \hbzk_{A}.
$$
Here, \(\hat{\boldsymbol{\Omega}}_{\tSz, N}\) and \(\tilde{\boldsymbol{\Omega}}_{\tSz, N}\) denote the estimated precision matrices on the feature set \(\tSz = S_0 \cup A\), computed using the original full sample and the projected sample, respectively. $\hat{\boldsymbol{B}}$ represents an aggregated estimator of the non-transferable source parameters.

\begin{corollary}
\label{partial_rate}
Given the assumptions in Theorem \ref{oracle} (1) and $n_{S} > n_{T} \gtrsim s^2 \log p$, if we run separate Lasso regression on each source sample and the target sample with penalty $\lambda_{\operatorname{Lasso}} \asymp \sqrt{\log p / n_T}$, and construct the initial estimator as
$$
\hbz_{\text{init}} = \hbz_{\text{Lasso}}, \quad \hbdk_{\text{init}} = \hbk_{\text{Lasso}} - \hbz_{\text{Lasso}}, \quad k \in [K],
$$ then by choosing $\lambda_{0} = \lambda_{1} =c_{0}\sqrt{\frac{\log p}{n_{T}}}$ for some constant $c_0$, we have
with probability larger than $1-c_{1} \exp \left(-c_{2} \log p \right)$,
\begin{align}
\label{partial_detectable_bound}
\|\hbz -  \bz\|_{2}^2  \lesssim \tilde{\kappa}_{F}^2\|\hat{\boldsymbol{\Omega}}_{\tSz, N}\|_{\infty}^2 \frac{s\log (s+a)}{N} + \|\tilde{\boldsymbol{\Omega}}_{\tSz, N}\|_{\infty}^2   (1 + \|\hat{\boldsymbol{B}}\|_{\infty})^2 \frac{a\log p}{n_{T}}.
\end{align}
where we define $s = |S_0|$, $a = |A|$,  $ \tilde{\kappa}_{F} := \left\| [\tbX_{\tSz}^\top\tbX_{\tSz}]^{-1}\tbX_{\tSz}^\top \boldsymbol{\epsilon} \right\|_{\infty} / \left\|  [\bX_{\tSz}^\top\bX_{\tSz}]^{-1}\bX_{\tSz}^\top \boldsymbol{\epsilon} \right\|_{\infty}$.

{\small
\begin{itemize}
\item[(1)] If for all $k \in [K]$, $\bXk_{S_k} \perp \bXk_{S_0}$, $\bXk_{(S_0 \cup S_k) \backslash A} \perp \bXk_{A}$, $\bXz_{S_0 \backslash A} \perp \bXz_{A}$, then $\hbz_{(S_0 \cup A)^c} =  \bz_{(S_0 \cup A)^c}$,
\begin{align*}
\|\hbz_{S_0\backslash A} -  \bz_{S_0 \backslash A}\|_{2}^2  \lesssim \|\hat{\boldsymbol{\Omega}}_{S_0 \backslash A, N}\|_{\infty}^2 \frac{s\log s}{N}, \quad \|\hbz_{A} -  \bz_{A}\|_{2}^2  \lesssim \|\hat{\boldsymbol{\Omega}}^{(0)}_{A, n_T}\|_{\infty}^2 \frac{a\log a}{n_T} + \|\hat{\boldsymbol{\Omega}}_{A, n_T}\|_{\infty}^2 \frac{aK\log p}{n_T}.
\end{align*}
\item[(2)] If for all $k \in [K]$, $S_k \supset S_0$,$\bXk_{(S_0 \cup S_k) \backslash A} \perp \bXk_{A}$, $\bXz_{S_0 \backslash A} \perp \bXz_{A}$, then $\hbz_{(S_0 \cup A)^c} =  \bz_{(S_0 \cup A)^c}$,
\begin{align*}
\|\hbz_{S_0\backslash A} -  \bz_{S_0 \backslash A}\|_{2}^2  \lesssim \|\hat{\boldsymbol{\Omega}}^{(0)}_{S_{0} \backslash A, n_T}\|_{\infty}^2 \frac{s\log s}{n_T}, \quad \|\hbz_{A} -  \bz_{A}\|_{2}^2  \lesssim \|\hat{\boldsymbol{\Omega}}^{(0)}_{A, n_T}\|_{\infty}^2 \frac{a\log a}{n_T} + \|\hat{\boldsymbol{\Omega}}_{A, n_T}\|_{\infty}^2 \frac{aK\log p}{n_T}.
\end{align*}
\end{itemize}
}

\end{corollary}

Compared to the rate of the oracle estimator (\ref{ora-rate}), the rate of the F-AdaTrans estimator (\ref{partial_detectable_bound}) involves additional terms related to the non-detectable set. The first term measures the estimation rate with all the detectable transferable structures, whereas the second term accounts for the bias brought by penalizing the non-detectable signals. To better interpret the result, we may consider a special case when the non-detectable set is orthogonal to the detectable set, as illustrated in case (1) and case (2) in the corollary. In both the fully transferable case (\(\bXk_{S_k} \perp \bXk_{S_0}\)) and the fully non-transferable case (\(S_k \supset S_0\)), the F-AdaTrans estimator fully utilizes the detectable transferable structure and achieves a near-oracle rate on the detectable set. For coefficients in the non-detectable set, \(\bz_{A}\), estimation relies solely on the target data, therefore the error rate is constrained by the target sample size \(n_T\). Here, the estimation error comprises two components: the first term reflects the accuracy achievable with only the target dataset, and the second term captures the bias introduced by penalizing the non-detectable set. Unlike \(\|\hat{\boldsymbol{\Omega}}_{S_0 \backslash A, N}\|_{\infty}\) and \(\|\hat{\boldsymbol{\Omega}}^{(0)}_{A, n_T}\|_{\infty}\), which is bounded with high probability, \(\|\hat{\boldsymbol{\Omega}}_{A, n_T}\|_{\infty}\) may diverge when \(N \gg n_T\). Further analysis and potential methods for refining estimation on the non-detectable set are provided in Appendix \ref{app:non-detectable_set}. In the more general case where the non-detectable set is correlated with the detectable set, the estimation rate for the detectable set will also be slower convergence due to the influence of the non-detectable components.


In summary, we demonstrate that \textit{F-AdaTrans}, a transfer learning method based on folded concave penalization, effectively leverages the sparsity of \(\bz\) and the transferable structure of source tasks to estimate \(\bz\) accurately. Appendix \ref{sec:comparison} provides a detailed comparison between \textit{F-AdaTrans} and existing adaptive transfer learning algorithms, focusing on key assumptions, estimation rates, and convergence guarantees. Generally, when source samples exhibit feature-wise transferability, \textit{F-AdaTrans} achieves better performance, particularly in estimating the detectable active features in the target sample. In the ideal scenario where the entire transferable structure is detectable, \textit{F-AdaTrans} attains a strictly faster convergence rate than existing methods. Additionally, unlike approaches that require nonsmooth optimization and rely on approximate solvers, \textit{F-AdaTrans} can be efficiently solved using soft-thresholding, with guaranteed convergence.


\section{Sample-wise Adaptive Transfer Learning}
\label{sample_wise}
The success of the feature-wise adaptive transfer learning relies on the assumption that the target signal is sufficiently strong and the target-source contrast possesses a transferable structure, i.e., $\bdk$ is a sparse vector with $h_k^\wedge$ sufficiently large. 
In applications where the source samples' quality is low or unreliable, the source parameter can deviate from that of the target significantly, yielding dense  target-source contrasts. In this setting, theoretical guarantees of the feature adaptive method~\eqref{obj2} hardly hold due to the hardness of identifying the transferable elements of the sources. To address the problem
and limit the impact of less informative sources, we introduce a sample-wise adaptive transfer learning method that weights the source samples according to their informative level. 

Let $\boldsymbol{w} = [w_0, \ldots, w_K]$ with $w_k \geq 0$ being the weight assigned to the $k$-th source, and recall that $n_k = n_S$ for $k\in [K]$ and $n_k = n_T$ for $k=0$.
We propose to estimate the target parameter $\bz$ 
by solving:
\begin{subequations}\label{P:sample-transfer}
\begin{align}\label{obj}
 & \min_{\bz, \{\bd^{(k)}\}_{k = 1}^K}   \Bigg\{ \frac{1}{N} \sum^{K}_{k=0} w_k\|\byk-\bXk (\bz + \bdk)\|_{2}^{2}+\lambda_{0}  \sqrt{\sum_{k=0}^{K}\frac{n_{k}}{N}w_{k}^2} \|\bz\|_{1} + \lambda_{1}  \sum^{K}_{k=1}w_k\|\bdk\|_{1}  \Bigg\}, \\
    \label{obj_constraint}
  & \text{s.t.} \quad \frac{1}{n_T}\|(\bXz)^\top (\byz - \bXz \bz)\|_{\infty} \le \lambda_{T},
\end{align}
\end{subequations}
where $\lambda_0$, $\lambda_1$ and $\lambda_T$ are nonnegative tuning parameters.

Problem~\eqref{P:sample-transfer} is designed based on the same fused-penalty as~(\ref{obj2}). However, two key differences are made to achieve sample-wise adaptive transfer. First, 
unlike  in problem (\ref{obj2}) where different weights are applied element-wise to $\bz$ and $\bdk$, a single scalar $w_k$  is assigned to both the loss term and the penalty on the parameters associated with the $k$-th source, adjusting its importance in estimating the target parameter. {The weight assigned to $\|\bz\|_{1}$, $\sqrt{\sum_{k=0}^K (n_k/N) w_k^2}$, demonstrates the influence of source weights on the shrinkage of $\bz$.} Second, we introduce  constraint (\ref{obj_constraint})  to identify $\bz$: The left hand side of (\ref{obj_constraint}) corresponds to  the gradient of target loss function and is equal to zero at the ground truth $\bz$ in the noiseless case $\bez_{i} = 0$; the slackness $\lambda_T$ is introduced to account for the presence of noise.
Intuitively, constraint (\ref{obj_constraint}) narrows down the search space for $\bz$ using primary target data, and the objective function (\ref{obj}) proceeds to use the full sample, improving the estimation accuracy. 

We note that the solution of problem~\eqref{P:sample-transfer} is invariant under positive  scaling of the $w_k$'s. To remove such ambiguity, we restrict the weights to satisfy the normalization constraint 
\begin{align}\label{eq:weight-normalize}
 \sum^{K}_{k=0}\frac{n_k}{N} w_k  =  1, \ w_k \geq 0, \ k = 0, \ldots,K.
\end{align}
To better understand the formulation (\ref{P:sample-transfer}), we examine two  cases.  If we set the target weight $w_{0} = N/n_T$ and the source weight $w_k = 0$ for all $k \in [K]$,  problem (\ref{P:sample-transfer}) reduces to a single-task lasso regression problem where only the target data is used to estimate $\bz$. In this case, there is no information transfer from the source samples and should be chosen when the $\bdk$'s are large. On the other hand, if we set all $w_k = 1$  for all $k = 0, \ldots, K$, then equal importance is assigned to all samples in estimating $\bz$. Such weight assignment is ideal if all the $\bdk$'s are zero. 
By adjusting the $w_k$'s, we interpolate the two extremes and aim to achieve the optimal balance between the source and target samples.


\subsection{Weight Choice with Known  Informative Level}
\label{sample_known}

We first discuss the choice of $w_k$ with a known informative level, characterized by  the following parameter space:
\begin{align}\label{def:parameter_space}
{
\Theta_{2}:=\left\{\bz, \{\bdk\}_{k=1}^K:  \|\bz\|_0 \leq s, \|\bdk\|_1 \leq h_{k}, k \in [K]\right\}}.
\end{align}
The target parameter $\bz$ is still assumed to be $s$-sparse.  However, the informative level of the $k$-th source is quantified by the $\ell_1$ norm of the whole target-source contrast $\bdk$, allowing for dense $\bdk$s with small elements.
 A larger $h_k$ indicates that the $k$-th source is less similar to the target.

 The following result establishes the estimation error of $\hbz$ assuming both $s$ and $h_k$'s are known.

\begin{theorem}
\label{onesteprate}
Under Assumption \ref{A1} and \ref{A2}, assume $n_{S} > n_{T}$, $\frac{s\log p}{n_{T}} + K\bar{h}(\boldsymbol{w})\sqrt{\frac{\log p}{n_T}} = o(1)$ and the weights satisfy (\ref{eq:weight-normalize}), with each weight either satisfying \( w_k \geq \underline{w} \) for some small constant \(\underline{w} > 0\) or being zero. If we choose 
{\small\begin{align*}
   &\lambda_{0} = c_{0} \left[ \Bigg( \sum_{k=0}^{K}\frac{n_{k}}{N}w_{k}^2\Bigg)^{-\frac{1}{2}} \Bigg(\frac{\bar{h}(\boldsymbol{w})^2 \log p}{s^2 n_{T}} \Bigg)^{\frac{1}{4}}\!\! + \Bigg(\frac{\log p}{N}\Bigg)^{\frac{1}{2}} \right], \\
   &\lambda_{1} = c_{0} \frac{n_{S}}{N} \Bigg(\frac{\log p}{n_{T}} \Bigg)^{\frac{1}{2}},  \text{ and }\lambda_T = c_{1} \Bigg(\frac{\log p}{n_{T}} \Bigg)^{\frac{1}{2}},
\end{align*}}
for some appropriate constants $c_{0}$ and $c_{1}$, then the solution of S-AdaTrans problem (\ref{P:sample-transfer}) satisfies
\begin{align}
\label{onestepFusion}
\hspace{-0.3cm}\|\hbz - \bz\|_2^2 \lesssim  \kappa_{S}(\boldsymbol{w}) \frac{s \log p}{N} + c_{\bSig}\bar{h}(\boldsymbol{w})\sqrt{\frac{\log p}{n_{T}}},
\end{align}
 with probability larger than $1-c_{1} \exp \left(-c_{2} \log p \right)$,
 where 
  \begin{align}
     \kappa_{S}(\boldsymbol{w}):= 
     \sum_{k=0}^{K}\frac{n_{k}w_{k}^2}{N}, \quad 
      \bar{h}(\boldsymbol{w}): = \sum^{K}_{k=1}\frac{n_k w_k}{N} h_k,
 \end{align}
 and $c_{\bSig}$ is a universal constant.
\end{theorem}

 Term $\bar{h}(\boldsymbol{w})$  is the weighted non-transferable level and $c_{\bSig}$ is a universal constant depending on the largest and smallest eigenvalues of $\{\bSig^{(k)}\}_{k=0}^{K}$. To interpret $\kappa_{S}(\boldsymbol{w})$, we notice that under the normalization constraint~\eqref{eq:weight-normalize}, by solving the KKT system we can conclude $\kappa_{S}(\boldsymbol{w})$ attains minimum value $1$ with the corresponding $w_k = 1$ for $k = 0, \ldots, K$. In such a case, the first term is $s \log p / N$ and corresponds to a near minimax optimal rate using  $N$ target samples~\citep{raskutti2011minimax} . 
 Therefore, in the same way as $\kappa_{F}$ defined in~\eqref{kappa}, $\kappa_{S}(\boldsymbol{w})$ can be viewed as a metric quantifying the transferability of the source datasets. 

To further understand the result, we revisit the two extreme cases  previously discussed. 
Under the weight choice $w_{0} = N/n_T$ and $w_k = 0$ for all $k \in [K]$, problem (\ref{P:sample-transfer}) reduces to the single-task lasso regression problem and ~\eqref{onestepFusion} correspondingly degenerates to the near minimax optimal bound $s \log p / n_T$. When  
 setting $w_{k} =1$, on the other hand,~\eqref{onestepFusion} becomes  $s \log p / N + \tilde{h} \sqrt{\log p/n_{T}}$ with $\tilde{h} := \frac{1}{K}\sum^{K}_{k=1} h_k$ being the  average of all $h_k$s. In this case, the first term of~\eqref{onestepFusion} achieves its minimum, at the cost of introducing a nonzero second term accounting for the negative impact of non-transferable target-source contrasts $\bdk$.
Such a rate is again near minimax optimal for multi-source sparse regression problems under sufficiently small $\tilde{h}$ and large $n_T$ \citep{li2022transfer}. 

With a flexible choice of the weights $w_k$'s, the upper bound (\ref{onestepFusion}) represents a middle ground between the two extremes. As $\kappa_{S}(\boldsymbol{w}) \geq 1$, the first term, $\kappa_{S}(\boldsymbol{w}) s \log p/N$, reflects a loss in the effective sample size when the source samples are not fully utilized. 
 The second term, $\bar{h}(\boldsymbol{w})\sqrt{\log p/n_{T}}$, also depending on the weight $w_k$, shows the potential reduction of negative transfer if $h_k$ is large but the corresponding source sample is down-weighted.
Consequently, the weights should be carefully chosen to balance the tradeoff between these two terms  to achieve the optimal estimation rate. 

To gain theoretical insights, we investigate the impact of the sample size $n_S$, $n_T$, and the informative level $h_k$ on the optimal weights that minimize the bound (\ref{onestepFusion}). {To better understand the problem, we introduce the sample-adjusted weight $w_0^\prime = \frac{n_T}{N}w_0$, $w_k^\prime = \frac{n_S}{N}w_k$ and the simplex $\boldsymbol{\Delta}^{K} = \{\boldsymbol{w}^{\prime} \in \mathbb{R}^{(K+1)}: \boldsymbol{1}^\top \bw^{\prime} = 1 \ \text{and} \ \bw^{\prime} \geq \boldsymbol{0}\}$.
With these definitions, we can reformulate the optimization problem as follows:}
\begin{align}
\label{prob:reformulated_w}
\min_{\boldsymbol{w}^{\prime} \in \boldsymbol{\Delta}^K}~\left\{Q(\boldsymbol{w}^\prime) := 
 \frac{s \log p}{n_T}(w_0^\prime)^2 +  \sum_{k=1}^K \left[\frac{s \log p}{n_{S}}(w_{k}^{\prime})^2 + c_{\bSig} h_{k}\sqrt{\frac{\log p}{n_T}}w_k^{\prime}\right]  \right\},
\end{align}
{which is  minimizing 
a quadratic objective function over a simplex constraint \citep{duchi2008efficient}. The following corollary provides the expression of the optimal weights for \textit{S-AdaTrans}.}

\begin{corollary}
\label{choice_of_w}
Let $h_{(j)}$ denote the $j$-th smallest  $h_{k}$ for $k \in [K]$ with some $K \geq 1$.
The  weights under constraint~\eqref{eq:weight-normalize} that minimizes the estimation error bound (\ref{onestepFusion}) are $w_0 = (N/n_T) w_0'$ and $w_k = (N/n_S) w_k'$, with
\begin{align}
\label{weight:generalK}
     &w_{k}^{\prime}=\max \left\{\frac{n_{S}}{n_{T}+\rho n_{S}}\left(1+\sum_{j=1}^{\rho} \frac{h_{(j)}}{2} \frac{c_{\bSig}\sqrt{\log p / n_{T}}}{s \log p / n_{S}}\right)-\frac{h_{k}}{2} \frac{c_{\bSig}\sqrt{\log p / n_T}}{{s \log p/n_{S}}}, 0\right\}, \\
     &w_{0}^{\prime}=\frac{n_{T}}{n_{T}+\rho n_{S}}\left(1+\sum_{j=1}^{\rho} \frac{h_{(j)}}{2} \frac{c_{\bSig}\sqrt{\log p / n_{T}}}{s \log p / n_{S}}\right), \label{weight:w0prime}
\end{align}
where $\rho$ is the number of strictly positive $w_k$s for $k \in [K]$.  As a special case, when $K = 1$, we have 
\begin{align}
\label{weight:K=1}
{w}_{1}^{\prime} = \max \left\{
 \frac{n_S}{n_T+n_S} - \frac{n_T}{n_T+n_S}\frac{h_1}{2}\frac{c_{\bSig} \sqrt{\log p / n_T}}{s \log p / n_S}, 0\right\},  \ w_{0}^{\prime} = 1 - w_{1}^{\prime},
\end{align}
and under the assumptions of Theorem \ref{onesteprate}, the solution of (\ref{P:sample-transfer}) with such a choice of $w_k$'s satisfies
\begin{align}
\label{onesteporacleFusion}
\hspace{-0.3cm}\|\hbz - \bz\|_2^2 \lesssim  \frac{s \log p}{N} + c_{\bSig}h_{1}\sqrt{\frac{\log p}{n_{T}}} \wedge \frac{s \log p}{n_T},
\end{align}
 with probability larger than $1-c_{1} \exp \left(-c_{2} \log p \right)$.
\end{corollary}

To understand the results in Corollary \ref{choice_of_w}, we start with the special case when $K=1$. According to (\ref{weight:K=1}), when $h_1 = 0$, we have $w_1^\prime = n_S/N$ and $w_0^\prime = n_T/N$, which implies that $w_0 = w_1 = 1$, corroborating the fact that source and target samples should be equally weighted when sharing the same ground truth parameter. As $h_1$ increases, 
the weight $w_1$ assigned to the source sample decreases, indicating we should down-weight the source when it becomes less informative. Considering the impact of the sample size, we see that the normalized weight $w_1'$ decreases either when $n_S$ decreases or  $n_T$ increases. The implication is intuitive, showing that under these two cases, the target becomes relatively more reliable and a larger weight should be allocated. With the optimally selected weights, (\ref{onesteporacleFusion}) implies that $\hbz$ is always no worse than the Lasso benchmark, and achieves a faster rate when the source is informative, i.e., $h_1$ is small. Furthermore, when $\sqrt{\log p / n_T} \lesssim h_1$ holds (\ref{onesteporacleFusion}) recovers the minimax optimal rate for estimating $\bz$ under the parameter space (\ref{def:parameter_space})  \citep{li2022transfer}.

\begin{figure}
    \centering
    \includegraphics[width=\linewidth]{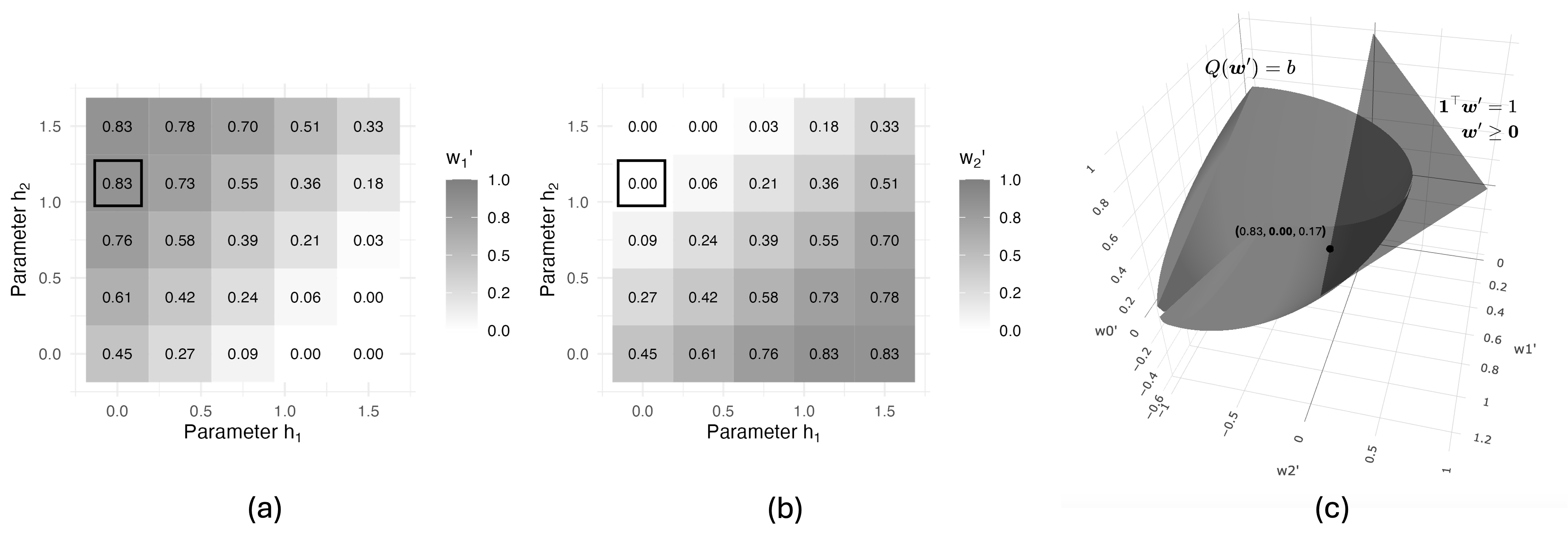}
    \caption{(a,b) the influence of varying $h_1$ and $h_2$ on the weights $w^{\prime}_{1}$ and $w^{\prime}_{2}$ that minimize the bound (\ref{onestepFusion}) under constraints ($K=2$, $p=500$, $n_T=50$, $n_S=250$, $s=8$). (c) the geometric interpretation when solving the problem with $h_1=0$ and $h_2=1$, showing how the ellipsoid level set hits the ``edges" of the constraint and shrinks $w_2^\prime$ to zero.}
    \label{fig:3D}
\end{figure}

For the general case when $K \geq 1$, the optimal choice of $w_k$ not only depends on $h_k$, the informative level of the $k$th source to the target but is also influenced by the relative informative levels among sources.  
Figure \ref{fig:3D}(a) and (b) provide an illustration for $K=2$. According to (\ref{weight:generalK}), if we keep $h_k$ fixed and increase $\{h_j\}_{j=1,j\neq k}^K$, $w_k$ will increase, resulting in a higher weight for the $k$th source and lower weights for other sources. On the other hand, if $h_k$ increases while $\{h_j\}_{j=1,j\neq k}^K$ are held constant, the weight $w_k$ decreases, eventually shrunk to $0$. {In fact, we can show that $w_k$ is Lipschitz continuous in $h_k$. See Appendix \ref{weight_change} for a more detailed explanation.} Figure \ref{fig:3D}(c) provides a geometric illustration, showing how the level set of the objective function, which is  an ellipsoid, is  influenced by the $h_k$s.



We finish by discussing the assumptions of Theorem \ref{onesteprate}. The condition $s\log p / n_T= o(1)$ shows that the dimension $p$ can grow up to an exponential order with respect to the target sample size. Such a sample-dimension dependency is standard in the high-dimensional setting.
The requirement that $K\bar{h}(\boldsymbol{w})\sqrt{\log p/n_T}$ is bounded can be fulfilled if either $\bar{h}(\boldsymbol{w})$ diminishes or $n_T$ grows  with the task number $K$. The lowerbound on $w_k$ prevents the weights to be $o(1)$ as $p$ grows. In practice it can be set as a small numerical constant.
These two requirements are technical conditions imposed to deal with the higher order terms that arise due to the non-strong convexity and smoothness of the empirical loss, \textit{challenges unique to the high dimensional setting} (cf. Lemma A.5 and Section 4.1). Notably, our bound does not depend on the whole covariance matrices $\Sigk$'s but only their eigenvalue bounds,
demonstrating the robustness of our method to covariate shifts. At the technical level, the proof of the theorem involves a  non-asymptotic analysis of the coupled structure of $\bz$ and $\bdk$ in problem (\ref{P:sample-transfer}), together with a fine treatment of both the non-strong convexity and smoothness,  which may be of independent interest.

\subsection{Weight Choice with Unknown  Informative Level}
\label{sec:sample_unknown_weights}
Obtaining optimal weights by minimizing the error bound in~\eqref{onestepFusion} requires knowing the sparsity level $s$ of $\bz$, the informative level $h_k$, and the universal constant $c_{\bSig}$. In this section, we provide a data-driven approach to compute the weights by solving the following optimization problem:  \begin{align} 
     & \min_{\boldsymbol{w}^\prime} \left\{
\sum_{k=0}^K \frac{\|\hbz_{\text{init}}\|_{0} \log p}{n_k}(w_k^{\prime})^{2} + \lambda_{W} \sum_{k=1}^K \|\hat{\bd}^{(k)}_{\text{init}}\|_{1} \sqrt{\frac{\log p}{n_T}}w_k^{\prime}  \right\}\nonumber\\  
& \textrm{\ s.t.}  \ \sum^{K}_{k=0}w_k^{\prime} = 1 \text{ and } w_k^{\prime} \geq 0, \ k=0, 1,\dots, K.\label{choose_weight}
\end{align}

In~\eqref{choose_weight}, we estimate $s$ and $h_k$ respectively by
$\|\hbz_{\text{init}}\|_{0}$ and $\|\hat{\bd}^{(k)}_{\text{init}}\|_{1}$ using 
some initial estimators, such as the Lasso estimator. The unknown constant $c_{\bSig}$ is 
handled as a hyperparameter $\lambda_{W}$, whose optimal choice is found through cross-validation on the target dataset. The optimization problem (\ref{choose_weight}) is convex and can be solved efficiently~\citep{duchi2008efficient}. With the weight computed by~\eqref{choose_weight}, we can then proceed to solve  problem (\ref{P:sample-transfer}) to obtain an estimator for $\bz$. Note that although problem (\ref{P:sample-transfer}) involves three hyperparameters, namely $\lambda_{0}$, $\lambda_{1}$, and $\lambda_{T}$, the relative magnitudes of $\lambda_{0}$ and $\lambda_{1}$ are fixed and thus can be jointly tuned. Algorithm \ref{alg:sadatrans} outlines the proposed \textit{S-AdaTrans} algorithm.

\begin{algorithm}
\caption{\textit{S-AdaTrans}}
\label{alg:sadatrans}
\begin{algorithmic}[1]
\State \textbf{Input:} target sample $\left(\bXz, \byz\right)$, source sample $\left\{\left(\bXk, \byk\right)\right\}_{k=1}^K$, tuning parameters $\lz$, $\lk$, $\lambda_{T}$ {(tuned jointly)} and $\lambda_{W}$.
\State \textbf{Output:} the estimated target coefficient $\hat{\boldsymbol{\beta}}^{(0)}$.
\State \textit{Weight Construction:} Run an initialization algorithm (e.g. separate Lasso regression) to obtain the initial estimator $\hbz_{\text{init}}$ and $\hbdk_{\text{init}}$, construct the sample-wise adaptive weight $\hat{w}^{(k)}$ by solving problem (\ref{choose_weight}).

\State \textit{Sample-wise Adaptive Transfer:} Compute $\hbz$ by solving
\Statex 
$\hbz, \{\hbdk\}_{k=1}^K \in \min_{\bz, \{\bd^{(k)}\}_{k = 1}^K}   \Bigg\{ \frac{1}{N} \sum^{K}_{k=0} \hat{w}_k\|\byk-\bXk (\bz + \bdk)\|_{2}^{2}+\lambda_{0}  \sqrt{\sum_{k=0}^{K}\frac{n_{k}}{N}\hat{w}_{k}^2} \|\bz\|_{1} + \lambda_{1}  \sum^{K}_{k=1}\hat{w}_k\|\bdk\|_{1}  \Bigg\}$,
\Statex $\text{s.t.} \quad \frac{1}{n_T}\|(\bXz)^\top (\byz - \bXz \bz)\|_{\infty} \le \lambda_{T}.$
\end{algorithmic}
\end{algorithm}


\section{Emperical Experiments}
\subsection{Simulation Study}

\label{simulation}
We evaluate the empirical performance of our proposed methods, \textit{F-AdaTrans}  and \textit{S-AdaTrans}, and compare with existing high-dimensional transfer learning methods \textit{TransGLM} \citep{tian2022transfer}, \textit{TransLasso} \citep{li2022transfer} and \textit{TransHDGLM} \citep{li2023estimation}. 
Specifically, the following methods are implemented:\\[1ex]
\textbf{Baseline:} \textit{Lasso (baseline)}: LASSO regression fitted on the target sample;\\[1ex]
\textbf{Feature-wise transfer learning methods:} {\textit{Ora-Est}}: oracle estimator with known transferable structure (\ref{oracle_form}); \textit{Ora-F-Ada} : \textit{F-AdaTrans} with known transferable structure (Section \ref{feature_known}); \textit{F-AdaTrans}: \textit{F-AdaTrans} with data-driven weights selection (Section~\ref{sec:feature_unknown_weights});\\
\textbf{Sample-wise transfer learning methods:} \textit{Ora-S-Ada}: \textit{S-AdaTrans} with known source informative level (Section~\ref{sample_known}); \textit{S-AdaTrans}: \textit{S-AdaTrans} with data-driven weights selection (Section~\ref{sec:sample_unknown_weights});
\\[1ex]
\textbf{Benchmark methods:} \textit{TransGLM} \citep{tian2022transfer}: a transfer learning algorithm with transferable source detection, each source is either included or discarded based on some decision rule;  \textit{TransLasso} \citep{li2022transfer}: a two-step transfer learning algorithm with source aggregation, a collection of estimators are constructed and aggregated together; \textit{TransHDGLM} \citep{li2023estimation}: a transfer learning algorithm with aggregation, the obtained estimator is combined with \textit{Lasso (baseline)} to prevent negative transfer.
\\[1ex] 
 Each simulation set is replicated with 100 independent trials with fixed random seeds for each, and we report the average estimation error of the target parameter, i.e., $\|\hbz - \bz\|_2^2$. Implementation details can be found in Appendix \ref{app:implement}.

\noindent\textbf{Simulation settings.} We consider a high-dimensional linear regression problem with dimension $p = 500$ and sparsity level $s = |S_0| = 8$. The target coefficient is set as $\bz_{j} = 0.3$ for $1 \le j \le s$ and $\bz_{j} = 0$ otherwise.
We generate $n_{T} = 50$ independent target samples $(\bXz, \byz)$  by $\byz = \bXz \bz + \bez$ with  $\bXz_{i \cdot} \sim N(0, \boldsymbol{I})$ and each $\bez_{i} \sim N(0, 1)$. 
Each $k$-th source sample $(\bXk, \byk)$ is generated  according to the model $\byk = \bXk (\bz+\bdk) + \bez$ with $\bXk_{i \cdot}\sim N(0, \Sigk)$ and $\bez_{i} \sim N(0, 1)$. We set $\Sigk = (\boldsymbol{A}^{(k)})^\top (\boldsymbol{A}^{(k)}) + \boldsymbol{I}$. Here $\boldsymbol{A}^{(k)}$ is a random matrix with each entry equals 0.3 with probability 0.3 and equals 0 with probability 0.7. We set $n_S = 250$ and $K = 4$ if not otherwise specified.


We consider the following two parameter configurations for the target-source contrast $\bdk$, corresponding respectively to feature- and sample-adaptive transfer learning scenarios. Let $\mathcal{D}(h^{\wedge}) := N(h^{\wedge}, (h^{\wedge}/3)^2)$ be a normal distribution parameterized by the non-transferable signal strength $h^\wedge$. 

\textit{Setting 1: feature adaptive transfer.} In this setting we generate two types of source samples with non-overlapping transferable features. Note that we define $s = |S_0|$ as the sparsity measure of $\bz$ and $s_k = |S_k|$ as the sparsity measure of $\bdk$. For $k=1,\dots, K$, if $k$ is odd then we set $\bdk_{j} \sim \mathcal{D}(h^{\wedge})$ for \textit{$1 \le j \le s/2$} and $\bdk_{j}=0$ otherwise. On the other hand if  $k$ is even, we set $\bdk_{j} \sim \mathcal{D}(h^{\wedge})$ for \textit{$s/2 < j \le s_k$} and $\bdk_{j}=0$ otherwise. We set $h^{\wedge} = 0.6$ and $s_k = 25$ if not otherwise specified. 
Figure~\ref{fig:parameter-settings} (a) illustrates the nonzero patterns of the target parameter $\bz$ and the $\bdk$'s.
In this setting, the odd-indexed sources have non-transferable features in the first half of the features of interest, while even-indexed sources have them in the second half.

\textit{Setting 2: sample adaptive transfer.} In this setting we generate two types of sources corrupted by dense non-transferable signals.
For odd $k$, we generate $\bdk_{j} \sim \mathcal{D}(h^{\wedge}/10)$ for \textit{$1 \le j \le s_k$} and $\bdk_{j}=0$ otherwise. For even $k$, we set $\bdk_{j} \sim \mathcal{D}(h^{\wedge})$ for \textit{$1 \le j \le s_k$} and $\bdk_{j}=0$ otherwise. We set $h^{\wedge} = 0.024$ and $s_k = 450$. Under this setting, the odd-indexed sources are more informative, while even-indexed ones are less, as shown in Figure~\ref{fig:parameter-settings} (b).

We investigate the impact of four key factors on the performance of \textit{F-AdaTrans} in Setting 1 and that of \textit{S-AdaTrans} in Setting 2, namely, the (i) non-transferable signal strength $h^{\wedge}$, (ii) source sample size $n_S$, (iii) number of source tasks $K$,  and (iv) number of non-transferable features $s_k$, reported in Table \ref{tab:feature}, \ref{tab:sample} and Figure \ref{fig:skandK} in the Appendix \ref{app:sim-suppl}.

\begin{table}[htbp]
  \centering
 \caption{Average estimation error under feature-wise adaptive transfer setting (Setting 1)}
\resizebox{0.75\columnwidth}{!}{%
    \begin{tabular}{rrrrrrrrrr}
    \toprule
    \multicolumn{1}{l}{$n_S$} & \multicolumn{1}{l}{$h^{\wedge}$} & \multicolumn{8}{c}{Method} \\
\cmidrule{1-10}          &       & \multicolumn{1}{l}{\textit{F-AdaTrans}} & \multicolumn{1}{l}{\textit{Ora-F-Ada}} & \multicolumn{1}{l}{\textit{Ora-Est}} & \multicolumn{1}{l}{\textit{S-Ada}} & \multicolumn{1}{l}{\textit{TransGLM}} & \multicolumn{1}{l}{\textit{TransLasso}} & \multicolumn{1}{l}{\textit{TransHDGLM}} & \multicolumn{1}{l}{\textit{Lasso (baseline)}} \\
\cmidrule{3-10}    150   & 0     & 0.00674 & 0.00139 & 0.00132 & 0.00746 & 0.07211 & 0.06111 & 0.45137 & 0.72592 \\
    150   & 0.3   & 0.05099 & 0.00289 & 0.00276 & 0.19843 & 0.37277 & 0.24010 & 0.21433 & 0.72592 \\
    150   & 0.6     & 0.03408 & 0.00283 & 0.00276 & 0.66213 & 0.80447 & 0.50966 & 0.55960 & 0.72592 \\
    150   & 0.9   & 0.03788 & 0.00286 & 0.00276 & 1.07917 & 0.87083 & 0.65184 & 0.67559 & 0.72592 \\
    \hline
    250   & 0     & 0.00374 & 0.00070 & 0.00069 & 0.00423 & 0.06325 & 0.04767 & 0.07370 & 0.72592 \\
    250   & 0.3   & 0.05555 & 0.00161 & 0.00159 & 0.20870 & 0.35686 & 0.24159 & 0.16099 & 0.72592 \\
    250   & 0.6     & 0.02272 & 0.00163 & 0.00159 & 0.39472 & 0.77665 & 0.50321 & 0.37737 & 0.72592 \\
    250   & 0.9   & 0.02190 & 0.00164 & 0.00159 & 0.12527 & 0.84094 & 0.62589 & 0.48945 & 0.72592 \\
    \hline
    350   & 0     & 0.00258 & 0.00053 & 0.00051 & 0.00315 & 0.07240 & 0.04978 & 0.03929 & 0.72592 \\
    350   & 0.3   & 0.06441 & 0.00117 & 0.00115 & 0.22955 & 0.40010 & 0.28158 & 0.18873 & 0.72592 \\
    350   & 0.6     & 0.01923 & 0.00117 & 0.00115 & 0.18779 & 0.78159 & 0.51945 & 0.41611 & 0.72592 \\
    350   & 0.9   & 0.01614 & 0.00119 & 0.00115 & 0.10227 & 0.82717 & 0.63348 & 0.51272 & 0.72592 \\
    \hline
    450   & 0     & 0.00196 & 0.00041 & 0.00039 & 0.00240 & 0.06440 & 0.04655 & 0.03516 & 0.72592 \\
    450   & 0.3   & 0.07132 & 0.00093 & 0.00091 & 0.20874 & 0.40389 & 0.27455 & 0.22851 & 0.72592 \\
    450   & 0.6     & 0.01618 & 0.00093 & 0.00091 & 0.12560 & 0.79897 & 0.55738 & 0.42812 & 0.72592 \\
    450   & 0.9   & 0.01455 & 0.00095 & 0.00091 & 0.11282 & 0.83521 & 0.62163 & 0.51907 & 0.72592 \\
    \bottomrule
    \end{tabular}%
    }
  \label{tab:feature}%
\end{table}%

\textbf{F-AdaTrans (Setting 1)}. 
(i) Table \ref{tab:feature} illustrates as the strength of non-transferable signals $h^{\wedge}$ increases, the estimation error of \textit{F-AdaTrans} slightly increases then decreases. This trend demonstrates the method's increased capability to detect the transferable structure with increasing signal strength. Once some non-transferable features are detected, they are subsequently filtered out, and thereby preventing further negative transfer. This finding is further supported by Figure \ref{fig:FAdaFusion-weight} in Appendix \ref{app:sim-suppl}, showing the detection accuracy of the non-transferable features increases as $h^\wedge$ increases. In contrast, existing methods show a rapidly increasing error, due to the fact that it can only either include or completely discard a source sample, but fails to transfer useful features within the source sample. 
(ii) Table \ref{tab:feature} shows the estimation error of \textit{F-AdaTrans} decreases rapidly with an increase in source sample size $n_S$, while other methods demonstrate minimal or no improvement. In addition, the \textit{Ora-F-Ada}, where weights are selected based on the true transferable structure, consistently demonstrates performance comparable to that of the Oracle Estimator, which validates our theoretical results in Section 3.1. (iii) Figure~\ref{fig:skandK} (top left) in the Appendix \ref{app:sim-suppl} shows as the task number $K$ increases, \textit{F-AdaTrans} achieves a smaller estimation error.
(iv) Figure~\ref{fig:skandK} (top right) shows  as $s_k$ increases, the error of \textit{F-AdaTrans} increases since the initial estimator gets worse as the cardinality of $\bdk$ grows, making it harder to identify the transferable structure in the source. However, it still achieves a better performance than the benchmark methods due to the partial inclusion of transferable features.
{Notice that even under the feature-wise transfer setting, the \textit{S-AdaTrans}   outperforms most other state-of-the-art methods.
This is because \textit{S-AdaTrans} also adopts the fused-penalty design, which is also able to capture the feature-wise transferable structure to some extent.}

\begin{table}[htbp]
  \centering
\caption{Average estimation error under sample-wise adaptive transfer setting (Setting 2)}
\resizebox{0.75\columnwidth}{!}{%
    \begin{tabular}{rrrrrrrrr}
    \toprule
    \multicolumn{1}{l}{$n_S$} & \multicolumn{1}{l}{$h^{\wedge}$} & \multicolumn{7}{c}{Method} \\
    \midrule
          &       & \multicolumn{1}{l}{\textit{S-AdaTrans}} & \multicolumn{1}{l}{Ora-S-Ada} & \multicolumn{1}{l}{\textit{F-Ada}} & \multicolumn{1}{l}{\textit{TransGLM}} & \multicolumn{1}{l}{\textit{TransLasso}} & \multicolumn{1}{l}{\textit{TransHDGLM}} & \multicolumn{1}{l}{\textit{Lasso (baseline)}} \\
\cmidrule{3-9}    150   & 0     & 0.00746 & 0.00673 & 0.00674 & 0.07211 & 0.06111 & 0.45137 & 0.72592 \\
    150   & 0.012   & 0.03638 & 0.02017 & 0.02128 & 0.35591 & 0.09798 & 0.36087 & 0.72592 \\
    150   & 0.024   & 0.05383 & 0.02939 & 0.04419 & 0.68311 & 0.08636 & 0.68557 & 0.72592 \\
    150   & 0.036   & 0.05763 & 0.04173 & 0.08163 & 0.31688 & 0.07959 & 0.71876 & 0.72592 \\
    \hline
    250   & 0     & 0.00423 & 0.00373 & 0.00374 & 0.06325 & 0.04767 & 0.07370 & 0.72592 \\
    250   & 0.012   & 0.01108 & 0.01203 & 0.01589 & 0.27767 & 0.06877 & 0.06915 & 0.72592 \\
    250   & 0.024   & 0.01536 & 0.01411 & 0.03601 & 0.82116 & 0.06644 & 0.38825 & 0.72592 \\
    250   & 0.036   & 0.03251 & 0.02777 & 0.06791 & 0.73847 & 0.06032 & 0.61596 & 0.72592 \\
    \hline
    350   & 0     & 0.00315 & 0.00257 & 0.00258 & 0.07240 & 0.04978 & 0.03929 & 0.72592 \\
    350   & 0.012   & 0.00827 & 0.00618 & 0.01453 & 0.25320 & 0.06946 & 0.06489 & 0.72592 \\
    350   & 0.024   & 0.01262 & 0.00828 & 0.03489 & 0.76618 & 0.06041 & 0.36792 & 0.72592 \\
    350   & 0.036   & 0.02396 & 0.01205 & 0.06586 & 0.91631 & 0.05411 & 0.59985 & 0.72592 \\
    \hline
    450   & 0     & 0.00240 & 0.00196 & 0.00196 & 0.06440 & 0.04655 & 0.03516 & 0.72592 \\
    450   & 0.012   & 0.00522 & 0.00456 & 0.01437 & 0.23055 & 0.06793 & 0.07968 & 0.72592 \\
    450   & 0.024   & 0.00907 & 0.00559 & 0.03582 & 0.73963 & 0.07062 & 0.37260 & 0.72592 \\
    450   & 0.036   & 0.00914 & 0.00685 & 0.06847 & 1.07909 & 0.05298 & 0.57727 & 0.72592 \\
    \bottomrule
    \end{tabular}%
    }
  \label{tab:sample}%
\end{table}%

\textbf{S-AdaTrans (Setting 2)}. 
(i) Table~\ref{tab:sample} shows  \textit{S-AdaTrans} is robust to increasing $h^\wedge$. In contrast, \textit{TransGLM} shows significant instability in the change of the estimation error for different choices of $n_S$: for $n_S = 150$ and $250$, the error first increase then decrease, while for $n_S = 350$ and $450$, the error monotonically increases.  This is primarily because \textit{TransGLM} adopts a binary decision choice. 
\textit{S-AdaTrans}, on the other hand, is able to adjust the importance of the source samples by assigning continuous values of weights, and thus more stable with respect to the change of source informative level. See Figure \ref{fig:SAdaFusion-weight} in Appendix \ref{app:sim-suppl} for the numerical evidence.
(ii) Similar to Setting 1, Table~\ref{tab:sample} shows  in Setting 2  the performance of \textit{S-AdaTrans} improves rapidly as the source sample size $n_S$ increases. 
(iii) Figure~\ref{fig:skandK} (bottom left) in the Appendix \ref{app:sim-suppl} shows the error of \textit{S-AdaTrans} decreases with $K$ increases. Further, it gets closer to \textit{Ora-S-Ada} with known source informative levels since the initial estimators become more accurate. Additionally, \textit{F-AdaTrans} demonstrates comparable performance to \textit{S-AdaTrans} in Setting 2 and outperforms other methods in most cases. (iv) The impact of increasing $s_k$ is similar to increasing $h^\wedge$, as shown in Figure~\ref{fig:skandK} (bottom right). This is because in Setting 2, increasing $s_k$ essentially implies an increase in $h_k$ [cf.~\eqref{def:parameter_space}]. Note that under the sample-wise adaptive transfer setting, \textit{TransLasso} shows fairly good performance compared to other benchmark methods, although under-performs \textit{S-AdaTrans}. This is because \textit{TransLasso}'s aggregation step also incorporates a weighting strategy to integrate multiple estimators. 

\subsection{Real Data Analysis}\label{app:real-data}

In the real data analysis, we examine the stock recovery of S\&P 500 companies during the COVID-19 pandemic \citep{guo2023statistical}, applying \textit{F-AdaTrans} and \textit{S-AdaTrans} to assess \textit{the impact financial metrics to the stock recovery capability from such adverse events}. 

After the outbreak of the COVID-19 pandemic, the U.S. stock market exhibited a unique V-shape pattern, due to the sudden nationwide lockdown and subsequent government and federal reserve interventions. It has been proved that the stock recovery capability can be predicted by various financial metrics of the corresponding companies \citep{guo2023statistical}. These companies, categorized into different \textit{sectors} by the Global Industry Classification Standard (GICS), show their different recovery capabilities towards adverse events likewise. 

To this end, the response variable $\by$ is the stock return measured from each of the S\&P 500 company's peak pre-pandemic prices in February 2020 to Augest 30, 2020. We consider the 50 over 200 moving average ratio on April 30, 2020 and $559$ accounting metrics from financial statements as potential features $\bX$, resulting in a feature dimension $p=560$. The accounting metrics, sourced from Yahoo Finance as of April 30, 2020, are derived from the most recent annual reports (fiscal years 2015-2019) and the first three quarters of 2019. We focus on the latest data and average growth rates of each metric. Our selection criteria included companies that released their latest financial reports between February and April 2020. This yielded a total sample of $N=455$ companies across the 9 sectors, with sector-wise sample sizes $n_k$ ranging from $24$ to $76$.

\begin{figure}
    \centering
    \includegraphics[width=\linewidth]{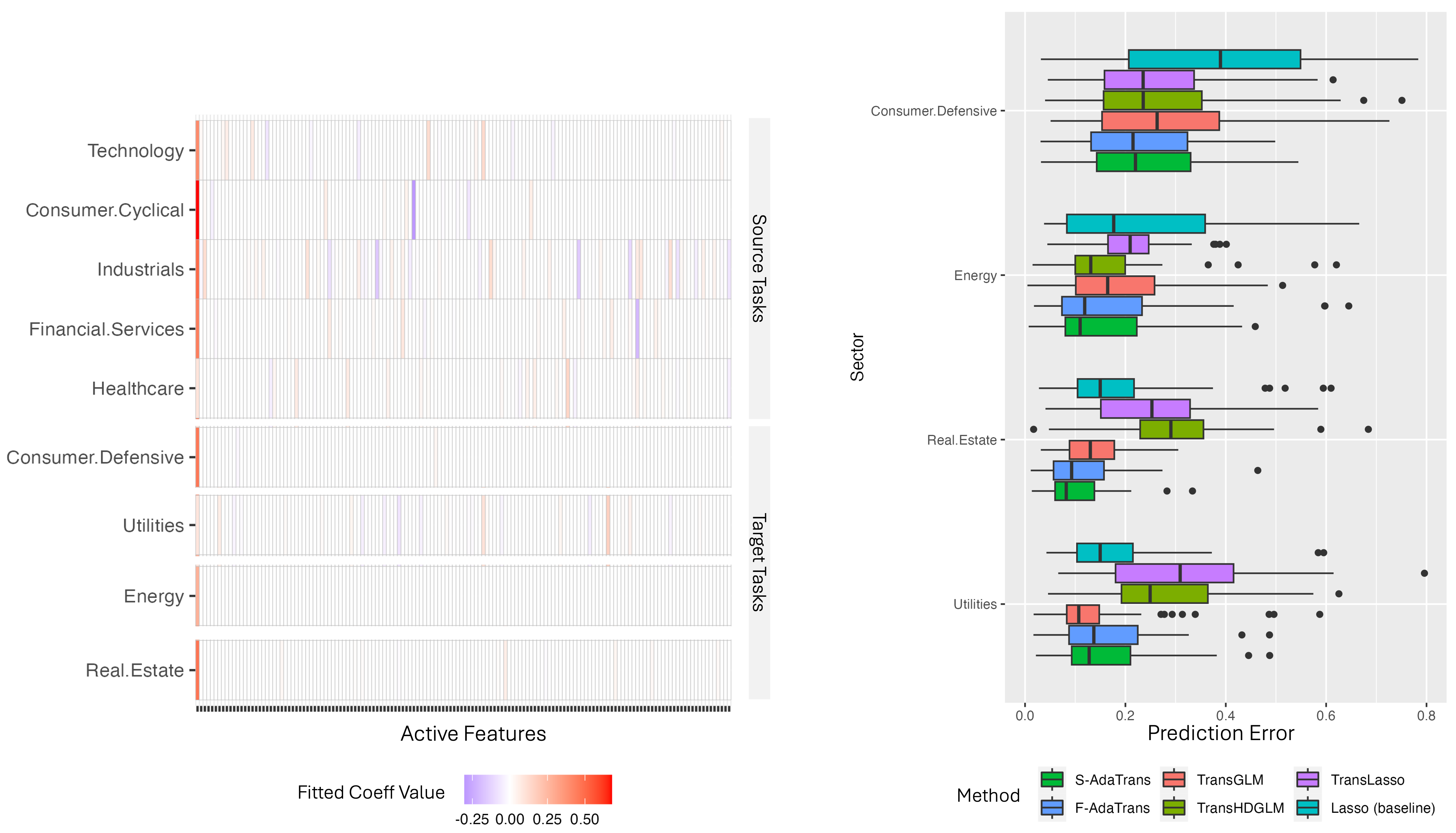}
    \caption{(Left) Fitted coefficient values of the 146 active features in different tasks. (Right) Prediction error of different methods on the target tasks.}
    \label{fig:boxplot}
\end{figure}

Treating stocks from each sector separately, we first conduct Lasso regression for each sector, selecting features that have non-zero effects in at least one of the sectors. 
Among all 560 metrics, this yields 146 active features, while the other 414 features demonstrate no effect across all sectors. Figure \ref{fig:boxplot} (left) shows the heterogeneous predictive effects (regression coefficients) of these 146 selected metrics across different sectors. We observe feature-wise transferable patterns among the sectors: some metrics have similar predictive effects across certain sectors, whereas others have distinct effects. In addition, some sectors, such as Industrials, display dense coefficient contrasts compared to other sectors, suggesting a sample-wise adaptive transfer approach. Based on Figure \ref{fig:boxplot} (left), we take sectors with top 5 largest sample sizes as sources, and each of the remaining $4$ sectors with smaller sample sizes as target. 

 All continuous data are standardized prior to analysis. The prediction performance is evaluated via 5-fold cross validation. This cross-validation process is replicated 10 times, yielding a total of 50 unique evaluations. 
 Figure \ref{fig:boxplot} (right) demonstrates the  $\ell_2$ prediction errors for six different transfer learning methods. 
 The results show that both \textit{F-} and \textit{S-AdaTrans} methods generally outperform other benchmarks in most experiments. In contrast, \textit{TransLasso} and \textit{TransHDGLM} consistently underperform the \textit{AdaTrans} method, and in some cases even underperform the Lasso baseline, suggesting potential negative transfer. This presumably because these methods fail to identify the transferable structure and informative source samples, resulting in loss of transferable information and introducing noises. Notably, \textit{AdaTrans} achieves its best performance in the Real Estate sector, the target with the smallest sample size among the four sectors analyzed. This highlights \textit{AdaTrans}'s capability to exploit adaptive transfer patterns effectively, especially when the target sample size is limited.
 
\section{Conclusion}
In this paper, we introduce an adaptive transfer learning method, \textit{AdaTrans}, enabling feature- and sample-specific transfer in high-dimensional regression. If the transferable information is known, we derive the optimal weights and the rate of these oracle estimators. When the transferable information is unknown, we further develop data-driven methods to compute the weights. We show both theoretically and empirically that by adaptively choosing the transferable information, \textit{AdaTrans} can outperform existing methods.


\bigskip

\appendix
\section*{\LARGE Appendix}

\renewcommand{\thesection}{\Alph{section}}

\section{Proof of Theorems, Propositions, and Corollaries}
\subsection{Proof of Theorem \ref{oracle_with_known_structure}}
Since we choose $w^{(k)}_{j} = \boldsymbol{1}_{\{j \in S^c_k\}}$ for $k=0,\dots,K$, the objective function (\ref{obj2}) can be simplified as 
\begin{align}
\label{LLA_formulation2}
\frac{n_T}{N} \mLz(\bz) + \sum^K_{k=1}\frac{n_S}{N} \mLk(\bz+ \bdk) +  \lz \sum_{j \in S_0^c}  |\bz_{j}| +  \lambda_{1} \sum^{K}_{k=1} \sum_{j \in S_k^c} |\bdk_{j}|,
\end{align}
where $\mLk(\bb) := \frac{1}{n_S} \|\byk-\bXk\bb\|_{2}^{2}$ is the local ordinary least squared (OLS) loss function. On the other hand, by the convexity of the OLS loss function, we find that for any $\bz, \bd^{(1)}, \dots, \bd^{(K)} \in \mathbb{R}^{p}$,
\begin{align}\label{convexity_property}
\begin{split}
&\left[\frac{n_T}{N} \mLz(\bz) + \sum^K_{k=1}\frac{n_S}{N} \mLk(\bz+ \bdk)\right] - \left[\frac{n_T}{N} \mLz(\hbzora) - \sum^K_{k=1}\frac{n_S}{N} \mLk(\hbzora+ \hbdkora) \right]  \\ 
&\ge \frac{n_T}{N} \sum^{p}_{j=1} \nabla_j \mLz(\hbzora) (\bz_j - \hbzoraj) + \sum^K_{k=1} \sum^p_{j=1} \frac{n_S}{N} \nabla_j \mLk(\hbzora+ \hbdkora) (\bz_j + \bdk_j - \hbzoraj - \hbdkoraj)  \\
&= \sum^p_{j=1} \left[ \frac{n_T}{N} \nabla_j \mLz(\hbzora)+ \sum^K_{k=1} \frac{n_S}{N} \nabla_j \mLk(\hbzora+ \hbdkora) \right] (\bz_j - \hbzoraj) \\
& \quad + \sum^p_{j=1} \sum^K_{k=1}  \frac{n_S}{N} \nabla_j \mLk(\hbzora+ \hbdkora) (\bdk_j - \hbdkoraj)  \\
&
= \sum_{j \in S_0^c} \left[ \frac{n_T}{N} \nabla_j \mLz(\hbzora)+ \sum^K_{k=1} \frac{n_S}{N} \nabla_j \mLk(\hbzora+ \hbdkora) \right] (\bz_j - \hbzoraj) \\
& \quad +  \sum^K_{k=1} \sum_{j \in S_k^c}  \frac{n_S}{N} \nabla_j \mLk(\hbzora+ \hbdkora) (\bdk_j - \hbdkoraj),    
\end{split}
\end{align}
where the last inequality uses the optimality conditions of problem (\ref{oracle_def}):
\begin{align*}
\frac{n_T}{N} \nabla_j \mLz(\hbzora)+ \sum^K_{k=1} \frac{n_S}{N} \nabla_j \mLk(\hbzora+ \hbdkora) = 0 \quad \text{for } j \in S_0,
\end{align*}
and
\begin{align*}
\nabla_j \mLk(\hbzora+ \hbdkora) = 0 \quad \text{for } j \in S_k, \quad k = 1, \dots, K.
\end{align*}

To prove that the oracle estimator is a solution to problem (\ref{obj2}), it suffices to show that the objective function gap
\begin{align*}
G(\bb, \bd) := &\left[\frac{n_T}{N} \mLz(\bz) + \sum^K_{k=1}\frac{n_S}{N} \mLk(\bz+ \bdk) + \lz \sum_{j \in S_0^c}   |\bz_{j}| +  \lk \sum^{K}_{k=1} \sum_{j \in S_k^c}  |\bdk_{j}|\right]\\
& \hspace{-1cm}-\left[\frac{n_T}{N} \mLz(\hbzora) + \sum^K_{k=1}\frac{n_S}{N} \mLk(\hbzora+ \hbdkora) + \lz \sum_{j \in S_0^c}  |\hbzoraj| + \lk  \sum^{K}_{k=1} \sum_{j \in S_k^c}   |\hbdkoraj|\right] 
\end{align*}
is non-negative for any $\bz, \bd^{(1)}, \dots, \bd^{(K)} \in \mathbb{R}^{p}$. From the definition of the oracle estimator (\ref{oracle_def}), we know that  for $j \in S_k^c$, $\hbdkoraj=0$ and for $j \in S_0^c$, $\hbzoraj=0$. This result, together with (\ref{convexity_property}), yields
\begin{align*}
G(\bb, \bd)\geq &\sum_{j \in S_0^c} \left[ \frac{n_T}{N} \nabla_j \mLz(\hbzora)+ \sum^K_{k=1} \frac{n_S}{N} \nabla_j \mLk(\hbzora+ \hbdkora) \right] (\bz_j - \hbzoraj)\\
& + \sum^K_{k=1}  \sum_{j \in S_k^c}  \frac{n_S}{N} \nabla_j \mLk(\hbzora+ \hbdkora) (\bdk_j - \hbdkoraj)+ \lz \sum_{j \in S_0^c}   |\bz_{j}| +  \lk \sum^{K}_{k=1} \sum_{j \in S_k^c}  |\bdk_{j}| \\
=&\sum_{j \in S_0^c} \left\{ \left[ \frac{n_T}{N} \nabla_j \mLz(\hbzora)+ \sum^K_{k=1} \frac{n_S}{N} \nabla_j \mLk(\hbzora+ \hbdkora) \right] \bz_j  + \lz   |\bz_{j}|\right\}  \\
&+ \sum^K_{k=1}   \sum_{j \in S_k^c}  \left\{ \frac{n_S}{N} \nabla_j \mLk(\hbzora+ \hbdkora) \bdk_j
+  \lk  |\bdk_{j}| \right\} \\
=&\sum_{j \in S_0^c} \left\{ \left[ \frac{n_T}{N} \nabla_j \mLz(\hbzora)+ \sum^K_{k=1} \frac{n_S}{N} \nabla_j \mLk(\hbzora+ \hbdkora) \right] \operatorname{sign}(\bz_{j}) + \lz\right\}  |\bz_{j}|  \\
&+ \sum^K_{k=1}   \sum_{j \in S_k^c}  \left\{ \frac{n_S}{N} \nabla_j \mLk(\hbzora+ \hbdkora) \operatorname{sign}(\bdk_{j})
+   \lk   \right\} |\bdk_{j}| \\
\ge & \sum_{j \in S_0^c} \left\{ \lz - \left| \frac{n_T}{N} \nabla_j \mLz(\hbzora)+ \sum^K_{k=1} \frac{n_S}{N} \nabla_j \mLk(\hbzora+ \hbdkora) \right| \right\}  |\bz_{j}|  \\
&+ \sum^K_{k=1}   \sum_{j \in S_k^c}  \left\{  \lk  - \left|\frac{n_S}{N} \nabla_j \mLk(\hbzora+ \hbdkora)\right|  \right\} |\bdk_{j}|.
\end{align*}

To finish the proof, we present the following lemma, proved in Appendix \ref{lemmaproof}: 
\begin{lemma}
\label{concentration_for_oracle}
Under Assumption \ref{A1} and \ref{A2}, if $n_{S} \gtrsim \log p$, then we have
\begin{align}
    \label{beta0_concentration} \left| \frac{n_T}{N} \nabla_j \mLz(\hbzora)+ \sum^K_{k=1} \frac{n_S}{N} \nabla_j \mLk(\hbzora+ \hbdkora) \right| \leq c_0\sqrt{\frac{\log p}{N}} \\
    \label{deltak_concentration}  \left|\frac{n_S}{N} \nabla_j \mLk(\hbzora+ \hbdkora)\right| \leq c_1\frac{n_S}{N}\sqrt{\frac{\log p}{n_S}}
\end{align}
with probability larger than $1 - \exp(-c_2 \log p) - \exp(-c_3 n_T)$, where $c_0$, $c_1$, $c_2$, $c_3$ are some universal constants.
\end{lemma}
Therefore, if we choose $\lambda_{0} > c_0 \sqrt{\frac{\log p}{N}}$ and $\lambda_{1} > c_1 \frac{n_S}{N} \sqrt{\frac{\log p}{n_S}}$, it always holds that $G(\bb, \bd) \ge 0$, which implies that the oracle estimator is a solution to problem (\ref{obj2}). Further noting that the strict equality $G(\bb, \bd) = 0$ holds if and only if $\bz_{S_0^c} = \boldsymbol{0}$ and $\bdk_{S_k^c} = \boldsymbol{0}$ for $k=1, \dots, K$. Given the uniqueness of the solution to problem (\ref{oracle_def}), it follows that the oracle estimator is indeed the unique solution to problem (\ref{obj2}), as claimed.

\subsection{Proof of Proposition \ref{OLS_trans_form} and Proposition \ref{oracle_rate}}
According to (\ref{oracle_def}), the oracle estimator satisfies
$$
\hbzora, \hat{\bd}_{\operatorname{ora}}^{(1)}, \dots, \hat{\bd}_{\operatorname{ora}}^{(K)} \in \underset{\bb^{(0)}_{S^c_0} = 0, \bd^{(1)}_{S^c_1} = 0, \dots, \bd^{(K)}_{S^c_K} = 0}{\text{argmin}} \left\{ \frac{1}{N} \left(\|\byz-\bXz_{S_0} \bz_{S_0}\|_{2}^{2} + \sum^{K}_{k=1} \|\byk-\bXk_{S_0} \bz_{S_0} - \bXk_{S_k}  \bdk_{S_k} \|_{2}^{2}\right)  \right\}.
$$
By the optimality conditions of the problem, we have
\begin{align}
    \label{opt_condition_1} &\left[ (\bXz_{S_0})^\top \bXz_{S_0} +\sum^{K}_{k=1}(\bXk_{S_0})^\top \bXk_{S_0} \right] \hbzoraS + \sum^{K}_{k=1}( \bXk_{S_0})^\top \bXk_{S_k} \hbdkoraS= (\bXz_{S_0})^\top \byz + \sum^K_{k=1}(\bXk_{S_0})^\top \byk, \\
    \label{opt_condition_2}
   &(\bXk_{S_k})^\top \bXk_{S_0} \hbzoraS + (\bXk_{S_k})^\top \bXk_{S_k} \hbdkoraS=(\bXk_{S_k})^\top \byk, \quad k \in [K].
\end{align}
Since we assume $|S_0| < n_T$, $\max_{1 \le k \le K} |S_k| < n_S$, $[(\bX_{S_{k}}^{(k)})^{\top} \bX_{S_{k}}^{(k)}]^{-1}$ is invertible for $k=0,\dots, K$. Therefore, plugging the results from (\ref{opt_condition_2}) into (\ref{opt_condition_1}) and subsequently canceling out $\hbdkoraS$s, we have
\begin{align}
\label{beta_oracle}
    \hbzoraS = \left[ (\bXz_{S_0})^\top \bXz_{S_0} + \sum^{K}_{k=1}(\bX^{(k)}_{S_0})^\top (\boldsymbol{I} - \boldsymbol{H}_{S_k}^{(k)})\bX^{(k)}_{S_0} \right]^{-1}  \left[ (\bXz_{S_0})^\top \byz + \sum^{K}_{k=1}(\bX_{S_0}^{(k)})^\top (\boldsymbol{I} - \boldsymbol{H}_{S_k}^{(k)})\byk \right] ,
\end{align}
and
\begin{align}
\label{delta_oracle}
\hbdkoraS=\left[\left(\bX_{S_{k}}^{(k)}\right)^{\top} \bX_{S_{k}}^{(k)}\right]^{-1}\left(\bX_{S_{k}}^{(k)}\right)^{\top}\left[\by^{(k)}-\bX_{S_{0}}^{(k)} \hbzoraS\right]
\end{align}
where $\Hk:=\bX_{S_k}^{(k)}[(\bXk_{S_k})^\top \bXk_{S_k}]^{-1}(\bXk_{S_k})^\top$ is the projection matrix. Therefore Proposition \ref{OLS_trans_form} is proved.

It remains to prove Proposition \ref{oracle_rate}.  Recall that we define $\bX_{S_0^c} =((\bXz_{S_0^c})^\top, ({\bX}_{S_0^c}^{(1)})^\top, \dots, ({\bX}_{S_0^c}^{(K)})^\top)^\top$, $\tbX_{S_0} =((\bXz_{S_0})^\top, (\tilde{\bX}_{S_0}^{(1)})^\top, \dots, (\tilde{\bX}_{S_0}^{(K)})^\top)^\top$ and $ \kappa_{F} := \left\| [\tbX_{S_0}^\top\tbX_{S_0}]^{-1}\tbX_{S_0}^\top \boldsymbol{\epsilon} \right\|_{\infty} / \left\|  [\bX_{S_0}^\top\bX_{S_0}]^{-1}\bX_{S_0}^\top \boldsymbol{\epsilon} \right\|_{\infty}$. According to the expression of $\hbzoraS$ outlined in (\ref{beta_oracle}), the $\ell_{\infty}$ estimation error bound satisfies
\begin{align*}
\|\hbzoraS - \bz_{S_0}\|_{\infty} &= \left\|  [\tbX_{S_0}^\top\tbX_{S_0}]^{-1}\tbX_{S_0}^\top \boldsymbol{\epsilon}  \right\|_{\infty} = \kappa_{F}\left\|  [\bX_{S_0}^\top\bX_{S_0}]^{-1}\bX_{S_0}^\top \boldsymbol{\epsilon}  \right\|_{\infty} \\
&= \kappa_{F} \max_{j \in S_0} \{ [e_j^\top  (\bX_{S_0}^\top\bX_{S_0})^{-1} ] \bX_{S_0}^\top \boldsymbol{\epsilon} \}
\le  \kappa_{F} \max_{j \in S_0}  \|e_j^\top  (\bX_{S_0}^\top\bX_{S_0})^{-1} \|_{1} \|\bX_{S_0}^\top \boldsymbol{\epsilon}\|_{\infty} \\
&\lesssim \kappa_{F}  \left\| \left(\frac{\bX_{S_0}^\top\bX_{S_0}}{N}\right)^{-1} \right\|_{\infty} \sqrt{\frac{\log s}{N}}.
\end{align*}
with probability larger than $1-c_{1} \exp \left(-c_{2} \log p \right)$, where in the last inequality we use the tail bound under Assumption \ref{A1} and \ref{A2}. For a detailed derivation of this tail bound, one can refer to the proof outlined in Lemma \ref{concentration}. This $\ell_{\infty}$ error bound, together with the $s$-sparsity of the oracle estimator $\hbzoraS$, implies
$$
\|\hbzora - \bz\|_{2} \leq \sqrt{s} \|\hbzora - \bz\|_{\infty} \lesssim \kappa_{F}  \left\| \left(\frac{\bX_{S_0}^\top\bX_{S_0}}{N}\right)^{-1} \right\|_{\infty} \sqrt{\frac{s \log s}{N}}
$$
as claimed. The two special cases discussed in the Proposition can be proved using similar arguments.

\subsection{Proof of Theorem \ref{oracle} and Proposition \ref{lasso_choice} }
We start with the fully detectable case (2). Based on the condition (\ref{localizable}) and the minimum signal strength condition, we have
$$
\begin{aligned}
& \left|\hbdkinitj\right| \leq\left|\bdk_{j}\right|+\left| \hbdkinitj-\bdk_j\right| \leq \frac{a_{2}}{2} \lambda_{1} \leq a_{2}\lambda_{1} \text { for } j \in S_{k}^{c} , \\
& \left|\hbdkinitj\right| \geq \left|\bdk_j\right|-\left| \hbdkinitj-\bdk_j\right| \geq h_{k}^{\wedge}-\frac{a_{2}}{2} \lambda_{1} \geq a \lambda_{1} \text { for } j \in S_{k} .
\end{aligned}
$$
for $k=1,\dots,K$. We can similarly establish that $|\hbz_j| \le a_2 \lambda_0$ for $j \in S_0^c$ and $|\hbz_j| \ge a \lambda_0$ for $j \in S_0$. Notice that the folded-concave penalty satisfies:
\begin{enumerate}
    \item $\mR_{\lambda}(t)$ is increasing and concave in $t \in[0, \infty)$ with $\mR_{\lambda}(0)=0$;
    \item $\mR_{\lambda}(t)$ is differentiable in $t \in(0, \infty)$ with $\mR_{\lambda}^{\prime}(0):=\mR_{\lambda}^{\prime}(0+) \geq a_1 \lambda$;
    \item $\mR_{\lambda}^{\prime}(t) \geq a_1 \lambda$ for $t \in\left(0, a_2 \lambda\right]$;
    \item $\mR_{\lambda}^{\prime}(t)=0$ for $t \in[a \lambda, \infty)$ with the pre-specified constant $a>a_2$.
\end{enumerate}
So according to the third and fourth properties of the folded concave penalty, we have
\begin{align}
\label{LLA_condition1}
\mR^{\prime}_{ \lambda_1} \left(| \hbdk_{\text{init},j} | \right)=0 \text{ for } j \in S_{k}, \quad \mR^{\prime}_{ \lambda_0} \left(| \hbz_{\text{init},j} | \right)=0 \text{ for } j \in S_{k}.
\end{align}
and
\begin{align}
\label{LLA_condition2}
\mR^{\prime}_{\lambda_1} \left(| \hbdk_{\text{init},j} | \right) \ge a_{1}\lambda_{1}  \text{ for } j \in S_{k}^c, \quad \mR^{\prime}_{ \lambda_0} \left(| \hbz_{\text{init},j} | \right) \ge a_{1}\lambda_{0} \text{ for } j \in S_{k}^c.
\end{align}
Therefore by setting $ \lambda_{0}w^{(0)}_{j} = \mR_{\lz}^{\prime}(|\bz_{\text{init}, j}|)$ and  $\lambda_{1}w^{(k)}_{j} = \mR_{\lambda_{1}}^{\prime}(|\bdk_{\text{init}, j}|)$, the arguments used in the proof of Theorem \ref{oracle_with_known_structure} can be directly applied here and yield the desired result. 

We then proceed to prove the first case. The proof consists of three steps. In the first step, we establish the convergence rate of a "sub-oracle" estimator. Then in the second step, we show that under some strict dual feasibility condition, such a sub-oracle estimator is the unique optimal solution to the problem (\ref{obj2}). Finally, we show that the strict dual feasibility condition holds under the given assumptions.

\paragraph{Step 1:} we start with analyzing the following sub-oracle estimator:
\begin{align}
\label{prob:suboracle}
\hbzsub,\left\{\hbdksub\right\}_{k=1}^{K} \in \underset{\bb_{\tSz^c}^{(0)}=0,\left\{\bdk_{\tSk^c}\right\}=0}{\operatorname{argmin}}\left\{\frac{1}{N} \sum_{k=0}^K \left\|\byk - \bXk (\bz+\bdk)\right\|_2^2 +\lambda_0\left\|\bz_{\tAz}\right\|_1+\lambda_1 \sum_{k=1}^K\left\|\bd_{\tAk}^{(k)}\right\|_1 \right\}
\end{align}

where we define $\tSz = S_0 \cup (\cup_{k=0}^K A_k)$, $\tSk = S_k \cup (\cup_{k=0}^K A_k)$ and $\tAz = \tAk = \cup_{k=0}^K A_k$. Compared to the definition of the oracle estimator outlined in (\ref{oracle_def}), the definition of the sub-oracle estimator involves an additional $\ell_{1}$ penalty on the non-detectable set $A_{k}$, $k=0,1, \cdots, K$. To handle this non-differentiable term, we first introduce the notation of subgradients $\hbzz \in \left.\partial\left\|\bz\right\|_{1}\right|_{\bz = \hbzsub}$ and $\hbzk \in \left.\partial\left\|\bdk\right\|_{1}\right|_{\bdk = \hbdksub}$ for $k \in[K]$. In the following arguments, we slightly abuse notation and denote by \(\hbzz_{\tAz}\) the vector in \(\mathbb{R}^{|\tSz|}\), constructed by selecting elements of \(\hbzz\) indexed by the set \(\tAz\), and setting all other elements to zero. $\hbzk_{\tAk}$ is defined in a similar way. Formally, we have
\[
(\hbzz_{\tAz})_i = \begin{cases} 
\hbzz_i & \text{if } i \in \tAz, \\
0 & \text{if } i \notin \tAz.
\end{cases}, \ 
(\hbzk_{\tAk})_i = \begin{cases} 
\hbzk_i & \text{if } i \in \tAk, \\
0 & \text{if } i \notin \tAk.
\end{cases}
\]

Then by the zero-subgradient condition of problem (\ref{prob:suboracle}), we have
{\small
\begin{align}
& {\left[\left(\bXztSz\right)^{\top} \bXztSz+\sum_{k=1}^{K}\left(\bXktSz\right)^{\top} \bXktSz\right]  \hbzsubS+\sum_{k=1}^{K}\left(\bXktSz\right)^{\top} \bXktSk \hbdksubS=\left(\bXztSz\right)^{\top} \by^{(0)}+\sum_{k=1}^{K}\left(\bXktSz\right)^{\top} \by^{(k)}-\frac{N}{2} \lambda_{0} \hbzz_{\tAz}} \label{opt_condition:S_0}\\
& \left(\bXktSk\right)^{\top} \bXktSz \hbzsubS+\left(\bXktSk\right)^{\top} \bXktSk \hbdksubS=\left(\bXktSk\right)^{\top} \by^{(k)}-\frac{N}{2} \lambda_{1} \hbzk_{\tAk}\label{opt_condition:S_k}
\end{align}
}
{\small
\begin{align}
& \left[\left(\bXztSzc\right)^{\top} \bXztSz+\sum_{k=1}^{K}\left(\bXktSzc\right)^{\top} \bXktSz\right] \hbzsubS+\sum_{k=1}^{K}\left(\bXktSzc\right)^{\top} \bXktSk \hbdksubS=\left(\bXztSzc\right)^{\top} \by^{(0)}+\sum_{k=1}^{K}\left(\bXktSzc\right)^{\top} \by^{(k)}-\frac{N}{2} \lambda_{0} \hbzz_{\tSzc} \label{opt_condition:S_0^c} \\
& \left(\bXktSkc\right)^{\top} \bXktSz \hbzsubS+\left(\bXktSkc\right)^{\top} \bXktSk \hbdksubS=\left(\bXktSkc\right)^{\top} \by^{(k)}-\frac{N}{2} \lambda_{1} \hbzk_{\tSkc} \label{opt_condition:S_k^c}
\end{align}
}
We start with the first two conditions (\ref{opt_condition:S_0}) and (\ref{opt_condition:S_k}). By the invertability of $\left(\bXktSk\right)^{\top} \bXktSk$, $k=0,1, \cdots, K$, we can reorganize (\ref{opt_condition:S_k}), plug the result into (\ref{opt_condition:S_0}) and get
$$
\begin{aligned}
\hbzsubS= & {\left[\left(\bXztSz\right)^{\top} \bXztSz+\sum_{k=1}^{K}\left(\tbXktSz\right)^{\top}\left(\tbXktSz\right)\right]^{-1} } \\
& \left[\left(\bXztSz\right)^{\top} \by^{(0)}+\sum_{k=1}^{K}\left(\tbXktSz\right)^{\top} \by^{(k)}-\frac{N}{2} \lambda_{0}  \hbzz_{\tAz}+\frac{N}{2} \lambda_{1} \sum_{k=1}^{K}\left(\bXktSz\right)^{\top} \bXktSk\left[\left(\bXktSk\right)^{\top}\bXktSk\right]^{-1}  \hbzk_{\tAk}\right] 
\end{aligned}
$$
where we define $\tbXktSz=\left(\boldsymbol{I}- \boldsymbol{H}_{\tSk}\right) \bXktSz$. This together with the closed form of the oracle estimator (\ref{oracle_form}) leads to the result in (\ref{form:sub_oracle_est}).

\paragraph{Step 2:} Now we show that under some feasibility conditions that will be specified later, the sub-oracle estimator is the unique solution to problem (\ref{prob:suboracle}). Similar to the proof of Theorem (\ref{oracle_with_known_structure}), we analyze the objective function gap
$$
\begin{aligned}
& \tilde{G}(\bb, \bd)=\left[\frac{n_T}{N} \mL^{(0)}\left(\bb^{(0)}\right)+\sum_{k=1}^{K} \frac{n_{S}}{N} \mL^{(k)}\left(\bb^{(0)}+\bd^{(k)}\right)+\lambda_{0}\left\|\bb_{\tAz \cup \tSzc}^{(0)}\right\|_{1}+\lambda_{1} \sum_{k=1}^{K}\left\|\bd_{\tAk \cup \tSkc }^{(k)}\right\|_{1}\right] \\
& -\left[\frac{n_T}{N} \mathcal{L}^{(0)}\left(\hbzsub\right)+\sum_{k=1}^{K} \frac{n_{S}}{N} \mathcal{L}^{(k)}\left(\hbzsub+\hbdksub\right)+\lambda_{0}\left(\hbzz_{\tAz \cup \tSzc }\right)^{\top} \hbz_{\text{sub}, \tAz \cup \tSzc }+\lambda_{1} \sum_{k=1}^{K}\left(\hbzk_{\tAk \cup \tSkc }\right)^{\top} \hbdk_{\text{sub}, \tAk \cup \tSkc}\right]
\end{aligned}
$$

By convexity of $\mathcal{L}^{(k)}(\cdot)$'s and the zero-subgradient condition (\ref{opt_condition:S_0}, \ref{opt_condition:S_k}, \ref{opt_condition:S_0^c}, \ref{opt_condition:S_k^c}) , we have
$$
\begin{aligned}
 \tilde{G}(\bb, \bd) \geqslant&\left[\frac{n_T}{N} \nabla \mathcal{L}^{(0)}\left(\hbzsub\right)+\sum_{k=1}^{K} \frac{n_{S}}{N} \nabla \mathcal{L}^{(k)}\left(\hbzsub+\hbdksub\right)\right]^{\top}\left(\bz-\hbzsub\right) \\
&+\sum_{k=1}^{K}\left[\frac{n_{S}}{N} \nabla \mathcal{L}^{(k)}\left(\hbzsub+\hbdksub\right)\right]^{\top}\left(\bd^{(k)}-\hbdksub\right) \\
& +\lambda_{0}\left\|\bb_{\tAz \cup \tSzc}^{(0)}\right\|_{1}+\lambda_{1} \sum_{k=1}^{K}\left\|\bd_{\tAk \cup \tSkc }^{(k)}\right\|_{1}-\lambda_{0}\left(\hbzz_{\tAz \cup \tSzc }\right)^{\top} \hbz_{\text{sub}, \tAz \cup \tSzc }-\lambda_{1} \sum_{k=1}^{K}\left(\hbzk_{\tAk \cup \tSkc }\right)^{\top} \hbdk_{\text{sub}, \tAk \cup \tSkc} \\
\geqslant&
-\lambda_{0}\left(\hbzz_{\tAz \cup \tSzc }\right)^{\top} \left(\bz_{\tAz \cup \tSzc } - \hbz_{\text{sub}, \tAz \cup \tSzc }\right)-\lambda_{1} \sum_{k=1}^{K}\left(\hbzk_{\tAk \cup \tSkc }\right)^{\top} \left(\bdk_{\tAk \cup \tSkc} - \hbdk_{\text{sub}, \tAk \cup \tSkc}\right)
\\
&+\lambda_{0}\left\|\bb_{\tAz \cup \tSzc}^{(0)}\right\|_{1}+\lambda_{1} \sum_{k=1}^{K}\left\|\bd_{\tAk \cup \tSkc }^{(k)}\right\|_{1}-\lambda_{0}\left(\hbzz_{\tAz \cup \tSzc }\right)^{\top} \hbz_{\text{sub}, \tAz \cup \tSzc }-\lambda_{1} \sum_{k=1}^{K}\left(\hbzk_{\tAk \cup \tSkc }\right)^{\top} \hbdk_{\text{sub}, \tAk \cup \tSkc}
\\
 =&\lambda_{0}\left(\left\|\bb_{\tAz \cup \tSzc}^{(0)}\right\|_{1} - \left(\hbzz_{\tAz \cup \tSzc }\right)^{\top} \hbz_{\text{sub}, \tAz \cup \tSzc }\right)+\lambda_{1} \left(\sum_{k=1}^{K}\left\|\bd_{\tAk \cup \tSkc }^{(k)}\right\|_{1} -\sum_{k=1}^{K}\left(\hbzk_{\tAk \cup \tSkc }\right)^{\top} \hbdk_{\text{sub}, \tAk \cup \tSkc}\right)
\end{aligned}
$$

Notice that by the definition of the sub-gradient, we have $\left\|\hbzz\right\|_{\infty} \leqslant 1$ and $\left\|\hbzk\right\|_{\infty} \leqslant 1$ for $k \in[K]$. Therefore, it must hold that $\tilde{G}(\bb, \bd) \geqslant 0$. If in addition, we have $\left\|\hbzz_{\tSzc}\right\|_{\infty}<1$ and $\left\|\hbzk_{\tSkc}\right\|_{\infty}<1$, then it implies that $\tilde{G}(\bb, \bd)=0$ only occurs when  $\bb_{\tSzc}^{(0)}=\hbz_{\text{sub},\tSzc} = \boldsymbol{0}$ and $\bd_{\tSkc}^{(k)}=\hbdk_{\text{sub}, \tSkc} = \boldsymbol{0}$. Given the invertability condition outlined in Theorem \ref{oracle}, the subproblem (\ref{prob:suboracle}) is strictly convex, and so has a unique minimizer $\hbzsub$ and $\hbdksub$. Therefore, in such case the sub-oracle estimator is the unique solution to problem (\ref{obj2}).

\paragraph{Step 3:}
It remains to establish the strict dual feasibility condition $\left\|\hbzz_{\tSzc}\right\|_{\infty}<1$ and $\left\|\hbzk_{\tSkc}\right\|_{\infty}<1$. We start with the zero-subgradient conditions (\ref{opt_condition:S_0^c}) and (\ref{opt_condition:S_k^c}). Substituting the results in (\ref{opt_condition:S_0}) and (\ref{opt_condition:S_k}) into (\ref{opt_condition:S_0^c}) and simplifying yields
\begin{align*}
\frac{N}{2} \lambda_0 \hbzz_{\tSzc}
=&\left(\bX_{\tSzc}^{(0)}\right)^{\top} \by^{(0)}+\sum_{k=1}^K\left(\bX_{\tSzc}^{(k)}\right)^{\top} \by^{(k)}\\
& -\left[(\bX_{\tSzc}^{(0)})^\top\bXztSz+\sum_{k=1}^K\left(\bX_{\tSzc}^{(k)}\right)^{\top} \bXktSz\right] \hbzsubS-\sum_{k=1}^K\left(\bX_{\tSzc}^{(k)}\right)^{\top} \bXktSk\hbdksubS \\
=&\bX_{\tSzc}^{\top} \by-\left(\bX_{\tSzc}^{\top} \bXtSz\right) \hbzsubS \\
& -\sum_{k=1}^K\left(\bX_{\tSzc}^{(k)}\right)^{\top} \bXktSk\left[\left(\bXktSk\right)^{\top} \bXktSk\right]^{-1}\left[\left(\bXktSk\right)^{\top} \by^{(k)}-\frac{N}{2} \lambda_1 \hbzk_{\tAk}-\left(\bXktSk\right)^{\top} \bXktSz \hbzsubS\right] \\
=&\tilde{\bX}_{\tSzc}^{\top} \by-\left(\tilde{\bX}_{\tSzc}^{\top} \tbXtSz\right) \hbzsubS+\frac{N}{2} \lambda_1 \sum_{k=1}^K \left(\bX_{\tSzc}^{(k)}\right)^{\top} \bXktSk\left[\left(\bXktSk\right)^{\top} \bXktSk\right]^{-1} \hbzk_{\tAk} \\
 =&\left(\tilde{\bX}_{\tSzc}^{\top} \tbXtSz\right) 
\left(\bz_{\tSz} - \hbzsubS\right)
+\tilde{\bX}_{\tSzc}^{\top} \boldsymbol{\epsilon} +\frac{N}{2} \lambda_1 \sum_{k=1}^K \left(\bX_{\tSzc}^{(k)}\right)^{\top} \bXktSk\left[\left(\bXktSk\right)^{\top} \bXktSk\right]^{-1} \hbzk_{\tAk} \\
 =&\tilde{\bX}_{\tSzc}^{\top} \boldsymbol{\epsilon} -\tilde{\bX}_{\tSzc}^{\top} \tbXtSz\left(\tbXtSz^{\top} \tbXtSz\right)^{-1}\left[\tbXtSz^{\top} \boldsymbol{\epsilon}-\frac{N}{2} \lambda_0 \hbzz_{\tAz}+\frac{N}{2} \lambda_1 \sum_{k=1}^K\left(\bXktSz\right)^{\top} \bXktSk\left[\left(\bXktSk\right)^{\top}\bXktSk\right]^{-1} \hbzk_{\tAk}\right] \\
& +\frac{N}{2} \lambda_1 \sum_{k=1}^K\left(\bX_{\tSzc}^{(k)}\right)^{\top} \bXktSk\left[\left(\bXktSk\right)^{\top} \bXktSk\right]^{-1} \hbzk_{\tAk},
\end{align*}
where the second-last equation is based on the model assumptions given in (\ref{m:target_model}) and (\ref{m:source_model}). 

Again by the definition of sub-gradients we have $\left\|\hat{\boldsymbol{z}}_{\tAk}^{(k)}\right\|_{\infty} \leqslant 1$. Applying Holder's inequality and triangle inequality leads to
$$
\begin{aligned}
\left\|\hbzz_{\tSzc}\right\|_{\infty} \leqslant & \frac{2}{N \lambda_{0}}\left\|\tilde{\bX}_{\tSzc}^{\top}\left(\boldsymbol{I}-\tbXtSz\left(\tbXtSz^{\top} \tbXtSz\right)^{-1} \tbXtSz^{\top}\right) \boldsymbol{\epsilon}\right\|_{\infty}+\left\|\tilde{\bX}_{\tSzc}^{\top} \tilde{\bX}_{\tSz}\left(\tilde{\bX}_{\tSz}^{\top} \tilde{\bX}_{\tSz}\right)^{-1}\hbzz_{\tAz}\right\|_{\infty} \\
& +\frac{\lambda_{1}}{\lambda_{0}}  \left\| \frac{1}{K}\sum_{k=1}^{K} \tbXtSzc^{\top}\tbH^{(K,k)}_{\tSz}\bXktSk\left[\left(\bXktSk\right)^{\top}\bXktSk\right]^{-1}\hbzk_{\tAk}\right\|_{\infty} \\
& +\frac{\lambda_{1}}{\lambda_{0}} \left\| \sum_{k=1}^K \left(\bX_{\tSzc}^{(k)}\right)^{\top} \bXktSk\left[\left(\bXktSk\right)^{\top} \bXktSk\right]^{-1} \hbzk_{\tAk} \right\|_{\infty},
\end{aligned}
$$
where we define $\tbH^{(K,k)}_{\tSz} =  \tilde{\bX}_{\tSz}\left(\tilde{\bX}_{\tSz}^{\top} \tilde{\bX}_{\tSz} / K\right)^{-1}\left(\bXktSz\right)^{\top}$.

Notice that by similar arguments as those in Lemma \ref{concentration}, we can show that by choosing $\lz \geq c_0 \sqrt{\frac{\log p}{N}}$ for some appropriate constant $c_0$, we have 
$$
 \frac{2}{N \lambda_{0}}\left\|\tilde{\bX}_{\tSzc}^{\top}\left(\boldsymbol{I}-\tbXtSz\left(\tbXtSz^{\top} \tbXtSz\right)^{-1} \tbXtSz^{\top}\right) \boldsymbol{\epsilon}\right\|_{\infty} \leq \alpha,
$$
with probability larger than $1-c_{1} \exp \left(-c_{2} \log p \right)$. This together with the conditions listed in Section \ref{mutual_cond:sec}, we then have $\left\|\hbzz_{\tSzc}\right\|_{\infty} <1 - \alpha < 1$.

We then proceed to show that $\left\|\hbzz_{\tSkc}\right\|_{\infty} <1 - \alpha < 1$. Following similar arguments, we can show that 
$$
\begin{aligned}
 \frac{N}{2} \lambda_{1} \hbzk_{S^c_k}=&\left(\bXktSkc\right)^{\top} \by^{(k)}-\left(\bXktSkc\right)^{\top} \bXktSz \hbzsubS-\left(\bXktSkc\right)^{\top} \bXktSk \hbdksubS \\
=&\left(\bXktSkc\right)^{\top} \by^{(k)}-\left(\bXktSkc\right)^{\top} \bXktSz \hbzsubS \\
& -\left(\bXktSkc\right)^{\top} \bXktSk\left[\left(\bXktSk\right)^{\top} \bXktSk\right]^{-1}\left[\left(\bXktSkc\right)^{\top} \by^{(k)}-\frac{N}{2} \lambda_{1} \hbzz_{\tAk}-\left(\bXktSk\right)^{\top} \bXktSz \hbzsubS\right] \\
=&\left(\tilde{\bX}_{\tSkc}^{(k)}\right)^{\top} \by^{(k)}-\left(\tilde{\bX}_{\tSkc}^{(k)}\right)^{\top} \tbXktSz \hbzsubS+\frac{N}{2} \lambda_{1}\left(\bXktSkc\right)^{\top} \bXktSk\left[\left(\bXktSk\right)^{\top} \bXktSk\right]^{-1} \hbzk_{\tAk} \\
=&\left(\tilde{\bX}_{\tSkc}^{(k)}\right)^{\top} \boldsymbol{\epsilon}^{(k)}+\left(\tilde{\bX}_{\tSkc}^{(k)}\right)^{\top} \tbXktSz\left(\bb_{\tSz}^{(0)}-\hbzsubS\right) \\
& +\frac{N}{2} \lambda_{1}\left(\bXktSkc\right)^{\top} \bXktSk\left[\left(\bXktSk\right)^{\top} \bXktSk\right]^{-1} \hbzk_{\tAk} \\
=&\left(\tilde{\bX}_{S_{k}^{c}}^{(k)}\right)^{\top} \boldsymbol{\epsilon}^{(k)}+\frac{N}{2} \lambda_{1}\left(\bXktSkc\right)^{\top} \bXktSk\left[\left(\bXktSk\right)^{\top} \bXktSk\right]^{-1} \hbzk_{\tAk}  \\
& -\left(\tbX_{\tSkc}^{(k)}\right)^{\top} \tbX_{\tSz}^{(k)}\left[\tbX_{\tSz}^{\top} \tbX_{\tSz}\right]^{-1}\left[\tilde{\bX}_{\tSz}^{\top} \boldsymbol{\epsilon}-\frac{N}{2} \lambda_{0} \hbzz_{\tAz}+\frac{N}{2} \lambda_{1} \sum_{k=1}^{K}\left(\bXktSz\right)^{\top} \bXktSk  \left[\left(\bXktSk\right)^{\top} \bXktSk\right]^{-1}  \hbzk_{\tAk}\right] 
\end{aligned}
$$
and
$$
\begin{aligned}
& \left\|\hbzk_{\tSkc}\right\|_{\infty} \leqslant \frac{2}{N \lambda_{1}}\left\|\left(\tilde{\bX}_{\tSkc}^{(k)}\right)^{\top}\left(\boldsymbol{\epsilon}^{(k)}-\tbXktSz\left(\tilde{\bX}_{\tSz}^{\top} \tilde{\bX}_{\tSz}\right)^{-1} \tilde{\bX}_{\tSz}^{\top}\boldsymbol{\epsilon}\right) \right\|_{\infty} +\left\|\left(\bXktSkc\right)^{\top} \bXktSk\left[\left(\bXktSk\right)^{\top} \bXktSk\right]^{-1} \hbzk_{\tAk}\right\|_{\infty} \\
& +\frac{\lambda_{0}}{\lambda_{1}}\left\|\left(\tilde{\bX}_{\tSkc}^{(k)}\right)^{\top} \tbXktSz\left(\tilde{\bX}_{\tSz}^{\top} \tilde{\bX}_{\tSz}\right)^{-1}\hbzz_{\tAz}\right\|_{\infty} +\left\| \frac{1}{K}\sum_{k=1}^{K} \left(\tilde{\bX}_{\tSkc}^{(k)}\right)^{\top}\bH_{\tSz}^{(k,k)} \bXktSk\left[\left(\bXktSk\right)^{\top}\bXktSk\right]^{-1} \hbzk_{\tAk}\right\|_{\infty},
\end{aligned}
$$
where we similarly define $\tbH_{\tSz}^{(k,k)} =  \tbXktSz\left(\tilde{\bX}_{\tSz}^{\top} \tilde{\bX}_{\tSz}/K\right)^{-1}\left(\bXktSz\right)^{\top}$.

Using similar arguments as those in the proof of Lemma \ref{concentration}, we can show that as long as we choose $\lk = c_{0}\sqrt{\frac{n_S}{N}\frac{\log p}{N}}$ for some appropriate constant $c_0$, the first term in the bound is again no larger than $\alpha$ with high probability. A further application of the conditions in Section \ref{mutual_cond:sec} would yield $\left\|\hbzk_{\tSkc}\right\|_{\infty} <1 - \alpha < 1$, the proof of the theorem is then finished.

\subsection{Proof of Corollary \ref{partial_rate}}
Notice that for any $k = 0 \dots, K$, $\tSk = S_k \cup (\cup_{k=0}^K A_k) \supset S_k$. So from the model setup (\ref{m:target_model},\ref{m:source_model}) and the property of the projection matrix $\bH^{(k)}_{\tSk}:=\bX_{\tSk}^{(k)}[(\bXk_{\tSk})^\top \bXk_{\tSk}]^{-1}(\bXk_{\tSk})^\top$, we have 
$$
\left(\bXztSz\right)^{\top}\by^{(0)} = \left(\bXztSz\right)^{\top}\bXz_{\tSz} \bb^{(0)}_{\tSz} + \left(\bXztSz\right)^{\top}\bez
$$
and
$$
\left(\bXktSz\right)^{\top}\left(\bI - \bH^{(k)}_{\tSk}\right)\by^{(k)} = \left(\bXktSz\right)^{\top}\left(\bI - \bH^{(k)}_{\tSk}\right)\bXk_{\tSz} \bb^{(k)}_{\tSz} + \left(\bXztSz\right)^{\top}\left(\bI - \bH^{(k)}_{\tSk}\right)\bek
$$

Similar to the arguments in Proposition \ref{oracle_rate}, we define $\bX_{\tSzc} =((\bXz_{\tSzc})^\top, ({\bX}_{\tSzc}^{(1)})^\top, \dots, ({\bX}_{\tSzc}^{(K)})^\top)^\top$, $\tbX_{\tSz} =((\bXz_{\tSz})^\top, (\tilde{\bX}_{\tSz}^{(1)})^\top, \dots, (\tilde{\bX}_{\tSz}^{(K)})^\top)^\top$ and $ \tilde{\kappa}_{F} := \left\| [\tbX_{\tSz}^\top\tbX_{\tSz}]^{-1}\tbX_{\tSz}^\top \boldsymbol{\epsilon} \right\|_{\infty} / \left\|  [\bX_{\tSz}^\top\bX_{\tSz}]^{-1}\bX_{\tSz}^\top \boldsymbol{\epsilon} \right\|_{\infty}$. Thus, under the partially detectable case in Theorem \ref{oracle}, we have
\begin{align}
\label{form:sub_oracle_decomp}
\hbz_{\tSz}= & {\bigg[\left(\bXztSz\right)^{\top} \bXztSz+\sum_{k=1}^{K}\left(\bXktSz\right)^{\top}\left(\bI - \bH_{\tSk}\right)\bXk_{\tSz}\bigg]^{-1} }  \bigg[\left(\bXztSz\right)^{\top} \by^{(0)} \nonumber \\
&+\sum_{k=1}^{K}\left(\bXktSz\right)^{\top}\left(\bI - \bH_{\tSk}\right)\by^{(k)}-\frac{N}{2} \lambda_{0}  \hbzz_{A}+\frac{N}{2} \lambda_{1} \sum_{k=1}^{K}\hat{\boldsymbol{B}}_{\tSz, \tSk}^{\top}  \hbzk_{A}\bigg], \nonumber \\
=&  \bz_{\tSz} + \left[\left(\tbX_{\tSz}\right)^{\top} \tbX_{\tSz}\right]^{-1} \left[\left(\tbX_{\tSz}\right)^{\top} \boldsymbol{\epsilon} \right] + \left[\frac{\left(\tbX_{\tSz}\right)^{\top} \tbX_{\tSz}}{N}\right]^{-1}\left(-\frac{1}{2}\lz\hbzz_{A} + \frac{1}{2}\lk \sum_{k=1}^{K}\hat{\boldsymbol{B}}_{\tSz, \tSk}^{\top}  \hbzk_{A}\right) 
\end{align}
and subsequently
\begin{align}
\label{partial_detectable_bound_app}
\|\hbz_{\tSz} -  \bz_{\tSz}\|_{\infty}  &\leq \tilde{\kappa}_{F} \max_{j \in \tSz}  \|e_j^\top  (\bX_{\tSz}^\top\bX_{\tSz})^{-1} \|_{1} \|\bX_{\tSz}^\top \boldsymbol{\epsilon}\|_{\infty} + \frac{1}{2}\|\tilde{\boldsymbol{\Omega}}_{\tSz, N}\|_{\infty} (\lz + \|\hat{\boldsymbol{B}}\|_{\infty}\lk) \\
&= \tilde{\kappa}_{F} \|\hat{\boldsymbol{\Omega}}_{\tSz, N}\|_{\infty} \|\bX_{\tSz}^\top \boldsymbol{\epsilon}\|_{\infty} + \frac{1}{2}\|\tilde{\boldsymbol{\Omega}}_{\tSz, N}\|_{\infty} (\lz + \|\hat{\boldsymbol{B}}\|_{\infty}\lk)
\end{align}
where we define $ \tilde{\kappa}_{F} := \left\| [\tbX_{\tSz}^\top\tbX_{\tSz}]^{-1}\tbX_{\tSz}^\top \boldsymbol{\epsilon} \right\|_{\infty} / \left\|  [\bX_{\tSz}^\top\bX_{\tSz}]^{-1}\bX_{\tSz}^\top \boldsymbol{\epsilon} \right\|_{\infty}$, $\hat{\boldsymbol{\Omega}}_{\tSz, N} = \left[\left(\bX_{\tSz}\right)^{\top} \bX_{\tSz}/N\right]^{-1}$, $\tilde{\boldsymbol{\Omega}}_{\tSz, N} = \left[\left(\tbX_{\tSz}\right)^{\top} \tbX_{\tSz}/N\right]^{-1}$ and $\hat{\boldsymbol{B}} = \sum_{k=1}^{K}\hat{\boldsymbol{B}}_{\tSz, \tSk} \hbzk_{A}$.

To proceed, we first present the following lemma on the properties of the initial estimator obtained from separate Lasso regression.

\begin{lemma}
\label{lasso_choice}
Given Assumptions \ref{A1} and \ref{A2}, and provided that $n_{S} > n_{T} \gtrsim s^2\log p$, if we run separate Lasso regression on each source sample and the target sample with penalty $\lambda_{\operatorname{Lasso}} \asymp \sqrt{\log p / n_T}$, and construct the initial estimator as
$$
\hbz_{\text{init}} = \hbz_{\text{Lasso}}, \quad \hbdk_{\text{init}} = \hbk_{\text{Lasso}} - \hbz_{\text{Lasso}}, \quad k \in [K],
$$ then by choosing $\lambda_{0} = \lambda_{1} =c_{0}\sqrt{\frac{\log p}{n_{T}}}$ for some constant $c_0$, we have
for all $k \in [K]$,  with probability larger than $1 - \exp(-c_1 \log p) - \exp(-c_2 n_T)$,
\begin{align}
\label{localizable}
\!\left\|\hbz_{\text{init}}-\bz\right\|_{\infty} \leq \frac{a_2}{2}  \lambda_0,  \ \left\|\hat{\bd}_{\text{init}}^{(k)}-\bd^{(k)}\right\|_{\infty} \leq \frac{a_2}{2} \lambda_1,
\end{align}
where $a_{2} \ge 0$ and $a_{1} \ge 0$ are constants specified in Appendix \ref{SCAD}.
\end{lemma}
Lemma \ref{lasso_choice} is a direct consequence of Corollary 9.27 in \cite{wainwright2019high}. It provides a choice of $\lz$ and $\lk$ that satisfies condition (\ref{localizable}) in Theorem \ref{oracle}. Plugging the result back into (\ref{partial_detectable_bound}) and applying the concentration inequality utilized in Lemma \ref{concentration} lead to
\begin{align}
\label{partial_with_lasso}
\|\hbz_{\tSz} -  \bz_{\tSz}\|_{\infty}  \lesssim \tilde{\kappa}_{F} \|\hat{\boldsymbol{\Omega}}_{\tSz, N}\|_{\infty} \sqrt{\frac{\log(s+a)}{N}} + \|\tilde{\boldsymbol{\Omega}}_{\tSz, N}\|_{\infty} (1 + \|\hat{\boldsymbol{B}}\|_{\infty})\sqrt{\frac{\log p}{n_{T}}}.
\end{align}
This together with (\ref{form:sub_oracle_decomp}) implies
\begin{align*}
\|\hbz -  \bz\|_{2}^2 = \|\hbz_{\tSz} -  \bz_{\tSz}\|_{2}^2 &= \|\hbz_{S_0\backslash A} -  \bz_{S_0 \backslash A}\|_{2}^2 + \|\hbz_{A} -  \bz_{A}\|_{2}^2 \\
&\lesssim \tilde{\kappa}_{F}^2\|\hat{\boldsymbol{\Omega}}_{\tSz, N}\|_{\infty}^2 \frac{s\log (s+a)}{N} + \|\tilde{\boldsymbol{\Omega}}_{\tSz, N}\|_{\infty}^2   (1 + \|\hat{\boldsymbol{B}}\|_{\infty})^2 \frac{a\log p}{n_{T}},
\end{align*}
then the proof of bound (\ref{partial_detectable_bound}) is finished. Next, we start back at (\ref{form:sub_oracle_decomp}) to discuss the rate under the two special cases mentioned in Corollary \ref{partial_rate}.

\paragraph{Case 1:} We first discuss the error bound under the fully transferable case $\bXk_{S_k} \perp \bXk_{S_0}$, $k\in [K]$. Since in addition we assume $\bXk_{S_k \backslash A} \perp \bXk_{A}$ for all $k \in [K]$ and $\bXk_{S_0 \backslash A} \perp \bXk_{A}$, we have 
{\small
\begin{align*}
\sum_{k=1}^{K}\hat{\boldsymbol{B}}_{\tSz, \tSk} \hbzk_{A} &= \sum_{k=1}^{K}\left[\left(\bX_{S_k\cup A}^{(k)}\right)^{\top}\left(\bX_{S_k\cup A}^{(k)}\right)\right]^{-1}\left(\bX_{S_k\cup A}^{(k)}\right)^{\top} \bX_{S_0\cup A}^{(k)} \hbzk_{A}\\
&= \sum_{k=1}^K 
\begin{pmatrix}
    \left[\left(\bX_{S_k\backslash A}^{(k)}\right)^{\top}\left(\bX_{S_k\backslash A}^{(k)}\right)\right]^{-1} & \boldsymbol{O} \\
    \boldsymbol{O} & \left[\left(\bX_{ A}^{k)}\right)^{\top}\left(\bX_{ A}^{(k)}\right)\right]^{-1}
\end{pmatrix} 
\begin{pmatrix}
    \left(\bX_{S_k\backslash A}^{(k)}\right)^{\top}\\
    \left(\bX_{ A}^{(k)}\right)^{\top}
\end{pmatrix} 
\bX_{S_0\cup A}^{(k)} \hbzk_{A} \\
&= \sum_{k=1}^K \hbzk_{A}
\end{align*}
}
Here without loss of generality, we assume that the sub-matrices indexed by $S_k \backslash A$ and $A$ occupy contiguous blocks in the matrix. On the other hand, since we assume $\bXk_{S_0 \backslash A} \perp \bXk_{A}$ for all $k=0, \dots, K$, we have
\begin{align*}
&\left[\left(\tbX_{\tSz}\right)^{\top} \tbX_{\tSz}\right]^{-1} \left[\left(\tbX_{\tSz}\right)^{\top} \boldsymbol{\epsilon} \right] \\
=& {\bigg[\left(\bXztSz\right)^{\top} \bXztSz+\sum_{k=1}^{K}\left(\bXktSz\right)^{\top}\left(\bI - \bH_{\tSk}\right)\bXk_{\tSz}\bigg]^{-1} }  \bigg[\left(\bXztSz\right)^{\top} \bez+\sum_{k=1}^{K}\left(\bXktSz\right)^{\top}\left(\bI - \bH_{\tSk}\right)\bek \bigg] \\
=& 
\begin{pmatrix}
\left[\sum_{k=0}^{K}\left(\bXk_{S_0 \backslash A}\right)^{\top}\bXk_{S_0 \backslash A}\right]^{-1} & \boldsymbol{O} \\
\boldsymbol{O} & \left[\left(\bXz_{A}\right)^{\top}\bXz_{A} \right]^{-1}
\end{pmatrix}
\begin{pmatrix}
    \sum_{k=0}^{K}\left(\bXk_{S_0 \backslash A}\right)^{\top}\bek \\
    \left(\bXz_{A}\right)^{\top}\bez
\end{pmatrix}
\end{align*}
Therefore, in this case, we find
\begin{align*}
    \hbz_{S_0 \backslash A} - \bz_{S_0 \backslash A} = \left[\left(\bX_{S_0 \backslash A}\right)^{\top} \bX_{S_0 \backslash A}\right]^{-1} \left[\left(\bX_{S_0 \backslash A}\right)^{\top} \boldsymbol{\epsilon} \right],
\end{align*}
and
\begin{align*}
    \hbz_{A} - \bz_{A} = \left[\left(\bXz_{A}\right)^{\top} \bXz_{A}\right]^{-1} \left[\left(\bXz_{A}\right)^{\top} \bez \right] + \left[\frac{\left(\bXz_{A}\right)^{\top} \bXz_{A}}{N}\right]^{-1}\left(-\frac{1}{2}\lz\hbzz_{A} + \frac{1}{2}\lk \sum_{k=1}^{K}\hbzk_{A}\right).
\end{align*}
this together with the fact that $\|\hbzz_{A}\|_{\infty} \leq 1$ and $\|\hbzk_{A}\|_{\infty} \leq 1$ leads to the result in case 1.

\paragraph{Case 2:} we then discuss the error bound under the fully non-transferable case $S_k  \supset S_0$ for $k \in [K]$. In this case, we have $\tbXk_{S_k \backslash A} = \boldsymbol{O}$. As we still assume $\bXk_{S_k \backslash A} \perp \bXk_{A}$ for all $k \in [K]$ and $\bXk_{S_0 \backslash A} \perp \bXk_{A}$, following the similar arguments as those in case 1 leads to
\begin{align*}
\sum_{k=1}^{K}\hat{\boldsymbol{B}}_{\tSz, \tSk} \hbzk_{A} = \sum_{k=1}^K \hbzk_{A}
\end{align*}
and
\begin{align*}
&\left[\left(\tbX_{\tSz}\right)^{\top} \tbX_{\tSz}\right]^{-1} \left[\left(\tbX_{\tSz}\right)^{\top} \boldsymbol{\epsilon} \right] \\
=& 
\begin{pmatrix}
\left[\left(\bXz_{S_0 \backslash A}\right)^{\top}\bXz_{S_0 \backslash A}\right]^{-1} & \boldsymbol{O} \\
\boldsymbol{O} & \left[\left(\bXz_{A}\right)^{\top}\bXz_{A} \right]^{-1}
\end{pmatrix}
\begin{pmatrix}
    \left(\bXz_{S_0 \backslash A}\right)^{\top}\bez \\
    \left(\bXz_{A}\right)^{\top}\bez
\end{pmatrix}
\end{align*}
Therefore, in this case, we find
\begin{align*}
    \hbz_{S_0 \backslash A} - \bz_{S_0 \backslash A} = \left[\left(\bXz_{S_0 \backslash A}\right)^{\top} \bXz_{S_0 \backslash A}\right]^{-1} \left[\left(\bXz_{S_0 \backslash A}\right)^{\top} \bez \right],
\end{align*}
and
\begin{align*}
    \hbz_{A} - \bz_{A} = \left[\left(\bXz_{A}\right)^{\top} \bXz_{A}\right]^{-1} \left[\left(\bXz_{A}\right)^{\top} \bez \right] + \left[\frac{\left(\bXz_{A}\right)^{\top} \bXz_{A}}{N}\right]^{-1}\left(-\frac{1}{2}\lz\hbzz_{A} + \frac{1}{2}\lk \sum_{k=1}^{K}\hbzk_{A}\right).
\end{align*}
Therefore the error bound could be established accordingly.

\subsection{Proof of Theorem \ref{onesteprate}}
\label{theoremproof}
Throughout the proof, we adopt the following notations to analyze the solution of the problem (\ref{P:sample-transfer}):
\begin{align}
\label{transrule}
\btheta^{*}= \left(\begin{array}{c}
(\btheta^{*})^{(1)} \\
(\btheta^{*})^{(2)} \\
\vdots \\
(\btheta^{*})^{(K)} \\
(\btheta^{*})^{(0)}
\end{array}\right) := 
\left(\begin{array}{c}
\bd^{(1)} \\
\bd^{(2)}\\
\vdots \\
\bd^{(K)} \\
\bb^{(0)}
\end{array}\right),
\end{align}
where  $\bz$ is the target model parameter and $\bdk$s are source task-specific signals. Under this transformation, solving problem (\ref{P:sample-transfer}) is equivalent as solving
\begin{equation}
\label{transobj}
\hat{\btheta} \in \underset{\btheta}{\operatorname{argmin}} \{\mL(\btheta)+\lambda_{0} \mR(\btheta)\}, \quad \text{subject to } \quad \frac{1}{n_T}\|(\bXz)^\top (\byz - \bXz \btheta^{(0)})\|_{\infty} \le \lambda_{T},
\end{equation}
where we define $\mL(\btheta) := \frac{1}{2N}\sum^{K}_{k=0} w_k\|\byk-\bXk(\btheta^{(0)}+\btheta^{(k)} \boldsymbol{1}_{\{k \neq 0\}})\|_{2}^{2}$ and $\mR(\btheta) :=\sqrt{\sum_{k=0}^{K}\frac{n_{k}}{N}w_{k}^2} \|\btheta^{(0)} \|_{1} + \sum^{K}_{k=1} \frac{\lk}{\lz} w_{k} \| \btheta^{(k)} \|_{1}$ for any $\btheta \in \mathbb{R}^{(K+1)p}$. Since there exists a one-to-one transformation between $\btheta^{*}$ and $\bb$, we can quantify the estimation error $\hat{\bb} - \bb$ by analyzing $\hat{\btheta} - \btheta^{*}$. 

Recall that we define $S_0$ as the support set of the target coefficient $\bz$ and $S_0^c$ as the complement set of $S_0$. We define $\btheta^{(k)}_{S_0}$ as sub-vector indexed by $S_0$. Throughout the proof, we define $n_k = n_S$ for $k = 1, \dots, K$ and $n_k = n_T$ for $k = 0$.

We first establish an essential property of sub-Gaussian design matrices, which serves as fundamental building blocks for our subsequent analysis. The following lemma \ref{RSC} shows that the least square objective function has a \textit{restricted} strongly convex (RSC) and \textit{restricted} smooth (RSM) property \citep{loh2011high,agarwal2010fast}. 
\begin{lemma}[RSC and RSM property] 
\label{RSC}
Under Assumption \ref{A1}, for any $\boldsymbol{\bDelta} \in \mathbb{R}^{p}$, with probability at least $1 - c_1 \exp (-c_2 n_k)$,
    \begin{align*}
    \frac{1}{n_k}\left\|\bX^{(k)}\boldsymbol{\bDelta} \right\|_2^2
&=
\boldsymbol{\bDelta}^{\top}\hat{\boldsymbol{\Sigma}}^{(k)}\boldsymbol{\bDelta} \geq \alpha_k\|\boldsymbol{\bDelta}\|_2^2-\beta_k \frac{\log p}{n_k}\|\boldsymbol{\bDelta}\|_1^2, \\
\frac{1}{n_k}\left\|\bX^{(k)}\boldsymbol{\bDelta} \right\|_2^2
&=
\boldsymbol{\bDelta}^{\top}\hat{\boldsymbol{\Sigma}}^{(k)}\boldsymbol{\bDelta} \leq \gamma_k\|\boldsymbol{\bDelta}\|_2^2+\tau_k \frac{\log p}{n_k}\|\boldsymbol{\bDelta}\|_1^2, 
    \end{align*}
    where $\alpha_{k} = \frac{1}{2}\Lambda_{\min}(\Sigk) \geq 1/c$, $\gamma_{k} = 2\Lambda_{\max}(\Sigk) \leq c$ and $\beta_{k}, \tau_{k} \leq c$, $n_0 = n_T$, and 
    $n_{k} = n_{S}$ for $k = 1, \dots, K$.
\end{lemma}
 
Define $\hat{\bDelta}:=\hat{\btheta}-\btheta^{*}$ as the estimation error of $\hat{\btheta}$ and the corresponding $k$-th block {$\hat{\bDelta}^{(k)}:=\hat{\btheta}^{(k)}-(\btheta^{*})^{(k)}$}. {For brevity, we will omit the superscript and write $(\btheta^{*})^{(k)}$ as $\btheta^{(k)}$ for $0 \le k \le K$ when there is no ambiguity. } Our goal is to establish an upper bound for $\|\hat{\bDelta}^{(0)}\|_2^2$, i.e., the $\ell_{2}$ estimation error of the target parameter $\bb^{(0)}$.

The proof of Theorem \ref{onesteprate} relies on three key technical lemmas, whose proofs are provided in Appendix \ref{lemmaproof}.
The first lemma establishes an upper bound for the first-order term of the Taylor series expansion of $\mL(\btheta)$.  

\begin{lemma}
\label{concentration}
Under Assumption \ref{A1} and \ref{A2}, if $n_{S} \gtrsim \log p$, then by choosing $\lambda_{0} \gtrsim c_{0}  \sqrt{\frac{\log p}{N}}$ and $\lk  \gtrsim c_{0} \sqrt{\frac{n_{S}}{N} \frac{\log p}{N}}$ for some appropriate constant $c_{0}$, we have for any $\bDelta = \left(\left(\bDelta^{(1)}\right)^{\top}, \dots, \left(\bDelta^{(K)}\right)^{\top}, \left(\bDelta^{(0)}\right)^{\top} \right)^{\top}\in \mathbb{R}^{(K+1) p}$,
$$
\left|\left\langle\nabla \mL\left(\btheta^{*}\right), \bDelta\right\rangle\right|
\leq \sum_{k=1}^{K} \frac{\lk}{2} w_k \left\|\bDelta^{(k)}\right\|_{1}+\frac{\lz}{2}\sqrt{\sum_{k=0}^{K}\frac{n_{k}}{N}w_{k}^2} \left\|\bDelta^{(0)}\right\|_{1} 
$$
with probability larger than $1-c_{1} \exp \left(-c_{2} \log p \right)$.   
\end{lemma}

 The next lemma establishes a restricted set of directions in which $\hbDelta$ lies.

\begin{lemma}
    \label{Thm1Lemma1}
     Under Assumption \ref{A1} and \ref{A2}, and the conditions of Lemma \ref{concentration}, the estimation error $\hat{\bDelta}$ satisfies the inequality
    \begin{align*}
\sum_{k=1}^{K}\lk w_k \left\|\hat{\bDelta}^{(k)}\right\|_{1}+\lambda_{0}\sqrt{\sum_{k=0}^{K}\frac{n_{k}}{N}w_{k}^2}\left\|\hat{\bDelta}^{(0)}\right\|_{1} \leq 4 \lambda_{0}\sqrt{\sum_{k=0}^{K}\frac{n_{k}}{N}w_{k}^2}\left\|\hat{\bDelta}_{S_0}^{(0)}\right\|_{1}+4 \sum_{k=1}^{K}\lk w_k h_{k}
    \end{align*}
with probability larger than $1-c_{1} \exp \left(-c_{2} \log p \right)$.
\end{lemma}

The following lemma ensures a property analogous to restricted strong convexity for $\hat{\bDelta}$.
\begin{lemma}    \label{Thm1Lemma2}
      Define $S_N = \{k: w_k = 0\}$ and $\underline{w} = \min_{k \in S_N^c} w_k$. Under Assumption \ref{A1} and \ref{A2} and the conditions of Lemma \ref{Thm1Lemma1}, if $n_S>n_T$,  the estimation error $\hat{\bDelta}$ satisfies 
\begin{align}
\label{WRSC}
& \mathcal{L}\left(\btheta^{*}+\hat{\bDelta}\right)-\mathcal{L}\left(\btheta^{*}\right)-\left\langle\nabla \mathcal{L}\left(\btheta^{*}\right), \hat{\bDelta}\right\rangle \\
& \geq \left(\frac{\alpha_{\min }^{2}}{\gamma_{0}}  \sum^{K}_{k=0} \frac{n_k w_k}{N} - u_n\right) \left\|\hat{\bDelta}^{(0)}\right\|_{2}^{2}+ \frac{\alpha_{\min }^{2}}{\gamma_{0}}\sum_{k=1}^{K} \frac{n_{k} w_k}{N} \left\|\hat{\bDelta}^{(k)}\right\|_{2}^{2} -\frac{2 \alpha_{\max}}{\gamma_{0}} \sum^{K}_{k=1} \frac{n_{k}w_k}{N}\lambda_{T} \left\|\hat{\bDelta}^{(k)}\right\|_{1} - v_n  \sum_{k=1}^{K} \lk w_k h_{k}
\end{align}
with probability larger than $1 - \exp(c_1 n_{T}) - \exp(c_2 \log p)$, where
\begin{align*}
u_n &=  \frac{64(\alpha_{\max} \tau_{0} + \beta_{\max} \gamma_{0})}{\gamma_{0}} \frac{n_S}{n_T} \frac{\log p}{N} \cdot \frac{\lz^2 s (\sum_{k=0}^{K}\frac{n_{k}}{N}w_{k}^2)}{\lk^2 \underline{w} \wedge [ (\lz^2 \sum_{k = 0}^K \frac{n_S}{N} w_k^2)/ ( (n_T/n_S) w_0 + \sum_{k = 1}^K  w_k)]},\\
v_n &= \frac{64(\alpha_{\max} \tau_{0} + \beta_{\max}\gamma_{0})}{\gamma_{0}} \frac{n_S}{n_T} \frac{\log p}{N} \cdot \frac{\sum_{k=1}^{K} \lk w_k h_{k}}{\lk^2 \underline{w} \wedge [ (\lz^2 \sum_{k = 0}^K \frac{n_S}{N} w_k^2)/ ( (n_T/n_S) w_0 + \sum_{k = 1}^K  w_k)]}, \\
\alpha_{\min} &= \min_{0 \le k \le K} \alpha_{k}, \quad \alpha_{\max} = \max_{0 \le k \le K} \alpha_{k} \text{ and} \quad \beta_{\max} = \max_{0 \le k \le K} \beta_{k}
\end{align*}
with RSC constants $(\alpha_{k}, \beta_{k})$ and RSM constants  $(\gamma_{k}, \tau_{k})$ defined in Lemma \ref{RSC}.
\end{lemma}

We now turn to the proof of the theorem. Recall for convenience the function $F: \mathbb{R}^{(K+1)p} \rightarrow \mathbb{R}$ defined in~\eqref{def:F}, given by $$F(\bDelta)=\mL\left(\btheta^{*}+\bDelta\right)-\mL\left(\btheta^{*}\right)+\lambda_{0}\mR \left(\btheta^{*}+\bDelta\right)-\lambda_{0}\mR \left(\btheta^{*}\right),$$
 where $\btheta^{*}$ is the transformed model parameter defined in (\ref{transrule}), and $\bDelta=\left(\left(\bDelta^{(1)}\right)^{\top}, \ldots, \left(\bDelta^{(K)}\right)^{\top}, \left(\bDelta^{(0)}\right)^{\top}\right)^{\top} \in \mathbb{R}^{(K+1)p}$.

By Lemma \ref{concentration}, the triangle inequality, and the fact that $\left\|\btheta^{(0)}_{S_0^c}\right\|_{1} = 0$,  $\left\|\btheta^{(k)}\right\|_{1} \le h_k$ for $1 \le k \le K$, we have
$$
\begin{aligned}
F(\hat{\bDelta})
=
& \mL\left(\btheta^{*}+\hat{\bDelta}\right)-\mL\left(\btheta^{*}\right)+\lambda_{0}\mR \left(\btheta^{*}+\hat{\bDelta}\right)-\lambda_{0}\mR \left(\btheta^{*}\right) \\
\geq & -\left\|\nabla \mL\left(\btheta^{*}\right)\right\|_{\infty} \| \hat{\bDelta} \|_{1}+ \hat{\bDelta}^{\top} \nabla^2 \mL\left(\btheta^*+\phi \hat{\bDelta}\right) \hat{\bDelta} \quad(\phi \in(0,1))  \\
& +\sum_{k=1}^{K}\lk w_k\left(\left\|\btheta^{(k)}+\hat{\bDelta}^{(k)}\right\|_{1}-\left\|\btheta^{(k)}\right\|_{1}\right) + \lambda_{0}\sqrt{\sum_{k=0}^{K}\frac{n_{k}}{N}w_{k}^2} \left(\left\|\btheta^{(0)} + \hat{\bDelta}^{(0)}\right\|_1 - \left\|\btheta^{(0)}\right\|_1\right) \\
\geq 
& -\sum_{k=1}^{K} \frac{\lk w_k}{2}\left\|\hat{\bDelta}^{(k)}\right\|_{1}-\frac{\lambda_{0} \sqrt{\sum_{k=0}^{K}\frac{n_{k}}{N}w_{k}^2}}{2}\left\|\hat{\bDelta}^{(0)}\right\|_{1}+ \hat{\bDelta}^{\top} \nabla^2 \mL\left(\btheta^*+\phi \hat{\bDelta}\right) \hat{\bDelta} \\
& +\sum_{k=1}^{K}\lk w_k\left(\left\|\hat{\bDelta}^{(k)}\right\|_{1}-2\left\|\btheta^{(k)}\right\|_{1}\right) \\
& +\lambda_{0} \sqrt{\sum_{k=0}^{K}\frac{n_{k}}{N}w_{k}^2}\left(\left\|\btheta_{S_0}^{(0)}\right\|_{1}-\left\|\hat{\bDelta}_{S_0}^{(0)}\right\|_{1}+\left\|\hat{\bDelta}_{S_0^c}^{(0)}\right\|_{1}-\left\|\btheta_{S_0^c}^{(0)}\right\|_{1}-\left\|\btheta_{S_0}^{(0)}\right\|_{1}-\left\|\btheta_{S_0^c}^{(0)}\right\|_{1}\right)\\
\geq 
&  \hat{\bDelta}^{\top} \nabla^2 \mL\left(\btheta^*+\phi \hat{\bDelta}\right) \hat{\bDelta}+\frac{\lambda_{0} \sqrt{\sum_{k=0}^{K}\frac{n_{k}}{N}w_{k}^2}}{2}\left(\left\|\hat{\bDelta}_{S_0^c}^{(0)}\right\|_{1}-3\left\|\hat{\bDelta}_{S_0}^{(0)}\right\|_{1}\right) +\sum_{k=1}^{K} \frac{\lk w_k}{2}\left\|\hat{\bDelta}^{(k)}\right\|_{1}-2 \sum_{k=1}^{K}\lk w_k h_{k}.
\end{aligned}
$$
with probability larger than $1- \exp \left(-c_{1} \log p \right) -  \exp \left(-c_{2} n_T \right)$.

Applying Lemma \ref{Thm1Lemma2} we can lowerbound the quadratic term $ \hat{\bDelta}^{\top} \nabla^2 \mL\left(\btheta^*+\phi \hat{\bDelta}\right) \hat{\bDelta}$, leading to
\begin{align}
\label{RSC_Thm1}
F(\hat{\bDelta}) 
\geq& 
\left(\frac{\alpha_{\min }^{2}}{\gamma_{0}}  \sum^{K}_{k=0} \frac{n_k w_k}{N} - u_n\right) \left\|\hat{\bDelta}^{(0)}\right\|_{2}^{2}+ \frac{\alpha_{\min }^{2}}{\gamma_{0}}\sum_{k=1}^{K} \frac{n_{S} w_k}{N} \left\|\hat{\bDelta}^{(k)}\right\|_{2}^{2} -\frac{2 \alpha_{\max}}{\gamma_{0}} \sum^{K}_{k=1} \frac{n_{S}w_k}{N}\lambda_{T} \left\|\hat{\bDelta}^{(k)}\right\|_{1} - v_n  \sum_{k=1}^{K} \lk w_k h_{k} \nonumber \\
 &+ \frac{\lambda_{0} \sqrt{\sum_{k=0}^{K}\frac{n_{k}}{N}w_{k}^2}}{2}\left(\left\|\hat{\bDelta}_{S_0^c}^{(0)}\right\|_{1}-3\left\|\hat{\bDelta}_{S_0}^{(0)}\right\|_{1}\right) +\sum_{k=1}^{K} \frac{\lk w_k}{2}\left\|\hat{\bDelta}^{(k)}\right\|_{1}-2 \sum_{k=1}^{K}\lk w_k h_{k} \nonumber \\
 \geq& 
 \left(\frac{\alpha_{\min }^{2}}{\gamma_{0}}  \sum^{K}_{k=0} \frac{n_k w_k}{N} - u_n\right) \left\|\hat{\bDelta}^{(0)}\right\|_{2}^{2}- \frac{3\sqrt{s}\lambda_{0} \sqrt{\sum_{k=0}^{K}\frac{n_{k}}{N}w_{k}^2}}{2}\left\|\hat{\bDelta}^{(0)}\right\|_{2}  \nonumber \\
  &+ \left( \sum_{k=1}^{K} \frac{\lk w_k}{2} - \frac{2 \alpha_{\max}}{\gamma_{0}} \sum^{K}_{k=1} \frac{n_{S}w_k}{N}\lambda_{T}\right) \left\|\hat{\bDelta}^{(k)}\right\|_{1} - (2+v_n)  \sum_{k=1}^{K} \lk w_k h_{k}. 
\end{align}
probability larger than $1- \exp \left(-c_{1} \log p \right) -  \exp \left(-c_{2} n_T \right)$.

Recall that we select $\lk  = c_{1} \frac{n_S}{N} \sqrt{\frac{\log p}{n_T}}$ and $\lambda_T = c_{2} \sqrt{\frac{\log p}{n_T}}$, therefore, with a proper choice of the constants $c_{1}$ and $c_{2}$, we have $\lk \geq 4 \frac{\alpha_{\max}}{\gamma_{0}} \frac{n_S}{N} \lambda_{T}$. 
In addition, notice that $\hat{\btheta}$ is the solution to the problem (\ref{transobj}) and $\btheta^*$ is feasible (cf. Lemma \ref{Thm1Lemma1}), we have $F(\hat{\bDelta}) \leq 0$. These results, combining with (\ref{RSC_Thm1}), lead to
\begin{align}
\label{Delta0Quadratic}
0 \geq \left(\frac{\alpha_{\min }^{2}}{\gamma_{0}}  \sum^{K}_{k=0} \frac{n_k w_k}{N} - u_n\right) \left\|\hat{\bDelta}^{(0)}\right\|_{2}^{2}- \frac{3\sqrt{s}\lambda_{0} \sqrt{\sum_{k=0}^{K}\frac{n_{k}}{N}w_{k}^2}}{2}\left\|\hat{\bDelta}^{(0)}\right\|_{2}  - (2+v_n)  \sum_{k=1}^{K} \lk w_k h_{k}, 
\end{align}
which is an  inequality quadratic in $\|\hat{\bDelta}^{(0)}\|_2$. 

To establish the  convergence rate of $\|\hat{\bDelta}^{(0)}\|_2$, it remains to find out the order of $u_n$ and $v_n$. We now discuss by cases based on the order of $\lz$. Recall the parameter choices given in the theorem:
\begin{align*}
    \lambda_{0} = c_{0} \left[\Big( \sum_{k=0}^{K}\frac{n_{k}}{N}w_{k}^2\Big)^{-1/2} \Big(\frac{\bar{h}(\boldsymbol{w})^2 \log p}{s^2 n_{T}} \Big)^{1/4} + \Big(\frac{\log p}{N}\Big)^{1/2} \right], \quad \lambda_{1} = c_{0} \frac{n_{S}}{N} \Big(\frac{\log p}{n_{T}} \Big)^{1/2}, \quad \text{and} \quad \lambda_T = c_{1} \Big(\frac{\log p}{n_{T}} \Big)^{1/2}.
\end{align*}

\subsubsection{Case 1:}\label{sec:case_study}
If $\frac{s \log p}{N} \lesssim \Big( \sum_{k=0}^{K}\frac{n_{k}}{N}w_{k}^2\Big)^{-1} \bar{h}(\boldsymbol{w})\sqrt{\frac{ \log p}{n_{T}} }$, then we have $\lz \asymp \Big( \sum_{k=0}^{K}\frac{n_{k}}{N}w_{k}^2\Big)^{-1/2} \left(\frac{\bar{h}(\boldsymbol{w})^2 \log p}{s^2 n_T}\right)^{1 / 4}$, $\lk \asymp \frac{n_S}{N} \sqrt{\frac{\log p}{n_T}}$ and $\lambda_{T} \asymp\sqrt{\frac{\log p}{n_T}}$. Recalling that $n_k = n_S$ for $k \in [K]$, we consider the following two cases:

\paragraph{Case 1.1:} If $\lk^2 \underline{w} \lesssim  (\lz^2 \sum_{k = 0}^K \frac{n_k}{N} w_k^2)/ ( (n_T/n_S) w_0 + \sum_{k = 1}^K  w_k)$, then we have
$$
u_{n} \lesssim \frac{1}{K} \frac{\log p}{n_{T}} \frac{\lz^2 s (\sum_{k=0}^{K}\frac{n_{k}}{N}w_{k}^2)}{\lk^2} \lesssim K\bar{h}(\boldsymbol{w}) \sqrt{\frac{\log p}{n_T}} = o(1),
$$  
where the first inequality is based on the assumption that $\underline{w}$ is bounded away from $0$ and and $(\beta_{\max} \vee \tau_0) / \gamma_{0}$ is bounded above, while for the last equality we uses the assumption that $K\bar{h}(\boldsymbol{w}) \sqrt{\frac{\log p}{n_T}} = o(1)$.  Similarly, we can establish that 
$$
v_n \lesssim \frac{1}{K} \frac{\log p}{n_{T}} \frac{\left( \sum_{k=1}^{K} \lk w_k h_{k}\right)}{\lk^2} \lesssim \sqrt{\frac{\log p}{n_{T}}} \sum_{k=1}^{K} w_k h_k  \lesssim K\bar{h}(\boldsymbol{w}) \sqrt{\frac{\log p}{n_T}}= o(1).
$$
Therefore, in this case, solving the inequality (\ref{Delta0Quadratic}) yields
\begin{align}
\label{case1_bound}
\left\| \hat{\bDelta}^{(0)} \right\|^2_2 & \lesssim \frac{1}{ \left(\sum^{K}_{k=0} \frac{n_k w_k}{N} \right)^2 }   \bar{h}(\boldsymbol{w})\sqrt{\frac{\log p}{n_T}}  +  \frac{1}{ \sum^{K}_{k=0} \frac{n_k w_k}{N}  } \sum^{K}_{k=1} \frac{n_S w_k}{N} h_k \sqrt{\frac{\log p}{n_{T}}}.
\end{align}

\paragraph{Case 1.2:} If $\lk^2 \underline{w} \gtrsim  (\lz^2 \sum_{k = 0}^K \frac{n_S}{N} w_k^2)/ ( (n_T/n_S) w_0 + \sum_{k = 1}^K  w_k)$, we can similarly establish that  
$$
u_{n} \lesssim \frac{1}{K} \frac{\log p}{n_{T}} s ( (n_T/n_S) w_0 + \sum_{k = 1}^K  w_k) \lesssim \frac{s \log p}{n_{T}} = o(1)
$$  
where the last inequality follows from the assumptions that $\frac{s \log p}{n_{T}} = o(1)$. In this case, noting that 
$$
\frac{\left( \sum_{k=1}^{K} \lk w_k h_{k}\right)}{ \left( 1 / ( (n_T/n_S) w_0 + \sum_{k = 1}^K  w_k)\right) } \lesssim K \bar{h}(\boldsymbol{w}) \sqrt{\frac{\log p}{n_T}},
$$ thereby we have
$$
v_n \lesssim  \frac{1}{K}\frac{\log p}{n_{T}} \frac{\left( \sum_{k=1}^{K} \lk w_k h_{k}\right)}{ \left(\lz^2 \sum_{k=0}^{K}\frac{n_{k}}{N}w_{k}^2 / ( (n_T/n_S) w_0 + \sum_{k = 1}^K  w_k)\right) } \lesssim \frac{s \log p}{n_T} = o(1).
$$
Hence, the bound in (\ref{case1_bound}) still holds.

\subsubsection{Case 2:}
 If $\frac{s \log p}{N} \gtrsim \Big( \sum_{k=0}^{K}\frac{n_{k}}{N}w_{k}^2\Big)^{-1} \bar{h}(\boldsymbol{w})\sqrt{\frac{ \log p}{n_{T}} }$, then we have $\lz \asymp \sqrt{\frac{\log p}{N}}$, $\lk\asymp \frac{n_S}{N} \sqrt{\frac{\log p}{n_T}}$ and $\lambda_{T} \asymp\sqrt{\frac{\log p}{n_T}}$. In this case, we have $\lk^2 \gtrsim \frac{\lz^2}{K} \gtrsim \left(\lz^2 \sum_{k=0}^{K}\frac{n_{k}}{N}w_{k}^2 / ( (n_T/n_S) w_0 + \sum_{k = 1}^K  w_k)\right)$ as $n_S > n_T$. So following the discussion in the first case, we have
$$
u_{n} \lesssim \frac{1}{K} \frac{\log p}{n_{T}} s ( (n_T/n_S) w_0 + \sum_{k = 1}^K  w_k) \lesssim \frac{s \log p}{n_{T}} = o(1).
$$  
Further notice that in this case, it holds that 
$$
\sum_{k=1}^{K} \lk w_k h_{k} \lesssim \bar{h}(\boldsymbol{w})\sqrt{\frac{\log p}{n_{T}}} \lesssim \Big( \sum_{k=0}^{K}\frac{n_{k}}{N}w_{k}^2\Big) \frac{s \log p}{N}.
$$
Therefore, we further obtain
$$
v_n \lesssim  \frac{1}{K}\frac{\log p}{n_{T}} \frac{\left( \sum_{k=1}^{K} \lk w_k h_{k}\right)}{ \left(\lz^2 \sum_{k=0}^{K}\frac{n_{k}}{N}w_{k}^2 / ( (n_T/n_S) w_0 + \sum_{k = 1}^K  w_k)\right) } \lesssim \frac{1}{K} \frac{\log p}{n_{T}} s \left( (n_T/n_S) w_0 + \sum_{k = 1}^K  w_k \right) \lesssim \frac{s \log p}{n_T}= o(1).
$$
So in this case plugging in the choice of $\lz$ and $\lk$, the $\ell_2$ error bound for $\hat{\bDelta}^{(0)}$ becomes
\begin{align}
\label{case2_bound}
\left\| \hat{\bDelta}^{(0)} \right\|^2_2 & \lesssim \frac{1}{ \left(\sum^{K}_{k=0} \frac{n_k w_k}{N}\right)^2 }  (\sum_{k=0}^{K}\frac{n_{k}w_{k}^2}{N}) \frac{s \log p}{N} +  \frac{1}{\sum^{K}_{k=0} \frac{n_k w_k}{N} } \bar{h}(\boldsymbol{w}) \sqrt{\frac{\log p}{n_{T}}} 
\end{align}
In summary, by combining the results from the two cases discussed above and applying the constraint $\sum^{K}_{k=0}\frac{n_K w_k}{N} = 1$, we have
\begin{align*}
\left\| \hat{\bDelta}^{(0)} \right\|^2_2 \lesssim \Big(\sum_{k=0}^{K}\frac{n_{k}w_{k}^2}{N}\Big) \frac{s \log p}{N} +  \bar{h}(\boldsymbol{w}) \sqrt{\frac{\log p}{n_{T}}} 
\end{align*}
probability larger than $1- \exp \left(-c_{1} \log p \right)$ as $n_T \gtrsim \log p$. The proof is then finished.

\subsection{Proof of Corollary \ref{choice_of_w}}
The proof follows a similar arguments as those in the Section 3 of \cite{duchi2008efficient}. According to Theorem \ref{onesteprate}, the optimal weights can be formally described as the solution to the following problem:
\begin{subequations}
\label{weight_prob}
\begin{align}
&\min _{\bw^{\prime}}\left\{\sum_{k=0}^{K}\left(w_{k}^{\prime}\right)^{2}  \frac{s \log p}{n_{k}}+c_{\bSig} \sum_{k=1}^{K} w_{k}^{\prime} h_{k} \sqrt{\frac{\log p}{n_{T}}}\right\} \label{weight_prob:obj} \\
&\text{s.t. } \sum_{k=0}^{K} w_{k}^{\prime}=1, \  w_{k}^{\prime} \geqslant 0,\ k=0, \cdots, K. \label{weight_prob:constr}
\end{align}
\end{subequations}
where recall that we define $w_{k}^{\prime}=\frac{n_k}{N} w_{k}$ with $n_k=n_S$ for $k \in [K]$ and $n_k = n_T$ for $k=0$.

The corresponding Lagrangian function of problem (\ref{weight_prob}) is
$$
\mathcal{L}\left(\bw^{\prime}, \boldsymbol{\eta}\right)=\sum_{k=0}^{K}\left(w_{k}^{\prime}\right)^{2} s \frac{\log p}{n_{k}}+c_{\bSig} \sum_{k=1}^{K} w_{k}^{\prime} h_{k} \sqrt{\frac{{\log p}}{n_{T}}}+\xi\left(1-\sum_{k=1}^{K} w_{k}^{\prime}\right)-\sum_{k=1}^{K} \eta_{k} w_{k}^{\prime}
$$
where $\boldsymbol{\eta} \in \mathbb{R}_{+}^{K}$ and $\xi \in \mathbb{R}$ are Lagrange multipliers. According to the KKT conditions, we have
\begin{align}
& \frac{\partial \mathcal{L}\left(\bw^{\prime}, \boldsymbol{\eta}\right)}{\partial w_{k}^{\prime}}=2 s \frac{\log p}{n_{S}} w_{k}^{\prime}+c_{\bSig} h_{k} \sqrt{\frac{\log p}{n_{T}}}-\xi-\eta_{k}=0,  \ k \in[K] \label{weight_prob:KKT1}\\
& \frac{\partial \mathcal{L}\left(\bw^{\prime}, \boldsymbol{\eta}\right)}{\partial w_{0}^{\prime}}=2 s \frac{\log p}{n_{T}} w_{0}^{\prime}-\xi-\eta_{0}=0 \label{weight_prob:KKT2}
\end{align}
and $\eta_{k} \cdot w_{k}^{\prime}=0$ for $k=0, \cdots, K$. Notice for any source task that is assigned a non-zero weight $w_{k}$, we must have $\eta_{k}=0$. Therefore, the non-zero weights are tied to the single variable $\xi$. To find $\xi$, we use the following lemma, with its proof deferred to Section \ref{lemmaproof}.
\begin{lemma}
    \label{lemma:sort}
     Let $w_{0}^{\prime}, \cdots, w_{K}^{\prime}$ be the optimal solution to the problem (\ref{weight_prob}), then
     \begin{itemize}
         \item  For any $j, l \in[K]$ such that $h_{l}>h_{j}$, $w_{j}^{\prime}=0$ implies $w_{l}^{\prime}=0$.
         \item For any choice of $\boldsymbol{h} \in \mathbb{R}_{+}^{K}$, it always holds that $w_{0}^{\prime}>0$.
     \end{itemize}
\end{lemma}

Lemma \ref{lemma:sort} provides a way to determine non-zero weights based on $\boldsymbol{h}$. Denote $\rho=\#\left\{k \in[K]: w_{k}>0\right\}$ as the number of sources with non-zero weights and $h_{(1)} \leqslant h_{(2)} \leqslant \cdots \leqslant h_{(K)}$ as the sequence obtained by sorting the elements in $\boldsymbol{h}$ in an ascending order, then Lemma \ref{lemma:sort} combined with (\ref{weight_prob:KKT1}) and (\ref{weight_prob:KKT2}) implies
$$
\begin{aligned}
\sum_{k=1}^{K} \frac{2 s \log p}{n_{S}} w_{k}^{\prime} =\sum_{k=1}^{\rho}\left(\xi-c_{\bSig} h_{(k)} \sqrt{\frac{\log p}{n_{T}}}\right) \ \text{and} \
\frac{2 s \log p}{n_{T}} w_{0}^{\prime}  =\xi
\end{aligned}
$$
which together with (\ref{weight_prob:constr}) indicates
\begin{align}
\label{weight:xi}
\xi=\frac{2 s \log p+n_{S} \sum_{k=1}^{\rho} c_{\bSig} h_{(k)} \sqrt{\frac{\log p}{n_T}}}{\rho n_{S}+n_{T}}
\end{align}

Plugging the result back into (\ref{weight_prob:KKT1}) and (\ref{weight_prob:KKT2}) yields the desired result.

It remains to establish the bound (\ref{onesteporacleFusion}). We prove by discussing all the cases. In the case where \( s \frac{\log p}{n_T} \leq \frac{c_{\bSig}}{2} h_{1} \sqrt{\frac{\log p}{n_T}} \), according to (\ref{weight:K=1}), we have \( w_1 = 0 \). Under this scenario, the upper bound specified in (\ref{onestepFusion}) simplifies to 
\( \frac{c_{\bSig}}{2}h_{1}\sqrt{\frac{\log p}{n_{T}}} \wedge \frac{s \log p}{n_T}. \)
Conversely, if \( s \frac{\log p}{n_T} \geq \frac{c_{\bSig}}{2} h_{1} \frac{\log p}{n_T} \), substituting the expression for \( w_1 \) into the upper bound in (\ref{onestepFusion}) results in a bound of 
\( s\frac{\log p}{N} + \frac{n_S}{N} \left( c_{\bSig}  h_{1}\sqrt{\frac{\log p}{n_T}} \wedge 2 s \frac{\log p}{n_T} \right). \)
Combining the results from both cases leads to the bound in (\ref{onesteporacleFusion}).

\section{Proof of lemmas}
\label{lemmaproof}
\subsection{Proof of Lemma \ref{concentration_for_oracle}}
We begin by establishing the result in (\ref{beta0_concentration}). According to the definition of $\mathcal{L}^{(k)}, k=0,1, \dots, K$ and the oracle estimator, it holds that
\begin{align}
\label{score_func_expression}
& \max _{j \in S_{0}^{c}}\left|\frac{n_T}{N} \nabla_{j} \mL^{(0)}\left(\hbzora\right)+\sum_{k=1}^{K} \frac{n_{S}}{N} \nabla_{j} \mL^{(k)}\left(\hbzora+\hbdkora\right)\right|\nonumber \\
= & \left\|\frac{2}{N}\left(\bX_{S_{0}^{c}}^{(0)}\right)^{\top}\left(\by^{(0)}-\bX_{S_{0}}^{(0)} \hbzoraS\right)+\sum_{k=1}^{K} \frac{2}{N}\left(\bX_{S_{0}^{c}}^{(k)}\right)^{\top}\left(\by^{(k)}-\bX_{S_{0}}^{(k)} \hbzoraS-\bX_{S_{k}}^{(k)} \hbdkoraS \right)\right\|_{\infty}
\end{align}

Recalling the expression of $\hbdkoraS$ outlined in (\ref{delta_oracle}), this expression together with the property of the projection matrix $\Hk$ leads to
$$
\begin{aligned}
 \bX_{S_{k}}^{(k)} \hbdkoraS&=\bX_{S_{k}}^{(k)}\left[\left(\bX_{S_{k}}^{(k)}\right)^{\top} \bX_{S_{k}}^{(k)}\right]^{-1}\left(\bX_{S_{k}}^{(k)}\right)^{\top}\left[\by^{(k)}-\bX_{S_{0}}^{(k)}\hbzoraS\right] \\
& =\Hk\left[\bX_{S_{k}}^{(k)} \bd_{S_{k}}^{(k)}+\bX_{S_{0}}^{(k)} \bb_{S_{0}}^{(0)}+\bek\right]-\Hk\bX_{S_{0}}^{(k)} \hbzoraS \\
& =\bX_{S_{k}}^{(k)} \bd_{S_{k}}^{(k)}+\Hk \bX_{S_{0}}^{(k)}\left(\bb_{S_{0}}^{(0)}-\hbzoraS\right)+\Hk \bek \\
& =\bX_{S_{k}}^{(k)} \bd_{S_{k}}^{(k)}+\bX_{S_{0}}^{(k)}\left(\bb_{S_{0}}^{(0)}-\hbzoraS\right)+\tbX_{S_{0}}^{(k)}\left(\hbzoraS-\bb_{S_{0}}^{(0)}\right)+\Hk \bek
\end{aligned}
$$
where recall that we define $\tbXk_{S_{0}} = (\boldsymbol{I} - \Hk) \bXk_{S_0}$. Substituting the result into (\ref{score_func_expression}) yields
$$
\begin{aligned}
& \max _{j \in S_{0}^{c}}\left|\frac{n_T}{N} \nabla_{j} \mL^{(0)}\left(\hbzora\right)+\sum_{k=1}^{K} \frac{n_{S}}{N} \nabla_{j} \mL^{(k)}\left(\hbzora+\hbdkora\right)\right| \\
=&\left\| \frac{2}{N}\left(\bX_{S_{0}^c}^{(0)}\right)^{\top}\left(\bX_{S_{0}}^{(0)}\left(\bb_{S_{0}}^{(0)}-\hbzoraS\right)+\bez\right)+\sum_{k=1}^{K} \frac{2}{N}\left(\bX_{S_{0}^{c}}^{(k)}\right)^{\top}\left(\tbX_{S_{0}}^{(k)}\left(\bb_{S_{0}}^{(0)}-\hbzoraS\right)+\left(\boldsymbol{I}-\Hk\right) \bek\right) \right\|_{\infty} \\
=&\left\|\frac{2}{N} \bX_{S_{0}^{c}}^{\top}\left[\tbX_{S_{0}}\left(\bb_{S_{0}}^{(0)}-\hbzoraS\right)+\tilde{\boldsymbol{\epsilon}}\right]\right\|_{\infty}
\end{aligned}
$$
where we used the shorthand $\bX_{S_0^c} =((\bXz_{S_0^c})^\top, ({\bX}_{S_0^c}^{(1)})^\top, \dots, ({\bX}_{S_0^c}^{(K)})^\top)^\top$, $\tbX_{S_0} =((\bXz_{S_0})^\top, (\tilde{\bX}_{S_0}^{(1)})^\top, \dots, (\tilde{\bX}_{S_0}^{(K)})^\top)^\top$ and $\tilde{\boldsymbol{\epsilon}} =((\bez)^\top, (\tilde{\boldsymbol{\epsilon}}^{(1)})^\top, \dots, (\tilde{\boldsymbol{\epsilon}}^{(K)})^\top)^\top$. 

Plugging the result in (\ref{beta_oracle}) for $\hbzoraS$ further gives
$$
\begin{aligned}
 \frac{2}{N} \bX_{S_{0}^{c}}^{\top}\left[\tbX_{S_{0}}\left(\bb_{S_{0}}^{(0)}-\hbzoraS\right)+\tilde{\boldsymbol{\epsilon}}\right] &=  \frac{2}{N} \bX_{S_{0}^{c}}^{\top}\left[\boldsymbol{I}-\tbX_{S_{0}}\left(\tbX_{S_{0}}^{\top} \tbX_{S_{0}}\right)^{-1} \tbX_{S_{0}}^{\top}\right] \tilde{\boldsymbol{\epsilon}}
\end{aligned}
$$
Notice that for $k=0,\dots,K$, $\left(\boldsymbol{I}-\Hk\right)$ is a projection matrix with rank $p-|S_k| \le p$ and $\bX_{i \cdot}^{(k)}$'s and $\bek_{i}$'s are independent random vectors/variables. Therefore, first applying an orthogonal transformation and then following the arguments similar to those in the proof of Lemma \ref{concentration}, we have
\begin{align*}
&\max _{j \in S_{0}^{c}}\left|\frac{n_T}{N} \nabla_{j} \mL^{(0)}\left(\hbzora\right)+\sum_{k=1}^{K} \frac{n_{k}}{N} \nabla_{j} \mL^{(k)}\left(\hbzora+\hbdkora\right)\right| \\
&= 
 \left\|\frac{2}{N} \bX_{S_{0}^{c}}^{\top}\left[\boldsymbol{I}-\tbX_{S_{0}}\left(\tbX_{S_{0}}^{\top} \tbX_{S_{0}}\right)^{-1} \tbX_{S_{0}}^{\top}\right] \tilde{\boldsymbol{\epsilon}}\right\|_{\infty}
\lesssim \sqrt{\frac{\log p}{N}}
\end{align*}
with probability larger than  $1-c_{1} \exp \left(-c_{2} \log p \right)$, as claimed.

Next, we prove the result in (\ref{deltak_concentration}). Following a similar argument, we can show that
$$
\begin{aligned}
\max _{j \in S_{k}^{c}}\left|\frac{n_{k}}{N} \nabla_j \mathcal{L}^{(k)}\left(\hbzora+\hbdkora\right)\right| & =\left\|\frac{2}{N} \left(\bX_{S_{k}^{c}}^{(k)}\right)^{\top}\left(\by^{(k)}-\bX_{S_{0}}^{(k)} \hbzoraS-\bX_{S_{k}}^{(k)} \hbdkoraS\right)\right\|_{\infty} \\
& =\left\|\frac{2}{N} \left(\bX_{S_{k}^{c}}^{(k)}\right)^{\top}\left(\tilde{\bX}_{S_{0}}^{(k)}\left(\bb_{S_{0}}^{(0)}-\hbzoraS\right)+\left(\boldsymbol{I}-\Hk\right) \bek\right)\right\|_{\infty} \\
& = \frac{2}{N}\left\|\left(\bX_{S_{k}^{c}}^{(k)}\right)^{\top}\left[-\tilde{\bX}_{S_{0}}^{(k)}\left(\tilde{\bX}_{S_{0}}^{\top} \tbX_{S_{0}}\right)^{-1} \tbX_{S_{0}}^{\top} \tilde{\boldsymbol{\epsilon}}+\tilde{\boldsymbol{\epsilon}}^{(k)}\right]\right\|_{\infty} \\
& = \frac{2n_S}{N}\left\|\frac{1}{n_S}\left(\bX_{S_{k}^{c}}^{(k)}\right)^{\top}\left[-\tilde{\bX}_{S_{0}}^{(k)}\left(\tilde{\bX}_{S_{0}}^{\top} \tbX_{S_{0}}\right)^{-1} \tbX_{S_{0}}^{\top} \tilde{\boldsymbol{\epsilon}}+\tilde{\boldsymbol{\epsilon}}^{(k)}\right]\right\|_{\infty} \\
& \lesssim \frac{n_{S}}{N} \sqrt{\frac{\log p}{n_{S}}}
\end{aligned}
$$
with probability larger than  $1-c_{1} \exp \left(-c_{2} \log p \right)$, which finishes the proof.

\subsection{Proof of Lemma \ref{concentration}} By definition, 
$$-\nabla \mL\left(\btheta\right)=\left(\left( \frac{1}{N} w_{1}\left(\bX^{(1)}\right)^{\top} \boldsymbol{\epsilon}^{(1)}\right)^{\top}, \dots, \left( \frac{1}{N}w_{K} \left(\bX^{(K)}\right)^{\top} \boldsymbol{\epsilon}^{(K)}\right)^{\top}, \left( \frac{1}{N} \sum_{k=0}^{K}w_{k}\left(\bX^{(k)}\right)^{\top} \bek\right)^{\top}\right)^{\top}.$$

Therefore, by the Holder's inequality, we have
$$
\begin{aligned}
\left|\left\langle\nabla \mL\left(\btheta\right), \bDelta\right\rangle\right| &\leq \sum_{k=1}^{K} \left| \left\langle\frac{1}{N} w_{k}\left(\bX^{(k)}\right)^{\top} \bek, \bDelta^{(k)}\right\rangle \right| +\left|\left\langle\frac{1}{N}\sum_{k=0}^{K}  w_{k}\left(\bX^{(k)}\right)^{\top} \bek, \bDelta^{(0)}\right\rangle \right| \\
& \leq \sum_{k=1}^{K}\left\|\frac{1}{N} w_{k}\left(\bX^{(k)}\right)^{\top} \bek\right\|_{\infty}\left\|\bDelta^{(k)}\right\|_{1}+\left\|\frac{1}{N}\sum_{k=0}^{K}w_{k}  \left(\bX^{(k)}\right)^{\top} \bek\right\|_{\infty}\left\|\bDelta^{(0)}\right\|_{1} .
\end{aligned}
$$

Recall that we define $n_{k} = n_{S}$ for $1 \le k \le K$ and $n_0 = n_T$. Define
$$
\mathcal{A}^{(k)}=\left\{\max _{\substack{1 \leq j \leq p}}\left\{\frac{1}{n_{k}} \sum_{i=1}^{n_{k}}\left(x_{i j}^{(k)}\right)^{2}\right\} \leq 2\max_{1 \le j \le p}  E\left(x_{i j}^{(k)}\right)^{2}\right\}, $$
and
$$
\mathcal{A}=\left\{\max _{\substack{1 \leq j \leq p}}\left\{\frac{1}{N} \sum^{K}_{k=0}\sum_{i=1}^{n_{k}}\left(x_{i j}^{(k)}\right)^{2}\right\} \leq 2 \max_{1 \le k \le K, 1 \le j \le p}  E\left(x_{i j}^{(k)}\right)^{2}\right\}. $$
Since  $\bX^{(k)}$ is sub-Gaussian with uniformly bounded second moment and $n_{S} \gtrsim \log p$, we have $P\left(\overline{\mathcal{A}^{(k)}}\right) \leq c_{1} \exp (-c_{2} n_{S})$ for $1 \le k \le K$ and $P\left(\overline{\mathcal{A}}\right) \leq c_{1} \exp (-c_{2} N)$ for some universal constants $c_{1}$ and $c_{2}$.

In addition, as $\bek \sim N\left(0, \sigma_{k}^{2}\boldsymbol{I}\right)$ for some finite $\sigma_{k}$, by Proposition 5.10 in \citep{vershynin2010introduction}, we can establish that with some universal constant $c_{3}$, for $1 \le k \le K$,
\begin{align*}
P\left(\max _{1 \leq j \leq p}\left|\frac{1}{n_{S}} \sum_{i=1}^{n_{S}}w_{k} \epsilon_{i}^{(k)} x_{i j}^{(k)}\right| \geq t\right) 
&\leq P\left(\max _{1 \leq j \leq p}\left|\frac{1}{n_{S}} \sum_{i=1}^{n_{S}} w_{k}\epsilon_{i}^{(k)} x_{i j}^{(k)}\right| \geq t  \ \middle| \  \mathcal{A}^{(k)}\right) + P(\overline{\mathcal{A}^{(k)}}) \\
&\leq p \cdot e \cdot \exp \left(-\frac{c_3 n_{S} t^{2}}{4 \sigma_{k}^2 w_{k}^2 \max_{1 \le j \le p} E\left(x_{i j}^{(k)}\right)^{2}}\right) + c_{1}\exp{(-c_{2}n_{S})}.
\end{align*}

Since by Assumption \ref{A1}, there exist a constant $c$ such that $\max_{1 \leq k \leq K}\Lambda_{\max}(\Sigk) \leq c$, so $\max_{1 \le j \le p} E\left(x_{i j}^{(k)}\right)^{2}$ is uniformly bounded above. Therefore for $1 \le k \le K$, by choosing $t = \sqrt{c_{4} \log p / n_{S}}$ for some constant $c_{4}$,  with probability larger than $1-c_{1} \exp \left(-c_{2} \log p\right)$, we have
$$
\left\|\frac{1}{N} w_{k}\left(\bX^{(k)}\right)^{\top} \bek\right\|_{\infty} \lesssim w_{k}\frac{n_{S}}{N} \sqrt{\frac{\log p}{n_{S}}} = w_{k}\sqrt{\frac{n_{S}}{N}} \sqrt{\frac{\log p}{N}} .
$$

Similarly, we have
\begin{align*}
P\left(\max _{1 \leq j \leq p}\left|\frac{1}{N} \sum_{k=0}^{K} \sum_{i=1}^{n_{k}} w_{k}\epsilon_{i}^{(k)} x_{i j}^{(k)}\right| \geq t \right) 
&\leq P\left(\max _{1 \leq j \leq p}\left|\frac{1}{N} \sum_{k=0}^{K} \sum_{i=1}^{n_{k}} w_{k}\epsilon_{i}^{(k)} x_{i j}^{(k)}\right| \geq t  \ \middle| \ \mathcal{A}\right) + P(\overline{\mathcal{A}})  \\
&\leq p \cdot e \cdot \exp \left(-\frac{c_{4} N t^{2}}{4  \sum_{k=0}^{K}\frac{n_{k}}{N}w_{k}^2 \max _ {0 \le k \le K, 1 \le j \le p}\sigma_{k}^2E\left(x_{ij}^{(k)}\right)^{2}}\right) + c_{1} \exp \left(-c_{2} N\right).
\end{align*}

So we have with probability larger than $1-c_{1} \exp \left(-c_{2} \log p \right)$,
$$
\left\| \frac{1}{N}\sum_{k=0}^{K}  w_k \left(\bX^{(k)}\right)^{\top} \bek\right\|_{\infty}
\lesssim \sqrt{\sum_{k=0}^{K}\frac{n_{k}}{N}w_{k}^2} \sqrt{\frac{\log p}{N}}.
$$

Therefore, by choosing $\lk \gtrsim c_{0} 
 \sqrt{\frac{n_{S}}{N}} \sqrt{\frac{\log p}{N}}$ and $\lambda_{0}\gtrsim c_{0} \sqrt{\frac{\log p}{N}}$ for some sufficiently large constant $c_{0}$, we have the desired result.

\subsection{Proof of Lemma \ref{Thm1Lemma1}} 
We define the optimal value gap function $F: \mathbb{R}^{(K+1)p} \rightarrow \mathbb{R}$ as
\begin{align}\label{def:F}
    F(\bDelta)=\mL\left(\btheta^{*}+\bDelta\right)-\mL\left(\btheta^{*}\right)+\lz \mR \left(\btheta^{*}+\bDelta\right)- \lz \mR \left(\btheta^{*}\right),
\end{align}
and $\hat{\btheta}$ as the solution to the problem (\ref{transobj}). We then have $\hat{\bDelta}=\hat{\btheta} - \btheta^{*} = \underset{\bDelta}{\operatorname{argmin}}  F(\bDelta)$ and $F(0)=0$. By the proof of Lemma \ref{concentration}, we have that under Assumption \ref{A1} and \ref{A2}, $\frac{1}{n_T}\|(\bXz)^\top (\byz - \bXz \btheta^{(0)})\|_{\infty} \lesssim \sqrt{\frac{\log p}{n_T}}$ with probability greater than $1-c_{1} \exp \left(-c_{2} \log p \right)$, therefore, by choosing $\lambda_{T} = c_0 \sqrt{\frac{\log p}{n_T}}$ for some sufficiently large $c_0$, $\btheta^*$ is a feasible solution of problem (\ref{P:sample-transfer}).
Consequently, we conclude that $F(\hat{\bDelta}) \leq 0$ and both $\hat{\btheta}$ and $\btheta^*$ are feasible solutions to problem (\ref{transobj}).

Since $\mL$ is a convex function, by Lemma \ref{concentration}, we can choose $\lk \gtrsim c_0 \sqrt{\frac{n_{S}}{N}}\sqrt{\frac{\log p}{N}}$ and $\lambda_{0} \gtrsim c_0 \sqrt{\frac{\log p}{N}}$ so that 
\begin{align}
\label{Convexity}
\mL\left(\btheta^{*}+\hat{\bDelta}\right)-\mL\left(\btheta^{*}\right) \geq \left\langle\nabla \mL\left(\btheta^{*}\right), \hat{\bDelta}\right\rangle
&\geq -\sum_{k=1}^{K} \frac{\lk}{2} w_k \left\|\hat{\bDelta}^{(k)}\right\|_{1}-\frac{\lambda_{0}}{2} \sqrt{\sum_{k=0}^{K}\frac{n_{k}}{N}w_{k}^2}  \left\|\hat{\bDelta}^{(0)}\right\|_{1}
\end{align}
with probability larger than $1 - c_{1} \exp(c_2 \log p)$.

Since the $\ell_1$-norm function is decomposable and $\|\btheta_{S_0^c}^{(0)}\|_1 = 0$, by triangle inequality we have

\begin{align}
\label{RestrictedCone}
 & \ \lz \mR \left(\btheta^{*}+\hbDelta\right)- \lz \mR \left(\btheta^{*}\right)\\
=& \sum_{k=1}^{K} \lk w_{k} \left(\left\|\btheta^{(k)}+\hat{\bDelta}^{(k)}\right\|_{1}-\left\|\btheta^{(k)}\right\|_{1}\right)+\lambda_{0} \sqrt{\sum_{k=0}^{K}\frac{n_{k}}{N}w_{k}^2} \left(\left\|\btheta^{(0)}+\hat{\bDelta}^{(0)}\right\|_{1}-\left\|\btheta^{(0)}\right\|_{1}\right) \nonumber \\
\geq & \sum_{k=1}^{K}\lk w_k \left(\left\|\hat{\bDelta}^{(k)}\right\|_{1}-2\left\|\btheta^{(k)}\right\|_{1}\right)+\lambda_{0}\sqrt{\sum_{k=0}^{K}\frac{n_{k}}{N}w_{k}^2}\left(\left\|\hat{\bDelta}_{S_0^c}^{(0)}\right\|_{1}-\left\|\hat{\bDelta}_{S_0}^{(0)}\right\|_{1}-2\left\|\btheta_{S_0^c}^{(0)}\right\|_{1}\right) \nonumber \\
\geq & \sum_{k=1}^{K}\lk w_k \left\|\hat{\bDelta}^{(k)}\right\|_{1}-2 \sum_{k=1}^{K}\lk w_k h_{k}+\lambda_{0}\sqrt{\sum_{k=0}^{K}\frac{n_{k}}{N}w_{k}^2}\left(\left\|\hat{\bDelta}_{S_0^c}^{(0)}\right\|_{1}-\left\|\hat{\bDelta}_{S_0}^{(0)}\right\|_{1}\right).\nonumber \\
\end{align}

Combining (\ref{Convexity}) with (\ref{RestrictedCone}) yields
\begin{align}
\label{beforecone}
0 \geq F(\hat{\bDelta}) \geq \sum_{k=1}^{K} \frac{\lk}{2} w_k \left\|\hat{\bDelta}^{(k)}\right\|_{1}-2 \sum_{k=1}^{K}\lk w_k h_{k}+\frac{\lambda_{0}}{2} \sqrt{\sum_{k=0}^{K}\frac{n_{k}}{N}w_{k}^2}\left(\left\|\hat{\bDelta}_{S_0^c}^{(0)}\right\|_{1}-3\left\|\hat{\bDelta}_{S_0}^{(0)}\right\|_{1}\right),
\end{align}

which leads to the following inequality:
$$
\sum_{k=1}^{K}\lk w_k \left\|\hat{\bDelta}^{(k)}\right\|_{1}+\lambda_{0}\sqrt{\sum_{k=0}^{K}\frac{n_{k}}{N}w_{k}^2}\left\|\hat{\bDelta}^{(0)}\right\|_{1} \leq 4 \lambda_{0}\sqrt{\sum_{k=0}^{K}\frac{n_{k}}{N}w_{k}^2}\left\|\hat{\bDelta}_{S_0}^{(0)}\right\|_{1}+4 \sum_{k=1}^{K}\lk w_k h_{k}.
$$

\subsection{Proof of Lemma \ref{Thm1Lemma2}}
A second order Taylor expansion of $\mathcal{L}$ implies
$$
\begin{aligned}
& \mathcal{L}\left(\btheta^{*}+\hat{\bDelta}\right)-\mathcal{L}\left(\btheta^{*}\right)-\left\langle\nabla \mathcal{L}\left(\btheta^{*}\right), \hat{\bDelta}\right\rangle \\
\geqslant & \hat{\bDelta}^{\top} \nabla^{2} \mathcal{L}\left(\btheta^{*}+\phi \hat{\bDelta}\right) \hat{\bDelta}\left(\phi \in(0,1)\right) \\
= & \sum_{k=1}^{K} \frac{n_{S}}{N} w_k \left(\hat{\bDelta}^{(k)}+\hat{\bDelta}^{(0)}\right)^{\top} \hat{\bSig}^{(k)}\left(\hat{\bDelta}^{(k)}+\hat{\bDelta}^{(0)}\right)+\frac{n_{T}}{N} w_0 \left(\hat{\bDelta}^{(0)}\right)^{\top} \hat{\bSig}^{(0)} \hat{\bDelta}^{(0)} .
\end{aligned}
$$

A further application of RSM and RSC property leads to
\begin{align}
\label{RSC_byproduct}
& \mathcal{L}\left(\btheta^{*}+\hat{\bDelta}\right)-\mathcal{L}\left(\btheta^{*}\right)-\left\langle\nabla \mL \left(\btheta^{*}\right), \hat{\bDelta}\right\rangle \nonumber \\
 \geqslant & \sum_{k=1}^{K} \frac{n_{S} \alpha_{k} w_k }{N}\left\|\hat{\bDelta}^{(k)}+\hat{\bDelta}^{(0)}\right\|_{2}^{2}+\frac{n_{T}\alpha_{0}w_0}{N}\left\|\hat{\bDelta}^{(0)}\right\|_{2}^{2}-R_{1}(\hat{\bDelta})  \nonumber\\
 \geqslant & \sum_{k=1}^{K} \frac{n_{S} \alpha_{k} w_k}{N} \cdot \frac{1}{\gamma_{0}}\left(\hat{\bDelta}^{(k)}+\hat{\bDelta}^{(0)}\right)^{\top} \hat{\bSig}^{(0)}\left(\hat{\bDelta}^{(k)}+\hat{\bDelta}^{(0)}\right)-R_{1}(\hat{\bDelta}) \nonumber \\
& +\frac{n_{T} \alpha_{0}w_0}{N}\left\|\hat{\bDelta}^{(0)}\right\|_{2}^{2}-R_{2}(\hat{\bDelta}) \nonumber \\
 \geqslant & \sum_{k=1}^{K} \frac{n_{S} \alpha_{k} w_k}{N} \cdot \frac{1}{\gamma_{0}}\left[\left(\hat{\bDelta}^{(k)}\right)^{\top} \hat{\bSig}^{(0)} \hat{\bDelta}^{(k)}+\left(\hat{\bDelta}^{(0)}\right)^{\top} \hat{\bSig}^{(0)} \hat{\bDelta}^{(0)}+2\left(\hat{\bDelta}^{(k)}\right)^{\top} \hat{\bSig}^{(0)} \hat{\bDelta}^{(0)}\right] \nonumber \\
& +\frac{n_{T}\alpha_{0} w_0}{N} \left\| \hat{\bDelta}^{(0)} \right\|_{2}^2-R_{1}(\hat{\bDelta})-R_{2}(\hat{\bDelta}),
\end{align}
where 
\begin{align}
\label{R1R2}
R_{1}(\hat{\bDelta})&:=\sum_{k=1}^{K} \frac{\beta_{k} w_k \log p}{N}\left\|\hat{\bDelta}^{(k)}+\hat{\bDelta}^{(0)}\right\|_{1}^{2}+\frac{\beta_{0}w_0 \log p}{N}\left\|\hat{\bDelta}^{(0)}\right\|_{1}^{2}  \nonumber \\
R_{2}(\hat{\bDelta})&:=\sum_{k=1}^{K} \frac{n_{S} \alpha_{k} w_k}{N} \frac{\tau_0}{\gamma_0} \frac{\log p}{n_T}\left\|\hat{\bDelta}^{(k)}+\hat{\bDelta}^{(0)}\right\|_{1}^{2}.
\end{align}
In addition, noting that $\hbz$ satisfies the constraint outlined in problem (\ref{obj}).  Therefore, using the results in Lemma \ref{concentration} we have
$$
\begin{aligned}
\left\|\hat{\bSig}^{(0)} \hat{\bDelta}^{(0)}\right\|_{\infty} & =\left\|\frac{1}{n_{T}}\left(\bX^{(0)}\right)^{\top}\bX^{(0)}\left(\hat{\bb}^{(0)}-\bb^{(0)}\right)\right\|_{\infty} \\
& =\left\| \frac{1}{n_{T}}\left(\bX^{(0)}\right)^{\top}\left(\bX^{(0)} \hat{\bb}^{(0)}-\by^{(0)}\right)-\frac{1}{n_{T}} \left(\bX^{(0)}\right)^\top \left(\bX^{(0)} \bb^{(0)}-\by^{(0)}\right) \right\|_{\infty} \\
& \leq\left\|\frac{1}{n_{T}}\left(\bX^{(0)}\right)^{T}\left(\bX^{(0)} \hat{\bb}^{(0)}-\by^{(0)}\right)\right\|_{\infty}+\left\|\frac{1}{n_{T}}\left(\bX^{(0)}\right)^{\top} \boldsymbol{\epsilon}^{(0)}\right\|_{\infty} \\
& \leq 2 \lambda_{T}.
\end{aligned}
$$
Hence, a direct application of the H\"{o}lder's inequality leads to
\begin{align}
\label{Holderfortarget}
\left( \hat{\bDelta}^{(k)}\right)^\top \hat{\bSig}^{(0)} \hat{\bDelta}^{(0)} \leq 2 \lambda_{T} \left\| \hat{\bDelta}^{(k)} \right\|_1.
\end{align}

Recall that we define $n_k = n_S$ for $k = 1, \dots, K$ and $n_k = n_T$ for $k=0$. Combining (\ref{Holderfortarget}) with (\ref{RSC_byproduct}) and applying the RSC property again, we have
$$
\begin{aligned}
& \mathcal{L}\left(\btheta^{*}+\hat{\bDelta}\right)-\mathcal{L}\left(\btheta^{*}\right)-\left\langle\nabla \mathcal{L}\left(\btheta^{*}\right), \hat{\bDelta}\right\rangle \\
\geqslant & \sum_{k=1}^{K} \frac{n_{S} \alpha_{k} w_k}{N} \cdot \frac{1}{\gamma_{0}}\left[\alpha_{0}\left\|\hat{\bDelta}^{(k)}\right\|_{2}^{2}+\alpha_{0}\left\|\hat{\bDelta}^{(0)}\right\|_{2}^{2}-2 \lambda_{T}\left\|\hat{\bDelta}^{(k)}\right\|_{1}\right] \\
& +\frac{n_{T} \alpha_{0} w_0}{N}\left\|\hat{\bDelta}^{(0)}\right\|_{2}^{2}-R_{1}(\hat{\bDelta})-R_{2}(\hat{\bDelta})-R_{3}(\hat{\bDelta}) \\
\geqslant & \frac{\alpha_{\min }^{2}}{\gamma_{0}}\left[ \sum^{K}_{k=0} \frac{n_k w_k}{N} \left\|\hat{\bDelta}^{(0)}\right\|_{2}^{2}+\sum_{k=1}^{K} \frac{n_{S} w_k}{N} \left\|\hat{\bDelta}^{(k)}\right\|_{2}^{2}\right] -\frac{2 \alpha_{\max}}{\gamma_{0}} \sum^{K}_{k=1} \frac{n_{S}w_k}{N}\lambda_{T} \left\|\hat{\bDelta}^{(k)}\right\|_{1}-\sum_{t=1}^{3} R_{t}(\hat{\bDelta}),
\end{aligned}
$$
where $R_{1}, R_{2}$ are defined in (\ref{R1R2}) and
$$
R_{3}(\hat{\bDelta}):=\sum_{k=1}^{K} \frac{n_{S} \alpha_{k} w_k}{N} \frac{\beta_{0}}{\gamma_{0}} \frac{\log p}{n_{T}}\left(\left\|\hat{\bDelta}^{(k)}\right\|_{1}^{2}+\left\|\hat{\bDelta}^{(0)}\right\|_{1}^{2}\right).
$$

It remains to bound the term
\begin{align*}
   \sum_{t=1}^{3} R_{t}(\hat{\bDelta}) 
   & = 
   \sum_{k=1}^{K} \frac{\beta_{k} w_k \log p}{N}\left\|\hat{\bDelta}^{(k)}+\hat{\bDelta}^{(0)}\right\|_{1}^{2}
   +\frac{\beta_{0}w_0 \log p}{N}\left\|\hat{\bDelta}^{(0)}\right\|_{1}^{2}
   + \sum_{k=1}^{K} \frac{n_{S} \alpha_{k} w_k}{N} \frac{\tau_0}{\gamma_0} \frac{\log p}{n_T}\left\|\hat{\bDelta}^{(k)}+\hat{\bDelta}^{(0)}\right\|_{1}^{2}\\
   & + \sum_{k=1}^{K} \frac{n_{S} \alpha_{k} w_k}{N} \frac{\beta_{0}}{\gamma_{0}} \frac{\log p}{n_{T}}\left(\left\|\hat{\bDelta}^{(k)}\right\|_{1}^{2}+\left\|\hat{\bDelta}^{(0)}\right\|_{1}^{2}\right)\\
   \leq & 
   \left( \frac{\beta_{0}w_0 \log p}{N} + 2 \sum_{k=1}^{K} \frac{\beta_{k} w_k \log p}{N} + 2 \sum_{k=1}^{K} \frac{n_{S} \alpha_{k} w_k}{N} \frac{\tau_0}{\gamma_0} \frac{\log p}{n_T}\right)\left\|\hat{\bDelta}^{(0)}\right\|_{1}^{2}\\ 
   & + 
    \sum_{k=1}^{K} \left( \frac{2 \beta_{k} w_k \log p}{N} +   \frac{2 n_{S} \alpha_{k} w_k}{N} \frac{\tau_0}{\gamma_0} \frac{\log p}{n_T} \right) \left\|\hat{\bDelta}^{(k)}\right\|_{1}^{2}\\
    \leq &
     \left(  2 \sum_{k=0}^{K} \frac{\beta_{k} w_k \log p}{N} + 2 \sum_{k=1}^{K} \frac{n_{S} \alpha_{k} w_k}{N} \frac{\tau_0}{\gamma_0} \frac{\log p}{n_T}\right)\left\|\hat{\bDelta}^{(0)}\right\|_{1}^{2}
    + 
    \sum_{k=1}^{K} \left( \frac{2 \beta_{k} w_k \log p}{N} +   \frac{2 n_{S} \alpha_{k} w_k}{N} \frac{\tau_0}{\gamma_0} \frac{\log p}{n_T} \right) \left\|\hat{\bDelta}^{(k)}\right\|_{1}^{2}\\
    \leq & 
    \frac{2(\alpha_{\max} \tau_{0} + \beta_{\max}\gamma_0)}{\gamma_{0}}\left[\left( w_0 \frac{\log p}{N} + \sum_{k = 1}^K w_k \frac{n_S}{n_T} \frac{\log p}{N}\right) \left\|\hat{\bDelta}^{(0)}\right\|_{1}^{2} +  \sum_{k = 1}^K w_k \frac{n_S}{n_T} \frac{\log p}{N} \left\|\hat{\bDelta}^{(k)}\right\|_{1}^{2}\right].
\end{align*}
Since $w_k \geq \underline{w} $, Lemma \ref{Thm1Lemma1} together with the triangle inequality yields
\begin{align*}
 & \sum_{t=1}^{3} R_{t}(\widehat{\bDelta}) \cdot \left(\frac{2(\alpha_{\max} \tau_{0} + \beta_{\max} \gamma_{0})}{\gamma_{0}}\right)^{-1}\\
 \leq & 
 \sum_{k = 1}^K w_k \frac{n_S}{n_T} \frac{\log p}{N} \left\|\hat{\bDelta}^{(k)}\right\|_{1}^{2} +  \left( w_0  + \sum_{k = 1}^K w_k \frac{n_S}{n_T} \right) \frac{\log p}{N}\left\|\hat{\bDelta}^{(0)}\right\|_{1}^{2}\\
 \leq
 & \frac{n_S}{n_T} \frac{\log p}{N} \sum_{ k = 1}^K \frac{\lk^2 w_k^2}{\lk^2 w_k} \left\|\hat{\bDelta}^{(k)}\right\|_{1}^{2} + \frac{\log p }{N} \frac{1}{n_T} \left( n_T w_0 + \sum_{k = 1}^K n_S w_k\right) \left\|\hat{\bDelta}^{(0)}\right\|_{1}^{2}\\
 = &  \frac{n_S}{n_T} \frac{\log p}{N}  \left(  \sum_{ k = 1}^K \frac{\lk^2 w_k^2}{\lk^2 w_k} \left\|\hat{\bDelta}^{(k)}\right\|_{1}^{2} + 
  \frac{\lz^2 \sum_{k = 0}^K \frac{n_S}{N} w_k^2}{ (\lz^2 \sum_{k = 0}^K \frac{n_S}{N} w_k^2)/ ( (n_T/n_S) w_0 + \sum_{k = 1}^K  w_k)}\left\|\hat{\bDelta}^{(0)}\right\|_{1}^{2}
 \right) \\
 \leq &  \frac{n_S}{n_T} \frac{\log p}{N} \cdot \frac{1}{\lk^2 \underline{w} \wedge  (\lz^2 \sum_{k = 0}^K \frac{n_S}{N} w_k^2)/ ( (n_T/n_S) w_0 + \sum_{k = 1}^K  w_k)}
 \left(\sum_{k=1}^{K} \lk w_k \left\|\hat{\bDelta}^{(k)}\right\|_{1}+\lambda_{0}\sqrt{\sum_{k=0}^{K}\frac{n_{S}}{N}w_{k}^2}\left\|\hat{\bDelta}^{(0)}\right\|_{1}\right)^{2}\\
 \leq &  \frac{n_S}{n_T} \frac{\log p}{N} \cdot \frac{1}{\lk^2 \underline{w} \wedge  (\lz^2 \sum_{k = 0}^K \frac{n_S}{N} w_k^2)/ ( (n_T/n_S) w_0 + \sum_{k = 1}^K  w_k)}
 \left[32 \lambda_{0}^{2} s (\sum_{k=0}^{K}\frac{n_{k}}{N}w_{k}^2)\left\|\hat{\bDelta}^{(0)}\right\|_{2}^{2}+32\left( \sum_{k=1}^{K} \lk w_k h_{k}\right)^2 \right],
\end{align*}

Recall that we introduce the shorthand
\begin{align*}
u_n &=  \frac{64(\alpha_{\max} \tau_{0} + \beta_{\max}\gamma_{0})}{\gamma_{0}} \frac{n_S}{n_T} \frac{\log p}{N} \cdot \frac{\lz^2 s (\sum_{k=0}^{K}\frac{n_{k}}{N}w_{k}^2)}{\lk^2 \underline{w} \wedge [ (\lz^2 \sum_{k = 0}^K \frac{n_S}{N} w_k^2)/ ( (n_T/n_S) w_0 + \sum_{k = 1}^K  w_k)]}\\
v_n &= \frac{64(\alpha_{\max} \tau_{0} + \beta_{\max}\gamma_{0})}{\gamma_{0}} \frac{n_S}{n_T} \frac{\log p}{N} \cdot \frac{\sum_{k=1}^{K} \lk w_k h_{k}}{\lk^2 \underline{w} \wedge [ (\lz^2 \sum_{k = 0}^K \frac{n_S}{N} w_k^2)/ ( (n_T/n_S) w_0 + \sum_{k = 1}^K  w_k)]}.
\end{align*}
Therefore collecting all the pieces together, we have
\begin{align*}
& \mathcal{L}\left(\btheta^{*}+\hat{\bDelta}\right)-\mathcal{L}\left(\btheta^{*}\right)-\left\langle\nabla \mathcal{L}\left(\btheta^{*}\right), \hat{\bDelta}\right\rangle \\
& \geq \left(\frac{\alpha_{\min }^{2}}{\gamma_{0}}  \sum^{K}_{k=0} \frac{n_k w_k}{N} - u_n\right) \left\|\hat{\bDelta}^{(0)}\right\|_{2}^{2}+ \frac{\alpha_{\min }^{2}}{\gamma_{0}}\sum_{k=1}^{K} \frac{n_{k} w_k}{N} \left\|\hat{\bDelta}^{(k)}\right\|_{2}^{2} -\frac{2 \alpha_{\max}}{\gamma_{0}} \sum^{K}_{k=1} \frac{n_{k}w_k}{N}\lambda_{T} \left\|\hat{\bDelta}^{(k)}\right\|_{1} - v_n  \sum_{k=1}^{K} \lk w_k h_{k},
\end{align*}
which finishes the proof.

\subsection{Proof of Lemma \ref{lemma:sort}}

We begin by proving the first statement, which follows directly from the proof of Lemma 2 in (\cite{shalev2006efficient}). Assume by contradiction that $w_{j}^{\prime}=0$ yet $w_{l}^{\prime}>0$. Notice that if we interchange $w_{j}$ and $w_{l}$ in the problem (\ref{weight_prob}), the constraint (\ref{weight_prob:constr}) still holds. Therefore, by the optimality condition of the problem (\ref{weight_prob}), we have
$$
\begin{aligned}
0 \geqslant & {\left[\sum_{k=0}^{K}\left(w_{k}^{\prime}\right)^{2} s \frac{\log p}{n_{k}}+c_{\bSig} \sum_{k=1}^{k} w_{k}^{\prime} h_{k} \sqrt{\frac{\log p}{n_{T}}}\right] } \\
& -\left[\sum_{k=0}^{K}\left(w_{k}^{\prime}\right)^{2} s \frac{\log p}{n_{k}}+c_{\bSig} \sum_{k \neq j, l}^{k} w_{k}^{\prime} h_{k} \sqrt{\frac{\log p}{n_{T}}}+c_{\bSig}\left(w_{j}^{\prime} h_{l}+w_{l}^{\prime} h_{j}\right) \sqrt{\frac{\log p}{n_{T}}}\right] \\
= & c_{\bSig}\left[w_{j}^{\prime}\left(h_{j}-h_{l}\right)+w_{l}^{\prime}\left(h_{l}-h_{j}\right)\right] \sqrt{\frac{\log p}{n_{T}}} \\
 =&c_{\bSig}\left[w_{l}^{\prime}\left(h_{l}-h_{j}\right)\right] \sqrt{\frac{\log p}{n_{T}}}
\end{aligned}
$$

Since $c_{\bSig} \sqrt{\frac{\operatorname{logp}}{n_{T}}}>0$, we have $h_{l} \leqslant h_{j}$, which contradicts the fact that $h_{l}>h_{j}$.

Then we proceed to prove the second statement. Assume by contradiction that $w_{0}^{\prime}=0$. Then by (\ref{weight_prob:KKT2}) we have $\xi=-\eta_{0} \leqslant 0$. By the constraint (\ref{weight_prob:constr}), we have $\sum_{k=1}^K w_k^{\prime} =1$. Therefore, following similar arguments as those in the proof of Corollary \ref{choice_of_w}, we can show that $\rho>0$ and
$$
\frac{n_{S}}{2 s \log p} \sum_{k=1}^{\rho}\left(\xi-c_{\bSig} h_{(k)} \sqrt{\frac{\log p}{n_{T}}}\right)=1
$$
which further implies that 
$$
\xi = \frac{1}{\rho}\left( \frac{2s\log p}{n_S}  + c_{\bSig} \sum^{\rho}_{k=1} h_{(k)} \sqrt{\frac{\log p}{n_T}}\right) > 0,
$$
which contradicts the fact that $\xi \leq 0$.

\section{Additional Theoretical Results}
\label{additional_theory}
\subsection{Discussion on the Folded-concave Penalty}
\label{SCAD}
Folded-concave penalty, such as the SCAD penalty \citep{fan2001variable} or the MCP penalty \citep{zhang2010nearly}, is a class of non-convex penalties which satisfies the following four properties: 
\begin{enumerate}
    \item $\mR_{\lambda}(t)$ is increasing and concave in $t \in[0, \infty)$ with $\mR_{\lambda}(0)=0$;
    \item $\mR_{\lambda}(t)$ is differentiable in $t \in(0, \infty)$ with $\mR_{\lambda}^{\prime}(0):=\mR_{\lambda}^{\prime}(0+) \geq a_1 \lambda$;
    \item $\mR_{\lambda}^{\prime}(t) \geq a_1 \lambda$ for $t \in\left(0, a_2 \lambda\right]$;
    \item $\mR_{\lambda}^{\prime}(t)=0$ for $t \in[a \lambda, \infty)$ with the pre-specified constant $a>a_2$.
\end{enumerate}
Where $a_1$ and $a_2$ are two fixed positive constants. The derivative of the SCAD penalty is
$$
\mR_\lambda^{\prime}(t)=\lambda I_{\{t \leq \lambda\}}+\frac{(a \lambda-t)_{+}}{a-1} I_{\{t>\lambda\}} \quad \text { for some } a>2
$$
and the derivative of the MCP is $\mR_\lambda^{\prime}(t)=\left(\lambda-\frac{t}{a}\right)_{+}$, for some $a>1$. It is easy to see that $a_1=a_2=1$ for the SCAD, and $a_1=1-a^{-1}, a_2=1$ for the MCP.

This family of penalties applies a stronger penalty to small elements to promote sparse solutions while adding a constant penalty to large elements to prevent introducing bias, thereby it is a desirable choice for constructing the weights for feature-wise adaptive transfer learning.

\subsection{Incoherence Assumptions in Theorem \ref{oracle}}
\label{mutual_cond:sec}
In the partially detectable case, we adopt a mild incoherence condition on the non-detectable set. Differing from the mutual incoherence conditions commonly seen in the literature \citep{wainwright2019high}, our conditions only apply to the non-detectable sets \(A_0\) and \(\{A_k\}_{k=1}^K\). Therefore, these conditions are trivially satisfied if all transferable patterns are detectable. The following three conditions aim to ensure that no elements of the detectable non-active set are falsely included, i.e., \(\hbz_{\tSz^c} = \hbz_{\text{ora}, \tSz^c} = \boldsymbol{0}\) with $\tSzc = S_0^c \backslash (\cup_{k=1}^K A_k)$. Specifically, we assume there exist a positive constant \(\alpha \in (0,1)\) such that 
\begin{align}
&\left\|\tilde{\bX}_{\tSzc}^{\top} \tilde{\bX}_{\tSz}\left(\tilde{\bX}_{\tSz}^{\top} \tilde{\bX}_{\tSz}\right)^{-1}\hbzz_{\tAz}\right\|_{\infty}   < 1 - 4 \alpha, \label{mutual_cond1}\\
&\frac{\lambda_{1}}{\lambda_{0}}  \left\| \frac{1}{K}\sum_{k=1}^{K} \tbXtSzc^{\top}\tbH^{(K,k)}_{\tSz}\bXktSk\left[\left(\bXktSk\right)^{\top}\bXktSk\right]^{-1}\hbzk_{\tAk}\right\|_{\infty} \leq \alpha,\label{mutual_cond2} \\
&\frac{\lambda_{1}}{\lambda_{0}} \left\| \sum_{k=1}^K \left(\bX_{\tSzc}^{(k)}\right)^{\top} \bXktSk\left[\left(\bXktSk\right)^{\top} \bXktSk\right]^{-1} \hbzk_{\tAk} \right\|_{\infty} \leq \alpha. \label{mutual_cond3}
\end{align}
where recall that we define $\tbH^{(K,k)}_{\tSz} =  \tilde{\bX}_{\tSz}\left(\tilde{\bX}_{\tSz}^{\top} \tilde{\bX}_{\tSz} / K\right)^{-1}\left(\bXktSz\right)^{\top}$.

Condition (\ref{mutual_cond1}) relates to the coefficient matrix \(\tilde{\bX}_{\tSzc}^{\top} \tilde{\bX}_{\tSz}(\tilde{\bX}_{\tSz}^{\top} \tilde{\bX}_{\tSz})^{-1}\), the ordinary least squares estimator for predicting the columns of \(\tbX_{\tSz^c}\) using the columns in \(\tbX_{\tSz}\). There is an additional term $\hbzz_{\tAz}$ that extracts the columns related to the non-detectable set from the coefficient matrix. This condition requires that the features in the detectable non-active set \(\tSzc = S_{0}^c \backslash (\cup_{k=0}^K A_k)\) do not exhibit perfect linear correlation with those in the non-detectable set \(\tAz = \cup_{k=0}^K A_k\). 
As we usually choose $\lz$ and $\lk$ to be of the same order (c.f. Corollary \ref{partial_rate}), conditions (\ref{mutual_cond2}) and (\ref{mutual_cond3}) essentially also impose restrictions on the linear correlations between the detectable non-active set \(\tSzc = S_{0}^c \backslash (\cup_{k=1}^K A_k)\) and the non-detectable set \(\tAk =  \cup_{k=0}^K A_k = \tAz\).


On the other hand, to ensure no false inclusion of the detectable non-transferable set, i.e., \(\hbdk_{\tSkc} = \hbdk_{\text{ora}, \tSkc} = \boldsymbol{0}\), we impose another three conditions for each \(k \in [K]\):
\begin{align*}
&\left\|\left(\bXktSkc\right)^{\top} \bXktSk\left[\left(\bXktSk\right)^{\top} \bXktSk\right]^{-1} \hbzk_{\tAk}\right\|_{\infty}< 1 - 4 \alpha,\\
&\frac{\lambda_{0}}{\lambda_{1}}\left\|\left(\tilde{\bX}_{\tSkc}^{(k)}\right)^{\top} \tbXktSz\left(\tilde{\bX}_{\tSz}^{\top} \tilde{\bX}_{\tSz}\right)^{-1}\hbzz_{\tAz}\right\|_{\infty} \leq \alpha, \\
&\left\| \frac{1}{K}\sum_{k=1}^{K} \left(\tilde{\bX}_{\tSkc}^{(k)}\right)^{\top}\bH_{\tSz}^{(k,k)} \bXktSk\left[\left(\bXktSk\right)^{\top}\bXktSk\right]^{-1} \hbzk_{\tAk}\right\|_{\infty} \leq \alpha,
\end{align*}
where $\tbH_{\tSz}^{(k,k)} =  \tbXktSz\left(\tilde{\bX}_{\tSz}^{\top} \tilde{\bX}_{\tSz}/K\right)^{-1}\left(\bXktSz\right)^{\top}$.

Similar to the previously discussed conditions, these conditions involve an incoherence assumption on the relationship between the detectable transferable set $\tSkc = S_k^c \backslash (\cup_{k=0}^K A_k)$ and the non-detectable set $\tAk = \tAz = \cup_{k=0}^K A_k$. 

Overall, these assumptions are considered mild, as they mainly apply to non-detectable sets. When the transferable structure is mostly detectable, or the elements of the non-detectable set are orthogonal to other elements, these assumptions hold trivially.



\subsection{Refined Analysis on the Non-detectable Set}
\label{app:non-detectable_set}
As shown in the proof of Corollary (\ref{partial_rate}), if the detectable set is orthogonal to the non-detectable set in the sense that 
$\bXk_{S_k \backslash A} \perp \bXk_{A}$, we have 
{\small
\begin{align*}
\sum_{k=1}^{K}\hat{\boldsymbol{B}}_{\tSz, \tSk} \hbzk_{A} := \sum_{k=1}^{K}\left(\bXktSz\right)^{\top} \bXztSk\left[\left(\bXktSk\right)^{\top}\bXktSk\right]^{-1}  \hbzk_{A} = \sum_{k=1}^K \hbzk_{A}.
\end{align*}
}
Therefore, for the estimation on the non-detectable set $A$, we have 
$$
\hbz_{A} - \bz_{A} = \left[\frac{\left(\bXz_{A}\right)^{\top} \bXz_{A}}{n_T}\right]^{-1} \left[\frac{\left(\bXz_{A}\right)^{\top} \bez}{n_T}\right] + \left[\frac{\left(\bXz_{A}\right)^{\top} \bXz_{A}}{N}\right]^{-1}\left(-\frac{1}{2}\lz\hbzz_{A} + \frac{1}{2}\lk \sum_{k=1}^{K}\hbzk_{A}\right).
$$
Notice that under mild conditions, $\|[(\bXz_{A})^{\top} \bXz_{A} / n_T]^{-1}\|_{\infty}$ is bounded with high probability, whereas $\|[(\bXz_{A})^{\top} \bXz_{A} / N]^{-1}\|_{\infty}$ can be of the order $O(N/n_T)$. If we set \( \lambda_0 = \lambda_k = c_0 \sqrt{\log p / n_T} \) as specified in Corollary \ref{partial_rate}, the worst-case \(\ell_\infty\)-norm bound for the second term is of order \( O\left( K \frac{N}{n_T} \sqrt{\frac{\log p}{n_T}} \right) \). This bound becomes increasingly loose as \( N / n_T \to \infty \), reflecting an over-penalization of the non-detectable set. The root cause of this issue is that, in the current setup, we cannot precisely identify the non-detectable set and thus must apply a uniform penalty across all features, inevitably excessively penalizing non-detectable components. To improve the bound, we can use a generalized folded concave penalty for constructing the weights. For instance, consider the following modified SCAD penalty:
\[
\mR_\lambda^{\prime}(t) = \lambda \, I_{\{t \leq \lambda^\prime\}} + \frac{(a \lambda - t)_{+}}{a - 1} \, I_{\{t > \lambda^\prime\}} \quad \text{for some } a > 2,
\]
where \(\lambda\) controls the penalty magnitude, and \(\lambda^\prime\) determines the threshold for feature separation. Under the conditions of Corollary \ref{partial_rate}, with the Lasso as the initial estimator, we can set \(\lambda_0 = c_0 \sqrt{\log p / N}\) and \(\lambda_0^\prime = c_0 \sqrt{\log p / n_T}\) for the target sample, while setting \(\lambda_k = c_0 \frac{n_S}{N} \sqrt{\log p / n_S}\) and \(\lambda_k^{\prime} = c_0 \sqrt{\log p / n_T}\) for each source sample. With this choice of parameters, the bound for the second term can be improved to \( O\left( \frac{N}{n_T} \sqrt{\frac{\log p}{n_S}} \right) \), resulting in a tighter bound for the estimation error on the non-detectable set.

\subsection{Comparison with Existing Methods}
\label{sec:comparison}
{In Table \ref{tab:comparison}, we compare the \textit{F-AdaTrans} algorithm with the existing adaptive transfer learning algorithm in terms of key assumptions, estimation rate, and algorithmic guarantee. We have the following observations:}\\
$\bullet$ \textit{Key Assumptions}: TransLasso requires a minimum cardinality assumption (Eq. (20) in \cite{li2022transfer}) on the non-transferable sources for consistent detection. \textit{TransHDGLM} instead requires a bounded $\ell_{2}$ norm of the non-transferable signals to guarantee an accurate detection. In general, these assumptions are imposed sample-wise. On the other hand, F-AdaTrans requires a more detailed condition on the feature-wise structure, like an incoherence assumption to obtain the sub-oracle estimator and the minimum signal assumption for achieving the oracle rate.
\\
$\bullet$ \textit{Estimation rate}:
Under the given conditions, the \textit{F-AdaTrans} estimator achieves a sharper estimation rate when the data exhibits a feature-wise transferable pattern. To see this, we first discuss an ideal scenario where the entire transferable structure is detectable. When $\max_k s_k \gg s$, \textit{F-AdaTrans} guarantees a worse-case rate of $O(s \log s / n_T)$, which is strictly faster than the TransLasso's rate $O(r_{\bSig} s \log p /n_T + \log K / n_T)$ when $p \gg s$ or $r_{\bSig} \to \infty$. Here $r_{\bSig}$ measures the heterogeneity in the covariate covariance matrices and can be of order $O(\sqrt{p})$ with well-conditioned covariance matrices \citep{he2024transfusion}. Unlike \textit{TransLasso}, \textit{F-AdaTrans} is robust to covariate shift, owing to its fused-regularization structure. When \( s \gg \max_k s_k \), \textit{F-AdaTrans} achieves a rate of \( O(s \log s / N) \) under mild incoherence conditions, which is significantly faster than \textit{TransLasso}'s rate of \( O(s \log p / N + r_{\bSig} \max_k s_k \log p / n_T + \log K / n_T) \), especially as the latter is constrained by the target sample size \( n_T \). \textit{F-AdaTrans} avoids this bottleneck by excluding non-transferable features from the estimation. While \textit{TransHDGLM} improves on \textit{TransLasso} by removing dependence on \( c_{\bSig} \) and \( \log K \), it still achieves a slower rate than \textit{F-AdaTrans} for similar reasons. Additionally, if the non-transferable features are orthogonal to the active features in the target model, \textit{F-AdaTrans} achieves a near-oracle rate regardless of the size of the non-transferable set, whereas \textit{TransLasso} and \textit{TransHDGLM} experience substantial performance degradation. In a more realistic case where the transferable structure is only partially detectable, the estimation cost associated with the non-detectable set affects the performance of \textit{F-AdaTrans}. When the non-detectable set is small and weakly correlated with the detectable set, as in the scenarios discussed in Corollary \ref{partial_rate}, \textit{F-AdaTrans} still achieves a near-oracle rate on the detectable active set, performing effectively on the active target parameters, though less efficiently on the non-detectable set. However, if the non-detectable set is large and strongly correlated with the detectable set, the performance of \textit{F-AdaTrans} may lag behind other methods.

$\bullet$ \textit{Convergence guarantee}: both \textit{F-AdaTrans} and \textit{TransLasso} can be formulated into standard Lasso regression problems and thus can be efficiently solved using soft-thresholding with guaranteed convergence \citep{wu2008coordinate}. On the other hand,
\textit{TransHDGLM} involves solving a nonsmooth optimization problem with $2K+2$ constraints, which is computationally intense. In the paper, they provide an approximate solver for the optimization problem, but the convergence guarantee has not yet been studied.

\begin{table}[htbp]
  \centering
 \caption{Comparison between \textit{F-AdaTrans} and existing methods.}
\resizebox{\columnwidth}{!}{%
    \begin{tabular}{cccc}
    \toprule
Method &    Key Assumption   &  Estimation Rate   & Convergence Guarantee  \\
\midrule  
   \textit{TransLasso} \citep{li2022transfer}  & $ \sqrt{n_T} \gg \log p \wedge \log K$, minimum cardinality assumption  & $O(s \log p / N + r_{\bSig}  (s \wedge \max_{k} s_k) \log p / n_T + \log K / n_T)$ & Yes \\ 
   \textit{TransHDGLM} \citep{li2023estimation} & $\sqrt{n_T} \gg s \log p$,  $\max_{k \in [K]} \|\bdk\|_{2} = O(1)$  & $O(s \log p / N + (s \wedge \max_{k} s_k) \log p / n_T + 1 / n_T)$ & No  \\
    \multirow{2}{*}{\textit{F-AdaTrans} (Ours)} & $n_T > s$, $n_S > \max_{k \in [K]} |S_k|$, minimum signal assumption     & $O(\kappa_{F}^2 s \log s / N)$ & Yes \\
    & $n_T > (s^2+a)\log p$, $n_S > \max_{k \in [K]} |S_k|$, incoherence assumption   & $\tilde{\kappa}_{F}^2\|\hat{\boldsymbol{\Omega}}_{\tSz, N}\|_{\infty}^2 \frac{s\log (s+a)}{N} + \|\tilde{\boldsymbol{\Omega}}_{\tSz, N}\|_{\infty}^2   (1 + \|\hat{\boldsymbol{B}}\|_{\infty})^2 \frac{a\log p}{n_{T}}$ & Yes \\
    \bottomrule
    \end{tabular}%
    }
  \label{tab:comparison}%
\end{table}%

\subsection{Additional Explanation of the results in Theorem \ref{choice_of_w}}
\label{weight_change}


With a slight abuse of notation, we denote $h_0=0$ and $\boldsymbol{h} = (h_0, \dots, h_K) \in \mathbb{R}^{K+1}$. We will show that $\boldsymbol{w}^\prime \in \mathbb{R}^{K}$, as a function of $\boldsymbol{h}$, is Lipschitz continuous in $\boldsymbol{h}$. We start by revisiting the problem (\ref{choose_weight}). 
We can formulate the problem as
\begin{align}
\label{prob:matrix_form}
   \min_{\boldsymbol{w}^\prime} Q(\boldsymbol{w}^\prime, \boldsymbol{h})= \min_{\boldsymbol{w}^\prime} \left\{ \frac{1}{2} (\boldsymbol{w}^\prime)^\top \boldsymbol{D} \boldsymbol{w}^\prime + \boldsymbol{h}^\top \boldsymbol{w}^\prime\right\}, \ s.t. \  \boldsymbol{w}^\prime \in \boldsymbol{\Delta}^K .
\end{align}
where we denote $\boldsymbol{D} := \frac{2s \log p}{c_{\bSig} \sqrt{\log p / n_T}} \diag (\frac{1}{n_T}, \frac{1}{n_S}, \dots, \frac{1}{n_S})$. Now for any $\boldsymbol{h}_{1} \in \mathbb{R}^{K+1}$ and  $\boldsymbol{h}_{2} \in \mathbb{R}^{K+1}$, let $\boldsymbol{w}_{1}^\prime  \in \mathbb{R}^{K+1}$ and $\boldsymbol{w}_2^\prime  \in \mathbb{R}^{K+1}$ be the corresponding optimal solutions in problem (\ref{prob:matrix_form}). Then by the optimality conditions of problem (\ref{prob:matrix_form}), we have 
\begin{align*}
\left(\boldsymbol{w}_{2}^\prime - \boldsymbol{w}_{1}^\prime\right)^\top \nabla_{\boldsymbol{w}^\prime} Q (\boldsymbol{w}_1^\prime, \boldsymbol{h}_1) = \left(\boldsymbol{w}_{2}^\prime - \boldsymbol{w}_{1}^\prime\right)^\top (\boldsymbol{D}\boldsymbol{w}^\prime_{1} + \boldsymbol{h}_{1}) \geq 0
\end{align*}
and similarly
\begin{align*}
\left(\boldsymbol{w}_{1}^\prime - \boldsymbol{w}_{2}^\prime\right)^\top \nabla_{\boldsymbol{w}^\prime} Q (\boldsymbol{w}_2^\prime, \boldsymbol{h}_2) = \left(\boldsymbol{w}_{1}^\prime - \boldsymbol{w}_{2}^\prime\right)^\top (\boldsymbol{D}\boldsymbol{w}^\prime_{2} + \boldsymbol{h}_{2}) \geq 0
\end{align*}
which subsequently implies
\begin{align*}
    \left(\boldsymbol{w}_{1}^\prime - \boldsymbol{w}_{2}^\prime\right)^\top \left(\boldsymbol{h}_{2} - \boldsymbol{h}_{1}\right) \geq \left(\boldsymbol{w}_{1}^\prime - \boldsymbol{w}_{2}^\prime\right)^\top \boldsymbol{D} \left(\boldsymbol{w}_{1}^\prime - \boldsymbol{w}_{2}^\prime\right) \geq \Lambda_{\min}(\boldsymbol{D}) \left\|\boldsymbol{w}_{1}^\prime - \boldsymbol{w}_{2}^\prime\right\|_2^2 > 0
\end{align*}
as $\boldsymbol{D}$ is positive definite for fixed $s$, $p$, $n_T$ and $n_S$. Therefore, the solution map $\boldsymbol{w}^\prime$ is Lipschitz continuous in $\boldsymbol{h}$.

\section{Implementation Details and Additional Empirical Results}
\label{Numerical}
\subsection{Implementation Details}\label{app:implement}
\textit{F-AdaTrans} is executed by first rewrite it as a weighted LASSO problem \citep{zou2006adaptive}, and then solved using the R package \textit{glmnet} with standard settings. For \textit{S-AdaTrans}, we first solve it without the constraint using \textit{glmnet} then check the solution feasibility. If the solution is feasible then the problem is solved, otherwise we solve the orignal constrained problem~\eqref{P:sample-transfer} using \textit{CVXR}. 
We also use  \textit{CVXR}  to compute the  weights for \textit{S-AdaTrans} defined by~\eqref{choose_weight}. It is worth mentioning that Problem (\ref{obj2}) can be written as a standard Lasso regression problem and can be efficiently solved.
By exploring the problem structure, low complexity algorithms for both~\eqref{P:sample-transfer} and  \textit{S-AdaTrans} can also be designed based on solving the KKT system and ADMM, respectively. 

All hyperparameters are adjusted through 3-fold cross-validation. The hyperparameters $\lz$ and $\lambda_{k}$ (for the $k$-th source's non-transferable signal $\bdk$) are collectively optimized by setting $\lz = \lambda \sqrt{{\log p}/{N}}$, $\lk = \lambda \frac{n_k}{N} \sqrt{{\log p}/{n_T}}$, and tuning $\lambda$ accordingly.  For the SCAD penalty, we set the parameter $a$ at $3.7 \times (K/2)$. This choice is informed by the recommendation of \citep{fan2001variable}, who suggest $3.7$ based on a Bayesian statistical perspective and empirical studies. The factor $(K/2)$ serves as an adjustment to avoid over-penalization with increasing $K$. \textit{TransGLM} is implemented using R package \textit{transglm} provided in the paper \citep{tian2022transfer} with standard configurations. \textit{TransLasso} is implemented using the code from the GitHub link https://github.com/saili0103/TransLasso. \textit{TransHDGLM} is implemented based on the iterative algorithm provided in the supplementary material of \cite{li2023estimation}.
            
\subsection{Additional Simulation Result}
\label{app:sim-suppl}

\begin{figure}
    \centering
    \includegraphics[width=0.8\linewidth]{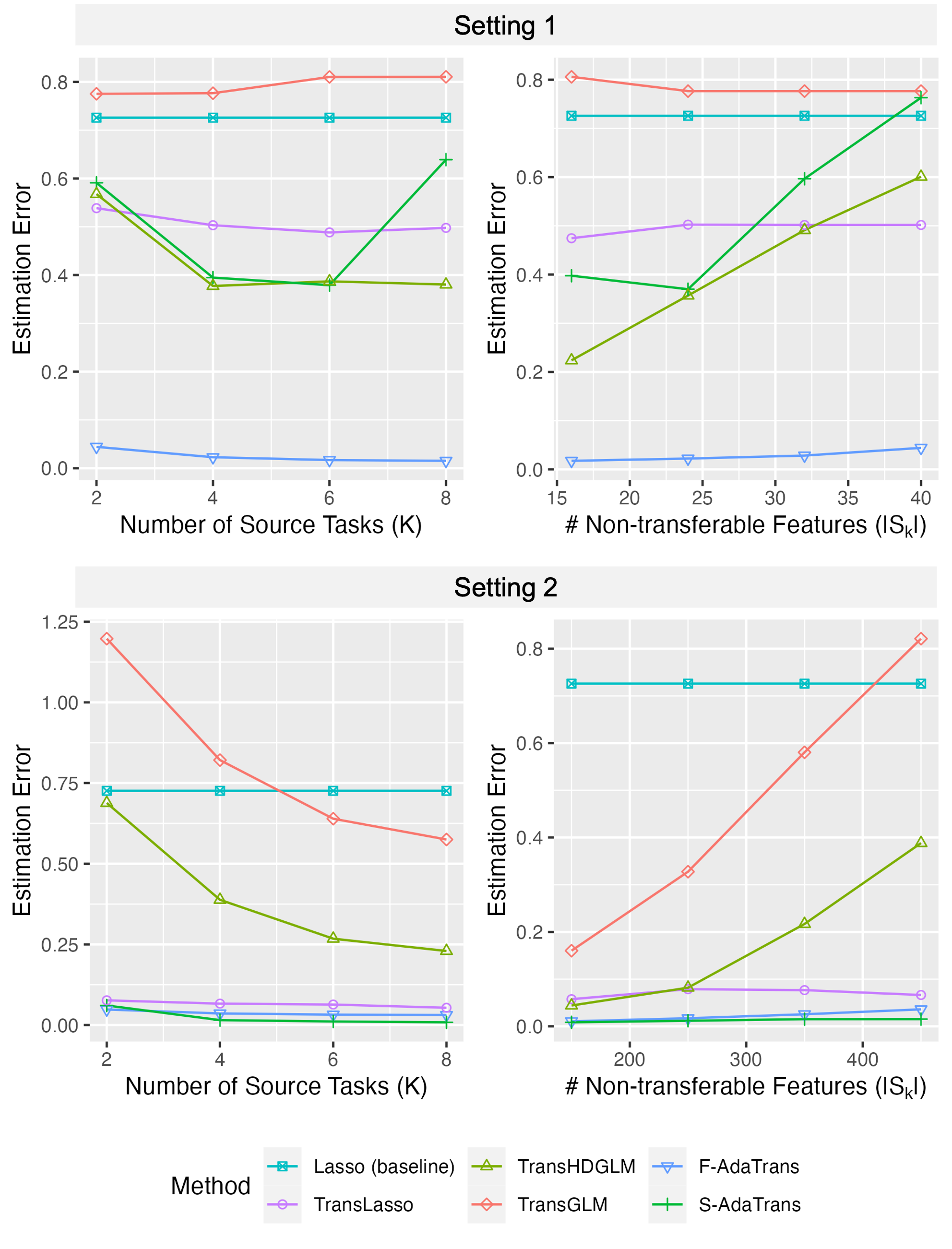}
    \caption{Average estimation error against $K$ and $s_k$ under under feature-wise (Setting 1) and sample-wise (Setting 2) adaptive transfer setting}
        \label{fig:skandK}
\end{figure}

\begin{figure}[htbp]
        \centering
        \includegraphics[width=0.8\linewidth]{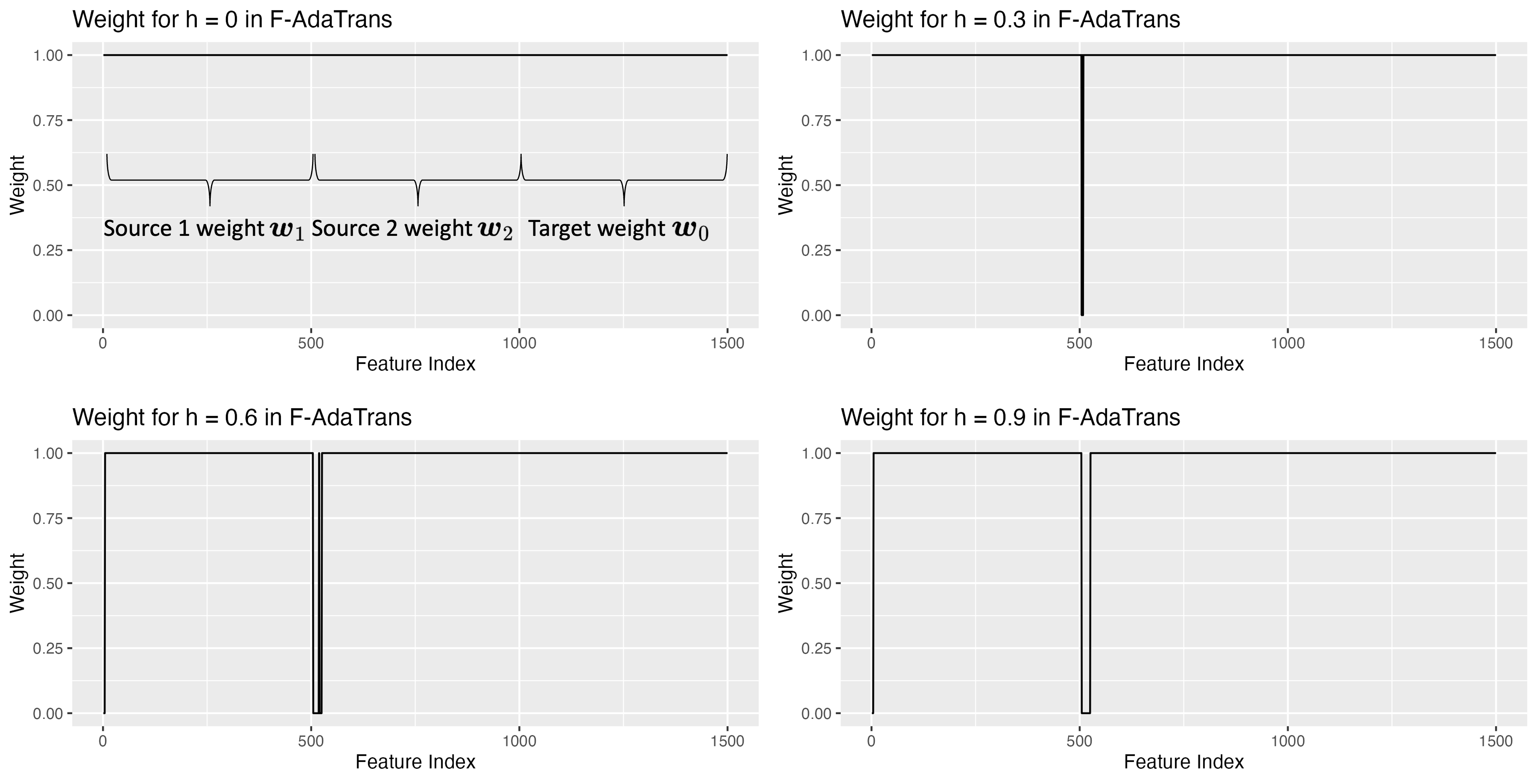}
    \vspace{.5cm}
        \caption{Feature selection result of \textit{F-AdaTrans} under feature-wise adaptive transfer setting (Setting 1) in Section \ref{simulation}.}
        \label{fig:FAdaFusion-weight}
\end{figure}

In Figure~\ref{fig:FAdaFusion-weight}, we plot the value of the weights computed by \textit{F-AdaTrans}, $\boldsymbol{w}_k \in \mathbb{R}^{p}$, $k=0,\dots,K$, in Setting 1 (i). Here we set $K=2$ and $n_S=250$. As $h^\wedge$ increases, the detection accuracy of the support of $\bdk$ gets higher (recall that $w_{kj} = 0$ means the $j$th feature of the $k$th source is not transferable). 

\begin{figure}[htbp]
        \centering
        \includegraphics[width=0.8\linewidth]{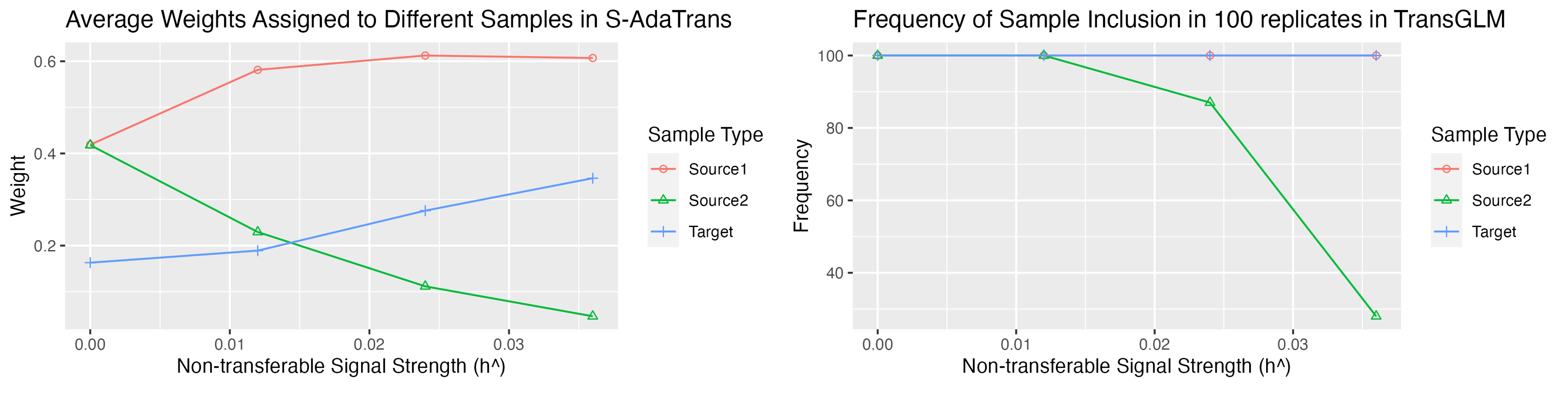}
        \vspace{.5cm}
        \caption{Source selection result of \textit{S-AdaTrans} and \textit{TransGLM} under sample-wise adaptive transfer setting (Setting 2) in Section \ref{simulation}.}
        \label{fig:SAdaFusion-weight}
\end{figure}

Figure~\ref{fig:SAdaFusion-weight} plots the weight used by \textit{S-AdaTrans}  and \textit{TransGLM} in Setting 2(i). Here we also set $K=2$ and $n_S=250$. The left figure plots the normalized weights of \textit{S-AdaTrans} averaged over 100 trials, showing as $h^\wedge$ gets stronger, the less informative source 2 is gradually excluded from the estimation. The right figure plots the frequency of each source sample being included in \textit{TransGLM} over the 100 trials. In each individual trial, each source is either included or discarded in the transfer learning procedure. The figure shows that as $h^\wedge$ increases, source 1 is always considered informative, but source 2 gets rejected more often. 

\bibliographystyle{plainnat}
\bibliography{bibfile}

\end{document}